\documentclass[12pt]{article}

\usepackage[utf8]{inputenc}
\usepackage[T1]{fontenc}

\usepackage{amsmath,
            mleftright,
            comment,
            amssymb,
            amsthm,
            nicefrac,
            xcolor,
            enumerate,
            authblk,
            color,
            etoolbox,
            mathrsfs,
            mathtools,
            bbm,
            xparse,
            stmaryrd,
            geometry,
            enumitem,
            mathtools,
            marginnote,
            tikz,
            xurl,
            graphicx,
            scalerel,
            relsize,
            stackengine,
            hyperref
            }

\definecolor{darkblue}{rgb}{0,0,0.75}
\hypersetup{colorlinks=true,linkcolor=darkblue,breaklinks=true,linktocpage=true}

\usepackage[colorinlistoftodos]{todonotes}
\usepackage[nocompress,sort]{cite}
            
\usepackage[sort,capitalize]{cleveref}

\crefformat{equation}{(#2#1#3)}
\crefname{enumi}{item}{items}
\crefname{equation}{}{}

\theoremstyle{plain}
\newtheorem{theorem}{Theorem}[section]
\newtheorem{lemma}[theorem]{Lemma}

\theoremstyle{remark}

\theoremstyle{definition}
\newtheorem{definition}[theorem]{Definition}

\numberwithin{equation}{section}

\crefname{subsection}{Subsection}{Subsections}

\DeclareFontEncoding{LS1}{}{}
\DeclareFontSubstitution{LS1}{stix}{m}{n}
\DeclareMathAlphabet{\mathscr}{LS1}{stixscr}{m}{n}

\newcommand{\cB}{\mathcal{B}}

\newcommand{\cF}{\mathcal{F}}

\newcommand{\cU}{\mathcal{U}}

\newcommand{\fB}{\mathfrak{B}}
\newcommand{\fC}{\mathfrak{C}}

\newcommand{\fJ}{{\bf\mathfrak{J}}}

\newcommand{\fL}{\mathfrak{L}}

\newcommand{\fa}{\mathfrak{a}}
\newcommand{\fb}{\mathfrak{b}}
\newcommand{\fc}{\mathfrak{c}}
\newcommand{\fd}{\mathfrak{d}}

\newcommand{\ff}{\mathfrak{f}}

\newcommand{\fm}{\mathfrak{m}}

\newcommand{\fp}{\mathfrak{p}}
\newcommand{\fq}{\mathfrak{q}}
\newcommand{\fr}{\mathfrak{r}}

\newcommand{\ft}{\mathfrak{t}}
\newcommand{\fu}{\mathfrak{u}}

\newcommand{\fx}{\mathfrak{t}}

\newcommand{\scrL}{\mathscr{L} \cfadd{def:lin_interp}}

\newcommand{\scrf}{\mathscr{f}}
\newcommand{\scru}{\mathscr{u}}

\newcommand{\with}{\curvearrowleft}
\newcommand{\lrSpace}{\ensuremath{\mkern-1.5mu}}

\newcommand{\eps}{\varepsilon}

\newcommand{\dpp}{\text{.}}
\newcommand{\dc}{\text{,}}
\newcommand{\dx}{\, {\rm d}}
\newcommand{\dxx}{{\rm d}}

\newcommand{\normmm}[1]{{\left\vert\kern-0.25ex\left\vert\kern-0.25ex\left\vert #1 
    \right\vert\kern-0.25ex\right\vert\kern-0.25ex\right\vert}
    \cfadd{DNN_norm}}

\DeclarePairedDelimiter{\pr}{(}{)}
\newcommand{\bpr}[1]{\pr[\big]{#1}}
\newcommand{\bbpr}[1]{\pr[\Big]{#1}}
\newcommand{\bbbpr}[1]{\pr[\bigg]{#1}}
\newcommand{\bbbbpr}[1]{\pr[\Bigg]{#1}}

\DeclarePairedDelimiter{\br}{[}{]}
\newcommand{\bbr}[1]{\br[\big]{#1}}
\newcommand{\bbbr}[1]{\br[\Big]{#1}}
\newcommand{\bbbbr}[1]{\br[\bigg]{#1}}
\newcommand{\bbbbbr}[1]{\br[\Bigg]{#1}}

\DeclarePairedDelimiter{\cu}{\{}{\}}
\newcommand{\bcu}[1]{\cu[\big]{#1}}
\newcommand{\bbcu}[1]{\cu[\Big]{#1}}
\newcommand{\bbbcu}[1]{\cu[\bigg]{#1}}

\DeclarePairedDelimiter{\abs}{\lvert}{\rvert}
\newcommand{\babs}[1]{\abs[\big]{#1}}
\newcommand{\bbabs}[1]{\abs[\Big]{#1}}
\newcommand{\bbbabs}[1]{\abs[\bigg]{#1}}

\DeclarePairedDelimiter{\norm}{\lVert}{\rVert\cfadd{DNN_norm}}

\newcommand{\N}{\ensuremath{\mathbb N}}
\newcommand{\Z}{\ensuremath{\mathbb Z}}

\newcommand{\R}{\ensuremath{\mathbb R}}

\newcommand{\E}{\ensuremath{\mathbb E}}
\renewcommand{\P}{\ensuremath{\mathbb P}}
\newcommand{\parallelization}{\operatorname{P} \cfadd{def:generalParallelization}}
\newcommand{\parallelizationSpecial}{\mathbf{P} \cfadd{def:simpleParallelization}}

\newcommand{\Rr}{\mathcal{R}_\mathfrak{r} \cfadd{def:ANNrealization}} 
\newcommand{\Ra}{\mathcal{R}_a \cfadd{def:ANNrealization}}
\newcommand{\1}{\mathbbm{1}}
\newcommand{\ii}{\mathfrak{i} \cfadd{padding}}

\newcommand{\interpolatingDNN}{\mathbf{F}}

\newcommand{\fwpr}{W}
\newcommand{\smallU}{u}
\newcommand{\smallF}{f}

\newcommand{\boundFG}{\mathfrak{L}}
\newcommand{\LipConstF}{L}
\newcommand{\mlp}{U}
\newcommand{\diffBd}{B}

\newcommand{\ANNhatfct}{\H}

\newcommand{\leaky}{\mathscr{a}}

\newcommand{\indexAct}{\nu}
\newcommand{\constantAssumpMainThm}{\kappa}
\newcommand{\constantMainThmDelta}{\eta}
\newcommand{\constantMainThmParam}{c}

\newcommand{\littleMM}{m}

\newcommand{\A}{\mathbf{A} \cfadd{linear}}
\newcommand{\U}{\mathbf{U}}

\newcommand{\F}{\mathbf{F}}
\newcommand{\G}{\mathbf{G}}

\renewcommand{\H}{\mathbf{H}}

\newcommand{\lbd}{u}
\newcommand{\ubd}{v}

\newcommand{\weight}{W \cfadd{linear}}
\newcommand{\bias}{B \cfadd{linear}}
\newcommand{\WNN}{\mathcal{W} \cfadd{def:ANN}}
\newcommand{\BNN}{\mathcal{B} \cfadd{def:ANN}}
\newcommand{\Mmulti}{\mathfrak{M} \cfadd{multi}}

\newcommand{\indexkmin}{0}
\newcommand{\indexk}{1}
\newcommand{\indexkplus}{2}

\newcommand{\smallsum}{\textstyle\sum}
\newcommand{\SmallSum}[2]{ {\textstyle\sum\limits_{#1}^{#2}}}

\newcommand{\idMatrix}{\operatorname{I} \cfadd{def:identityMatrix}}
\newcommand{\ANNsm}{\mathbf{N} \cfadd{def:ANN}}

\newcommand{\activation}{a}

\newcommand{\functionANN}[1]{\mathcal{R}_{#1} \cfadd{def:ANNrealization}}
\newcommand{\paramANN}{\mathcal{P} \cfadd{def:ANN}}

\newcommand{\lengthANN}{\mathcal{L} \cfadd{def:ANN}}
\newcommand{\inDimANN}{\mathcal{I} \cfadd{def:ANN}}
\newcommand{\compANN}[2]{{#1 \bullet \allowbreak #2} \cfadd{def:ANNcomposition}}

\newcommand{\outDimANN}{\mathcal{O} \cfadd{def:ANN}}
\newcommand{\longerANN}[1]{\mathcal{E}_{#1} \cfadd{def:ANNenlargement}}

\newcommand{\dims}{\mathcal{D} \cfadd{def:ANN}}
\newcommand{\hiddenLength}{\mathcal{H} \cfadd{def:ANN}}

\DeclareMathOperator{\id}{id}

\newcommand{\sumANN}{\mathfrak{S} \cfadd{def:ANN:sum}}
\newcommand{\extensionANN}{\mathfrak{T} \cfadd{def:ANN:extension}}
\newcommand{\dimANNlevel}{\mathbb{D} \cfadd{def:ANN}}
\newcommand{\power}[2]{#1^{\bullet #2} \cfadd{def:iteratedANNcomposition}}
\newcommand{\scalar}[2]{ #1 \circledast #2 \cfadd{def:ANNscalar}}
\newcommand{\transpose}{* \cfadd{def:Transpose}}
\newcommand{\oSum}{\oplus \cfadd{def:ANNsum:same}}
\newcommand{\OSum}[2]{{\mathop\oplus\limits_{#1}^{#2} } \cfadd{def:ANNsum:same} }
\newcommand{\bSum}{{\mathop\boxplus} \cfadd{def:ANN:sum_diff}}
\newcommand{\BSum}[3]{{\mathop\boxplus\limits_{#1,#2}^{#3}} \cfadd{def:ANN:sum_diff}}

\allowdisplaybreaks

\ExplSyntaxOn

\seq_new:N \g_cflist_loaded
\seq_new:N \g_cflist_pending

\NewDocumentCommand{\cfadd} { m } {
	\seq_if_in:NnF \g_cflist_loaded { #1 } {
		\seq_if_in:NnF \g_cflist_pending { #1 } {
			\seq_gput_right:Nn \g_cflist_pending { #1 }
		}
	}
}

\NewDocumentCommand{\cfconsiderloaded} { m } {
	\seq_gput_right:Nn \g_cflist_loaded {#1}
}

\NewDocumentCommand{\cfremove} { m } {
	\seq_gremove_all:Nn \g_cflist_pending { #1 }
}

\NewDocumentCommand{\cfload} { o } {
	\seq_if_empty:NTF \g_cflist_pending {\unskip\IfValueT{#1}{\ignorespaces}} {
		(cf.\ \cref{\seq_use:Nn \g_cflist_pending {,}})\IfValueTF{#1}{#1~}{\unskip}
		\seq_gconcat:NNN \g_cflist_loaded \g_cflist_loaded \g_cflist_pending
		\seq_gclear:N \g_cflist_pending
		\IfValueT{#1}{\ignorespaces}
	}
}

\NewDocumentCommand{\cfclear} {} {
	\seq_gclear:N \g_cflist_loaded
	\seq_gclear:N \g_cflist_pending
}

\NewDocumentCommand{\cfout} { o } {
	\seq_if_empty:NTF \g_cflist_pending {\unskip\IfValueT{#1}{\ignorespaces}} {
		(cf.\ \cref{\seq_use:Nn \g_cflist_pending {,}})\IfValueTF{#1}{#1~}{\unskip}
		\seq_gclear:N \g_cflist_pending
		\IfValueT{#1}{\ignorespaces}
	}
}

\NewDocumentCommand{\ifnocf} { m } {
	\seq_if_empty:NT \g_cflist_pending { #1 }
}

\ExplSyntaxOff

\ExplSyntaxOn

\NewDocumentEnvironment {athm} {m m o} {
	\str_if_eq:noTF {example} {#1} {
		\bool_gset_true:N \g_example_bool
	} {
		\bool_gset_false:N \g_example_bool
	}
	\IfNoValueTF{#3}{
		\begin{#1}\label{#2}\global\def\loc{#2}
		}{
			\begin{#1}[#3]\label{#2}\global\def\loc{#2}
			}
		}{
		\end{#1}
	}
	
	\NewDocumentEnvironment {adef} {m} {
		\begin{definition}\label{#1}\global\def\loc{#1}
		}{
		\end{definition}
	}
	
	\NewDocumentEnvironment{aproof} {} {
		\bool_if:NTF \g_example_bool {
			\begin{proof}[Proof~for~\cref{\loc}]
			} {
				\begin{proof}[Proof~of~\cref{\loc}]
				}
				\bool_gset_false:N \g_finishproof_bool
			}{
				\bool_if:NTF \g_finishproof_bool {}
				{\finishproofthus}
			\end{proof}
		}
		
		\NewDocumentCommand{\finishproofthus} {} {
			\bool_gset_true:N \g_finishproof_bool 
			\bool_if:NTF \g_example_bool {
				The~proof~for~\cref{\loc}~is~thus~complete.
			} {
				The~proof~of~\cref{\loc}~is~thus~complete.
			}
		}
		\NewDocumentCommand{\finishproofthis} {} {
			\bool_gset_true:N \g_finishproof_bool 
			\bool_if:NTF \g_example_bool {
				This~completes~the~proof~for~\cref{\loc}.
			} {
				This~completes~the~proof~of~\cref{\loc}.
			}
		}
		
\ExplSyntaxOff

\ExplSyntaxOn

\bool_new:N \g_noteobserve

\NewDocumentCommand{\setnote}{}{
	\bool_gset_true:N \g_noteobserve
}

\NewDocumentCommand{\setobserve}{}{
	\bool_gset_false:N \g_noteobserve
}

\NewDocumentCommand{\nobs}{ o }{
	\IfValueT{#1}{
		\str_if_eq:noTF {note} {#1} {
			\bool_gset_true:N \g_noteobserve
		} {
			\str_if_eq:noTF {Note} {#1} {
				\bool_gset_true:N \g_noteobserve
			} {
				\bool_gset_false:N \g_noteobserve
			}
		}
	}
	\bool_if:nTF { \g_noteobserve } {
		\bool_gset_false:N \g_noteobserve
		note
	} {
		\bool_gset_true:N \g_noteobserve
		observe
	}
	\IfValueF{#1}{~}
}

\NewDocumentCommand{\Nobs}{ o }{
	\IfValueT{#1}{
		\str_if_eq:noTF {note} {#1} {
			\bool_gset_true:N \g_noteobserve
		} {
			\str_if_eq:noTF {Note} {#1} {
				\bool_gset_true:N \g_noteobserve
			} {
				\bool_gset_false:N \g_noteobserve
			}
		}
	}
	\bool_if:nTF { \g_noteobserve } {
		\bool_gset_false:N \g_noteobserve
		Note
	} {
		\bool_gset_true:N \g_noteobserve
		Observe
	}
	\IfValueF{#1}{~}
}

\ExplSyntaxOff

\ExplSyntaxOn

\int_new:N \g_furthermore

\NewDocumentCommand{\Moreover}{ o o }{
	\IfValueT{#1}{
		\str_case:nn {#1} {
			{Next} {\int_gset:Nn {\g_furthermore} {0}}      
			{Furthermore} {\int_gset:Nn {\g_furthermore} {1}}
			{Moreover} {\int_gset:Nn {\g_furthermore} {2}}
			{In~addition} {\int_gset:Nn {\g_furthermore} {3}}
			{note} {\bool_gset_true:N \g_noteobserve}
			{observe} {\bool_gset_false:N \g_noteobserve}
		}
		\IfValueT{#2}{
			\str_case:nn {#2} {
				{Next} {\int_gset:Nn {\g_furthermore} {0}}        
				{Furthermore} {\int_gset:Nn {\g_furthermore} {1}}
				{Moreover} {\int_gset:Nn {\g_furthermore} {2}}
				{In~addition} {\int_gset:Nn {\g_furthermore} {3}}
				{note} {\bool_gset_true:N \g_noteobserve}
				{observe} {\bool_gset_false:N \g_noteobserve}
			}
		}
	}
	\int_case:nn { \int_mod:nn {\g_furthermore} {4} } {
		{ 0 } { Next~\nobs that}    
		{ 1 } { Furthermore,~\nobs that}
		{ 2 } { Moreover,~\nobs that}
		{ 3 } { In~addition,~\nobs that}
	}
	\int_incr:N \g_furthermore
	\peek_charcode:NTF , {  } { ~ }
}

\ExplSyntaxOff

\ExplSyntaxOn

\bool_new:N \g_hencetherefore

\NewDocumentCommand{\hence}{ o }{
	\IfValueT{#1}{
		\str_if_eq:noTF {hence} {#1} {
			\bool_gset_true:N \g_hencetherefore
		} {
			\str_if_eq:noTF {Hence} {#1} {
				\bool_gset_true:N \g_hencetherefore
			} {
				\bool_gset_false:N \g_hencetherefore
			}
		}
	}
	\bool_if:nTF { \g_hencetherefore } {
		\bool_gset_false:N \g_hencetherefore
		hence
	} {
		\bool_gset_true:N \g_hencetherefore
		therefore
	}
	\IfValueF{#1}{~}
}

\NewDocumentCommand{\Hence}{ o }{
	\IfValueT{#1}{
		\str_if_eq:noTF {hence} {#1} {
			\bool_gset_true:N \g_hencetherefore
		} {
			\str_if_eq:noTF {Hence} {#1} {
				\bool_gset_true:N \g_hencetherefore
			} {
				\bool_gset_false:N \g_hencetherefore
			}
		}
	}
	\bool_if:nTF { \g_hencetherefore } {
		\bool_gset_false:N \g_hencetherefore
		Hence,~we~obtain
	} {
		\bool_gset_true:N \g_hencetherefore
		Therefore,~we~obtain
	}
	\IfValueF{#1}{~}
}

\ExplSyntaxOff

\ExplSyntaxOn

\bool_new:N \g_eg

\NewDocumentCommand{\eg}{ o }{
   \IfValueT{#1}{
     \str_if_eq:noTF {example} {#1} {
       \bool_gset_true:N \g_eg
     } {
       \bool_gset_false:N \g_eg
     }
   }
   \bool_if:nTF { \g_eg } {
     \bool_gset_false:N \g_eg
     \unskip,~for~example,~
   } {
     \bool_gset_true:N \g_eg
     \unskip,~for~instance,~
   }
}

\ExplSyntaxOff

\ExplSyntaxOn

\seq_const_from_clist:Nn \g_prove_mru {
  establish,
  demonstrate,
  prove,
  show,
  imply,
  ensure
}

\prop_new:N \l__verbs
\prop_put:Nnn \l__verbs {show} {shows}
\prop_put:Nnn \l__verbs {imply} {implies}
\prop_put:Nnn \l__verbs {demonstrate} {demonstrates}
\prop_put:Nnn \l__verbs {prove} {proves}
\prop_put:Nnn \l__verbs {establish} {establishes}
\prop_put:Nnn \l__verbs {ensure} {ensures}
\prop_put:Nnn \l__verbs {assure} {assures}

\tl_new:N \g_wordtmp
\seq_new:N \l_mytmps

\cs_generate_variant:Nn \str_if_in:nnTF { nVTF }

\NewDocumentCommand{\prove}{ o }{
  \IfValueTF{#1}{
    \seq_clear:N \l_mytmps
    \seq_map_inline:Nn \g_prove_mru {
      \str_if_eq:nnTF {##1} {ensure} {
        \str_set:Nn \l_temps {n}
      } {
        \str_set:Nx \l_temps {\str_head_ignore_spaces:n {##1}}
      }
      \str_if_in:nVTF {#1} \l_temps {
        \seq_put_right:Nn \l_mytmps {##1}
      } { }
    }
    \seq_get_right:NN \l_mytmps \g_wordtmp
  } {
    \seq_get_right:NN \g_prove_mru \g_wordtmp
  }
  \tl_use:N \g_wordtmp
  \IfValueTF{#1}{}{~}
  \seq_gput_left:NV \g_prove_mru \g_wordtmp
  \seq_gremove_duplicates:N \g_prove_mru
}

\NewDocumentCommand{\proves}{ o }{
  \IfValueTF{#1}{
    \seq_clear:N \l_mytmps
    \seq_map_inline:Nn \g_prove_mru {
      \str_if_eq:nnTF {##1} {ensure} {
        \str_set:Nn \l_temps {n}
      } {
        \str_set:Nx \l_temps {\str_head_ignore_spaces:n {##1}}
      }
      \str_if_in:nVTF {#1} \l_temps {
        \seq_put_right:Nn \l_mytmps {##1}
      } { }
    }
    \seq_get_right:NN \l_mytmps \g_wordtmp
  } {
    \seq_get_right:NN \g_prove_mru \g_wordtmp
  }
  \str_set:NV \l_tmpa_str \g_wordtmp
  \prop_get:NVN \l__verbs \l_tmpa_str \l_tmpa_tl
  \tl_use:N \l_tmpa_tl
  \IfValueTF{#1}{}{~}
  \seq_gput_left:NV \g_prove_mru \g_wordtmp
  \seq_gremove_duplicates:N \g_prove_mru
}

\newcommand{\llabel}[1]{\savelabel{#1}\label{\loc.#1}}

\clist_new:N \l_localreflist
\clist_new:N \l_reflist

\NewDocumentCommand{\lref} { m } {
  \clist_set:No \l_localreflist {#1}
  \clist_clear:N \l_reflist
  \clist_map_inline:Nn \l_localreflist { \clist_put_right:Nn \l_reflist {\loc.##1} }
  \cref{\l_reflist}
}

\NewDocumentCommand{\Lref} { m } {
  \clist_set:No \l_localreflist {#1}
  \clist_clear:N \l_reflist
  \clist_map_inline:Nn \l_localreflist { \clist_put_right:Nn \l_reflist {\loc.##1} }
  \Cref{\l_reflist}
}

\NewDocumentCommand{\itref}{ m m }{
  \clist_set:No \l_localreflist {#2}
  \clist_clear:N \l_reflist
  \clist_map_inline:Nn \l_localreflist { \clist_put_right:Nn \l_reflist {#1.##1} }
  \cref{\l_reflist}~in~\cref{#1}
}

\seq_new:N \l_enum_seq
\int_new:N \l_num_items

\bool_new:N \g_commaused_bool

\providecommand{\comma}{}

\cs_new:Nn \enum_it:nn {
  \int_case:nnF {\l_num_items - #1} {
    {0} {
			\renewcommand{\comma}{}
      #2\space
    }
    {1} {
      \bool_gset_false:N \g_commaused_bool
			\renewcommand{\comma}{,~\bool_gset_true:N \g_commaused_bool}
			#2
			\bool_if:NTF \g_commaused_bool {} {,~}
      and~
    }
  } {
		\bool_gset_false:N \g_commaused_bool
		\renewcommand{\comma}{,~\bool_gset_true:N \g_commaused_bool}
    #2
		\bool_if:NTF \g_commaused_bool {} {,~}
  }
}

\cs_new:Nn \uc:n {\MFUsentencecase{#1}}
\cs_generate_variant:Nn \uc:n {x}

\cs_generate_variant:Nn \str_uppercase:n {x}
\cs_generate_variant:Nn \tl_if_eq:nnTF {onTF}

\cs_new:Nn \enum:nnnn {
  \seq_set_split:Nnn \l_enum_seq ; {#1}
  \seq_remove_all:Nn \l_enum_seq { }
	\seq_remove_all:Nn \l_enum_seq {#2}
  \seq_log:N \l_enum_seq
  \int_set:Nn \l_num_items {\seq_count:N \l_enum_seq}
  \int_log:N \l_num_items
  \int_case:nnF {\l_num_items} {
    { 0 } { 0 }
    { 1 } {
			\IfBooleanTF{#4} {
      	\text_titlecase_first:n {\seq_use:Nn \l_enum_seq {}}
			} {
				\seq_use:Nn \l_enum_seq {}
			}
      \space
			\tl_if_eq:onTF{#3}{-}{}{
				\proves[#3]~
			}
		}
		{ 2 } { 
			\IfBooleanTF{#4} {
      	\text_titlecase_first:n {\seq_use:Nn \l_enum_seq {~and~}}
			} {
				\seq_use:Nn \l_enum_seq {~and~}
			}
			\space
			\tl_if_eq:onTF{#3}{-}{}{
				\prove[#3]~
			}
    }
  } {
    \seq_indexed_map_function:NN \l_enum_seq \enum_it:nn
		\tl_if_eq:onTF{#3}{-}{}{
			\prove[#3]~
		}
  }
}

\cs_generate_variant:Nn \enum:nnnn {nxnn}

\NewDocumentCommand{\enum}{O{} m O{-} s}{
	\IfBooleanTF{#4}{
		\enum:nxnn {#2} {#1} {sindep} \BooleanFalse
	} {
		\enum:nxnn {#2} {#1} {#3} \BooleanFalse
	}
}

\NewDocumentCommand{\dott}{}{\ifnocf{.}\space}

\seq_new:N \l_arg_seq
\tl_new:N \l_dummy_tl

\bool_new:N \g_arg_start_bool

\NewDocumentCommand{\startnewargseq}{}{\bool_gset_true:N \g_arg_start_bool}

\cs_generate_variant:Nn \seq_if_in:NnTF {NxTF}
\cs_generate_variant:Nn \seq_remove_all:Nn {Nx}

\int_new:N \l_random_int

\cs_generate_variant:Nn \tl_if_head_eq_catcode:nNTF {oNTF}

\bool_new:N \g_debug_bool
\bool_gset_true:N \g_debug_bool

\NewDocumentCommand{\argument}{mom}{
	\seq_set_split:Nnn \l_arg_seq ; {#1}
  \seq_remove_all:Nn \l_arg_seq { }
	\seq_log:N \l_arg_seq
	\seq_if_in:NxTF \l_arg_seq {\lref{\g_label_tl}} {
		\seq_remove_all:Nx \l_arg_seq {\lref{\g_label_tl}}
		\seq_get_left:NNTF \l_arg_seq \l_dummy_tl {
			\tl_log:N \l_dummy_tl
			\tl_if_head_eq_catcode:oNTF {\l_dummy_tl} a {
				\int_case:nnF {\seq_count:N \l_arg_seq} {
					{1} {
						\int_case:nn {\int_rand:nn {0} {2}} {
							{0} {
								\enum:nxnn {#1} {\lref{\g_label_tl}} {-} {\BooleanTrue}
								\hence~
								\proves[sindep]~\ignorespaces #3
							}
							{1} {
								\enum[\lref{\g_label_tl}]{
									This;
									#1
								}*\ignorespaces #3
							}
							{2} {
								Combining~
								\enum[\lref{\g_label_tl}]{
									this;
									#1
								} \proves[sindep]~\ignorespaces #3
							}
						}
					}
				} {
					\int_case:nn {\int_rand:nn {0} {3}} {
					 	{0} {
							\enum:nxnn {#1} {\lref{\g_label_tl}} {-} {\BooleanTrue}
							\hence~
							\prove[sindep]~\ignorespaces #3
						}
						{1} {
							\enum[\lref{\g_label_tl}]{
								This;
								#1
							}*\ignorespaces #3
						}
						{2} {
							Combining~
							\enum[\lref{\g_label_tl}]{
								this;
								#1
							} \proves[sindep]~\ignorespaces #3
						}
						{3} {
							Combining~
							\enum:nxnn {#1} {\lref{\g_label_tl}} {-} {\BooleanFalse}
							\hence~
							\proves[sindep]~\ignorespaces #3
						}
					}
				}
			} {
				\int_case:nnF {\seq_count:N \l_arg_seq} {
					{1} {
						\enum[\lref{\g_label_tl}]{
							This;
							#1
						}*\ignorespaces #3
					}
				} {
					\int_case:nn {\int_rand:nn {0} {1}} {
						{0} {
							\enum[\lref{\g_label_tl}]{
								This;
								#1
							}*\ignorespaces #3		
						}
						{1} {
							Combining~
							\enum[\lref{\g_label_tl}]{
								this;
								#1
							} \proves[sindep]~\ignorespaces #3		
						}
					}
				}
			}
		} {
			\tl_if_head_eq_catcode:oNTF {#3} a {
				\int_case:nn {\int_rand:nn {0} {1}} {
					{0} {
						Hence,~we~obtain~\ignorespaces #3
					}
					{1} {
						This~\proves[sindep]~\ignorespaces #3
					}
				}
			} {
				This~\proves[sindep]~\ignorespaces #3
			}
		} 
	} {
		\int_compare:nNnTF {\seq_count:N \l_arg_seq} = {0} {
			\bool_if:NTF \g_arg_start_bool {
				\Nobs\unskip
				#3
			} {
				\Moreover~
				#3
			}
		} {
			\bool_if:NTF \g_arg_start_bool {
				\Nobs~that~
				\enum{
					#1
				}*\ignorespaces #3
			} {
				\int_compare:nNnTF {\seq_count:N \l_arg_seq} = {1} {
					\int_set:Nn \l_tmpa_int {4}
				} {
					\int_set:Nn \l_tmpa_int {6}
				}
				\int_case:nn {\int_rand:nn {0} {\l_tmpa_int}} {
					{0} {
						Moreover,~\nobs~that~
						\enum{
							#1
						}*\ignorespaces #3		
					}
					{1} {
						Furthermore,~\nobs~that~
						\enum{
							#1
						}*\ignorespaces #3		
					}
					{2} {
						In~addition,~\nobs~that~
						\enum{
							#1
						}*\ignorespaces #3		
					}
					{3} {
						Next,~\nobs~that~
						\enum{
							#1
						}*\ignorespaces #3		
					}
					{4} {
						In~the~next~step~we~\nobs~that~
						\enum{
							#1
						}*\ignorespaces #3		
					}
					{5} {
						Next~we~combine~
						\enum{
							#1
						}to~obtain~\ignorespaces #3
					}
					{6} {
						In~the~next~step~we~combine~
						\enum{
							#1
						}to~obtain~\ignorespaces #3
					}
				}
			}
		}
	}
	\bool_gset_false:N \g_arg_start_bool
	\cfload[.]\ignorespaces
}

\tl_new:N \g_label_tl
\tl_gset:Nn \g_label_tl { }

\NewDocumentCommand{\savelabel}{m}{\tl_gset:Nn \g_label_tl {#1}}

\ExplSyntaxOff

\input{abbreviations.tex}

\begin{document}

\title{Deep neural networks with ReLU, leaky ReLU, \\
and softplus activation
provably overcome the\\
curse 
of dimensionality 
for space-time solutions \\
of semilinear partial differential equations}

\author{
Julia Ackermann$^{1}$,
Arnulf Jentzen$^{2,3}$,\\ 
\vspace{-0.3cm}
Benno Kuckuck$^4$,
and
Joshua Lee Padgett$^{5}$
\bigskip
\\
\small{$^1$ Department of Mathematics \& Informatics,}
\vspace{-0.1cm}\\
\small{University of Wuppertal, Germany, e-mail: \texttt{jackermann@uni-wuppertal.de}}
\smallskip
\\
\small{$^2$ School of Data Science and Shenzhen Research Institute of Big Data,}
\vspace{-0.1cm}\\
\small{The Chinese University of Hong Kong, Shenzhen (CUHK-Shenzhen), \vspace{-0.1cm}\\China, e-mail: \texttt{ajentzen@cuhk.edu.cn}}
\smallskip
\\
\small{$^3$ Applied Mathematics: Institute for Analysis and Numerics,}
\vspace{-0.1cm}\\
\small{University of M{\"u}nster, Germany, e-mail: \texttt{ajentzen@uni-muenster.de}}
\smallskip
\\
\small{$^4$ Applied Mathematics: Institute for Analysis and Numerics,}
\vspace{-0.1cm}\\
\small{University of M{\"u}nster, Germany, e-mail: \texttt{bkuckuck@uni-muenster.de}}
\smallskip
\\
\small{$^5$ Data \& Analytics, Toyota Financial Services,}
\vspace{-0.1cm}\\
\small{Texas, USA, e-mail: \texttt{josh.padgett@toyota.com}}
}

\date{\today}

\maketitle

\begin{abstract}
    It is a very challenging topic in applied mathematics to solve high-dimensional nonlinear 
    partial differential equations (PDEs). 
    Standard approximation methods for nonlinear PDEs such as finite difference and finite element methods suffer under the so-called 
    curse of dimensionality (COD) 
    in the sense that the number of computational operations of the numerical approximation method grows at least exponentially in the PDE dimension 
    and with such methods it is essentially impossible to approximately solve high-dimensional PDEs even when the fastest currently available computers are used. 
    However, in the last years great progress has been made in this area of research through suitable 
    deep learning (DL) 
    based methods for PDEs in which deep neural networks (DNNs) 
    are used to approximate solutions of PDEs. 
    Despite the remarkable success of such DL methods in numerical simulations, it remains a fundamental open problem of research to prove (or disprove) that such methods can overcome the COD in the approximation of PDEs. 
    However, there are nowadays several partial error analysis results for DL methods for high-dimensional nonlinear PDEs in the literature which prove that DNNs can overcome the COD in the sense that the number of parameters of the approximating DNN grows at most polynomially in both the reciprocal $ \nicefrac{ 1 }{ \varepsilon } $ of the prescribed approximation accuracy $ \varepsilon > 0 $
    and the PDE dimension $ d \in \N = \{ 1, 2, 3, \dots \}$.
    In the main result of this article we prove that for all $T, p \in(0,\infty)$ it holds that solutions $u_d\colon [0,T]\times \R^d \to \R$, $d\in\N$, of semilinear heat equations with Lipschitz continuous nonlinearities 
    can be approximated in the $L^p$-sense  
    on space-time regions without the COD by DNNs with the 
    rectified linear unit (ReLU), 
    the leaky ReLU, or the softplus activation function. 
    In previous articles similar results have been established not for space-time regions but for the solutions $u_d( T, \cdot )$, $d \in \N$, at the terminal time $T$. 
\end{abstract}

\tableofcontents

%
%
%


\section{Introduction}
\label{sec:intro}

It is a very challenging topic in applied mathematics to solve high-dimensional \PDEs. 
The dimensionality corresponds here to the number of dimensions/degrees of freedom of the domain set on which solutions of the \PDE\ are defined. Classical deterministic numerical approximation methods for \PDEs\ such as finite difference methods (see, e.g., Jovanovi\'{c} \& S\"{u}li \cite{Sueli2014book}) typically suffer from the so-called 
\COD\ 
(cf., e.g., Bellman \cite{bellman2013dynamic}, Novak \& Ritter \cite{NovakRitter1997}, and Novak \& Wo{\'z}niakowski \cite[Chapter~1]{Novak2008}) in the sense that the number of computational operations of the numerical method grows at least exponentially in the \PDE\ dimension $ d \in \N = \{ 1, 2, 3, \dots \} $ and with such numerical methods it is basically impossible to approximately compute solutions of even moderate high-dimensional \PDEs, say, 30-dimensional \PDE\ solutions (corresponding to $ d = 30 $).

Great progress has been made in this field of research using suitable 
\DL\ 
based approximation methods for high-dimensional \PDEs. 
More specifically, in about the last 7 years there have arisen a large number of articles in which 
suitable \DL\ based approximation methods 
-- involving deep \ANNs\  
trained by stochastic gradient descent optimization methods --
have been proposed and used to approximately solve high-dimensional \PDEs. 
For example, 
we refer to \cite{weinan2017deep,Han2018PNAS,BeckJentzenE2019,ChanMikaelWarin2019,Kolmogorov,HurePhamWarin2019,PhamWarin2019,BeckBeckerCheridito2019,NueskenRichter2020,raissi2018forward,kremsner2020}
for \DL\ methods which are based on stochastic representations (involving forward \SDEs\  
or forward \BSDEs)  
of the \PDE\ under consideration such as deep \BSDE\ and deep Kolmogorov methods, 
we refer to \cite{RaissiEtAl2019PINN,sirignano2017dgm,hu2023tackling,lu2021DeepXDE,Berg2018AUD,gu2021selectnet}
for \DL\ methods which are based on the classical or strong formulation of the \PDE\ under consideration such as \PINN\ 
and \DG\ 
methods, 
and we refer to 
\cite{weinan2018deep,ValsecchiOliva2022xnodewan,ZangBaoYeZhou2020weak,chen2023friedrichs,BaoYeZangZhou2020inverseWAN}
for \DL\ methods which are based on weak or variational formulations of the \PDE\ under consideration. 
We also refer, for instance, to the survey articles Beck et al.~\cite{BeckHutzenthalerJentzenKuckuck2023anOverview}, Blechschmidt \& Ernst~\cite{BlechschmidtErnst2021}, Cuomo et al.~\cite{CuomoEtAl2022}, E et al.~\cite{EHanJentzen2021algorithms}, Germain et al.~\cite{germain2021neural}, and Karniadakis et al.~\cite{karniadakis2021physicsinformedML} and the monograph Jentzen et al.~\cite[Chapters~16--18]{jentzen2023mathematical} 
for further references and details.

Despite the remarkable success of such \DL\ methods in numerical simulations, it remains a fundamental open problem of research to prove (or disprove) that such methods can indeed overcome the \COD\ in the approximation of \PDEs. 
Actually, even in the situation of one-dimensional \PDEs\ and one-dimensional abstract target functions it remains a challenging open research problem to prove (or disprove) that such methods do indeed converge (cf., e.g., 
\cite{Gentile2022master,welper2023approximation,gentile2022approximation,ibragimov2022convergence,gonon2023random,jentzenriekert2024nonconvergence,cheridito2021nonconv,lu2020dying,shin2020trainability}).

However, there are nowadays several partial error analyses 
for \DL\ methods for high-dimensional \PDEs\ in the scientific literature which prove that \ANNs\ have the fundamental capacity to overcome the \COD\ in the sense that the number of parameters of the approximating \ANN\ grows at most polynomially in both the reciprocal $ \nicefrac{ 1 }{ \varepsilon } $ of the prescribed approximation accuracy $ \varepsilon > 0 $ and the dimension $ d \in \N $ of the PDE\footnote{This polynomial growth property in both the inverse $\nicefrac{1}{\varepsilon}$ of the prescribed approximation accuracy $ \varepsilon > 0 $ and the \PDE\ dimension $ d \in \N $ is sometimes referred to as polynomial tractability in the literature (cf., e.g., Novak \& Wo{\'z}niakowski \cite[Section~4.4.1]{Novak2008}).}. 
Such \ANN\ approximation results for high-dimensional \PDEs\ have first been obtained for linear \PDEs\ of the Kolmogorov type 
(see, e.g., \cite{BernerGrohsJentzen2018,
GononSchwab2020,
GononSchwab2023,
ElbraechterSchwab2018,
GrohsWurstemberger2018,
JentzenSalimovaWelti2018,
ReisingerZhang2019,
GrohsJentzenSalimova2019,
GononGrohsEtAl2019,
HornungJentzenSalimova2020,
BaggenstosSalimova2023,cheridito2023efficient,
xiao2023empirical}) and, thereafter, have been extended to certain classes of nonlinear \PDEs\ 
(see, e.g., \cite{ackermann2023deep,HutzenthalerJentzenKruse2019,cioicalicht2022deep,
neufeld2023rectifiedVlasov,neufeld2024rectifiedGradientDep,neufeld2023pide}). 
We also refer to the survey articles Beck et al.~\cite[Section~6]{BeckHutzenthalerJentzenKuckuck2023anOverview} 
and E et al.~\cite[Section~7]{EHanJentzen2021algorithms} 
and the monograph Jentzen et al.~\cite[Section~18.4]{jentzen2023mathematical}
for further reading on such \ANN\ approximation results.

In this work we prove in \cref{thm_intro} in this introductory section that 
for every arbitrarily large moment $p \in (0,\infty)$ and every arbitrarily large time horizon $T \in (0,\infty)$ 
it holds that deep \ANNs\ with the 
\ReLU, 
the leaky \ReLU,  or the softplus activation 
overcome the \COD\ in the 
$L^p([0,T]\times [0,1]^d;\R)$-approximation of solutions of a class of semilinear heat \PDEs\ with Lipschitz continuous nonlinearities 
(see \cref{eq:thm_intro_Lpconv} in \cref{thm_intro} for details). 
\cref{thm_intro} follows from the more general results in \cref{theorem:main}, \cref{cor_of_mainthm1}, and \cref{cor_of_mainthm2} in \cref{sec:ANN_approx_PDE} 
and in \cref{thm_intro} in this introductory section we restrict ourselves to measuring the error with respect to the Lebesgue integral on the simple space-time region 
$[0,T]\times [0,1]^d$ 
while in our more general results in \cref{theorem:main}, \cref{cor_of_mainthm1}, and \cref{cor_of_mainthm2} 
we consider more general measures on more general space-time regions to measure the error between the exact solution of the \PDE\ and its deep \ANN\ approximation. 
In our preliminary article \cite{ackermann2023deep} we also showed such an \ANN\ approximation result for semilinear heat \PDEs\ but we restricted ourselves to deep \ANN\ approximations for the \PDE\ solution on some spatial regions (subsets of $\R^d$) evaluated at the terminal time $T$ 
instead of on space-time regions as in this work. 
We now present the precise statement of \cref{thm_intro} in a self-contained fashion in full mathematical details and, thereafter, we provide further explanatory sentences regarding the statement of \cref{thm_intro}.


\begin{samepage}
\begin{athm}{theorem}{thm_intro}
Let 
$T, \constantAssumpMainThm, p \in (0,\infty)$, 
let
$f\colon \R \to \R$ be Lipschitz continuous, 
for every $d \in \N$ let $u_d \in C^{1,2}([0,T]\times \R^d,\R)$
satisfy for all $t\in[0,T]$, $x \in \R^d$ that 
\begin{equation}\label{eq:thm_intro_heateq}
\tfrac{\partial}{\partial t} u_d(t,x)
= 
\Delta_x u_d(t,x)
+
f\pr{ u_d(t,x) } , 
\end{equation}
let $\indexAct \in\{0,1\}$, $\leaky \in \R\backslash\{-1,1\}$, 
let $a\colon \R\to\R$ satisfy for all $x \in \R$ that $a(x) = \indexAct \max\{x,\leaky x\} + (1-\indexAct)  \ln(1+\exp(x))$,   
for every $d\in\N$, $x=(x_1,\dots,x_d)\in\R^d$
let $\A(x)\in\R^d$ satisfy 
\begin{equation}
    \A(x)=(a(x_1),\dots,a(x_d)), 
\end{equation}
for every $L\in\N$, $l_0,l_1,\dots,l_L \in \N$, 
$\Phi=( (W_1,B_1),\dots,(W_L,B_L) ) \in \pr{\times_{k=1}^L (\R^{l_{k}\times l_{k-1}}\times \R^{l_k})}$  
let $\functionANN{}(\Phi)\colon \R^{l_0} \to \R^{l_L}$
and $\paramANN(\Phi) \in \N$ satisfy for all 
$v_0 \in \R^{l_0}$, $v_1 \in \R^{l_1}$, $\dots$, $v_{L} \in \R^{l_{L}}$ 
with $\forall\, k \in \{1,2,\dots,L-1\}\colon v_k = \A(W_kv_{k-1}+B_k)$ that 
\begin{equation}\label{eq:thm_intro_realization_params}
    \textstyle{
    (\functionANN{}(\Phi))(v_0) 
    = W_L v_{L-1} + B_L
    \qquad \text{and} \qquad 
    \paramANN(\Phi) = \sum_{k=1}^L l_k (l_{k-1}+1),
    }
\end{equation} 
for every $d\in\N$ let 
\begin{equation}\label{eq:thm_intro_ANNs}
  \textstyle{
  \mathbf{N}_d = \cup_{ H \in \N } \cup_{ (l_0, l_1, \dots, l_{ H + 1 } ) \in \{ d \} \times \N^H \times \{ 1 \} } 
  \pr{ \times_{ k = 1 }^{ H + 1 } 
  \pr{
      \R^{ l_k \times l_{k-1} } \times \R^{ l_k } 
    }
  } },
\end{equation} 
and assume for all $d \in \N$, $\varepsilon \in (0,1]$ that there exists 
$\G \in \mathbf{N}_d$ such that for all $t \in [0,T]$, 
$x=(x_1,\dots,x_d) \in \R^d$ 
it holds that 
$\paramANN(\G)\le \constantAssumpMainThm d^{\constantAssumpMainThm} \varepsilon^{-\constantAssumpMainThm}$ and 
\begin{equation}\label{eq:thm_intro_assump_G}
    \textstyle{
    \varepsilon \bpr{ 
    \abs{u_d(t,x)} 
    + \sum_{k=1}^d \abs{
    \tfrac{\partial}{\partial x_k} u_d(0,x)
    }
    }
    + \abs{ u_d(0,x) - (\functionANN{}(\G))(x) }
    \le \varepsilon \constantAssumpMainThm d^{\constantAssumpMainThm} 
    \pr{1+
    \sum_{k=1}^d \abs{x_k}^{\constantAssumpMainThm} } 
    .}
\end{equation}
Then 
there exists 
$c \in \R$ 
such that for all $d \in \N$, $\varepsilon \in (0,1]$ 
there exists $\U \in \mathbf{N}_{d+1}$ 
such that 
\begin{equation}\label{eq:thm_intro_Lpconv}
    \textstyle{
    \bigl[ \int_{[0,T]\times [0,1]^d} \, \abs{u_d(y)-\pr{\functionANN{}(\U)}(y)}^{p} \, \dxx y  \bigr]^{\nicefrac{1}{p}}
    \le \varepsilon 
    \qquad \text{and} \qquad 
    \paramANN(\U) \le c d^c \varepsilon^{-c} 
    .
    }
\end{equation}
\end{athm}
\end{samepage}


\cref{thm_intro} is an immediate consequence of \cref{cor_of_mainthm2} in \cref{subsec:ANN_approx_PDE_specific_activation} below. 
\cref{cor_of_mainthm2}, in turn, follows from \cref{theorem:main}, which is the main theorem of this article. 
In the following we add a few explanatory comments on the conclusion of \cref{thm_intro} and the mathematical objects appearing in \cref{thm_intro}. 

The real number $T > 0$ in \cref{thm_intro} describes the time horizon of the \PDEs\ whose solutions we intend to approximate by \ANNs\ in \cref{thm_intro}. 
The real number $\constantAssumpMainThm > 0$ in \cref{thm_intro} is a constant which we use to formulate the regularity and approximation assumption in \cref{eq:thm_intro_assump_G} in \cref{thm_intro}. 
The real number $p > 0$ in \cref{thm_intro} is a constant which determines 
the way how we measure the error between the \PDE\ solution and its \ANN\ approximation, that is, we measure the error between the \PDE\ solution and its \ANN\ approximation in the $L^p$-distance; see \cref{eq:thm_intro_Lpconv} in \cref{thm_intro}. 

The function $f\colon \R\to\R$ in \cref{thm_intro} is the nonlinearity in the \PDEs\ whose solutions we intend to approximate by \ANNs\ in \cref{thm_intro}. 
It is assumed to be Lipschitz continuous in the sense that there exists $c \in \R$ such that for all $x, y \in \R$ we have that
\begin{equation}
  \abs{ f(x) - f(y) } \leq c \abs{ x - y } .
\end{equation}
In \cref{eq:thm_intro_heateq} in \cref{thm_intro} we present the semilinear heat \PDEs\ whose solutions we intend to approximate by \ANNs\ in \cref{thm_intro}
and 
the functions $u_d\colon [0,T]\times \R^d \to \R$, $d\in\N$,  in \cref{thm_intro} are the solutions of the \PDEs\ in \cref{eq:thm_intro_heateq}. 

The function $a\colon \R\to\R$ in \cref{thm_intro} is the activation function for the approximating \ANNs\ in \cref{thm_intro}. 
The real numbers $\indexAct,\leaky \in \R$ are two parameters that determine the concrete choice of the activation function $a\colon \R\to\R$. 
In particular, in the case $\indexAct=0$ we have that $a$ is nothing else but the \emph{softplus activation} (see, e.g., \cite[Section~1.2.5]{jentzen2023mathematical}), in the case $\indexAct = \leaky = 1$ we have that $a$ is nothing else but the \emph{\ReLU\ activation} (see, e.g., \cite[Section~1.2.3]{jentzen2023mathematical}), and in the case $\indexAct = 1, \leaky \in (0,1)$ we have that $a$ is nothing else but the \emph{leaky \ReLU\ activation} with leaky factor $\leaky$ (see, e.g., \cite[Section~1.2.11]{jentzen2023mathematical}). 
We also note that in \cref{thm_intro} we have for every 
$d\in\N$ that the function 
$\R^d \ni x \mapsto \A(x) \in \R^d$ is the $d$-dimensional version of the one-dimensional activation function $a\colon \R\to\R$. 

The sets $\ANNsm_d$, $d\in\N$, in \cref{eq:thm_intro_ANNs} describe the sets of the approximating \ANNs\ in \cref{thm_intro}. 
Moreover, we note that for all $d\in \N$ and every $\Phi \in \ANNsm_d$ we have that the function $\functionANN{}(\Phi) \colon \R^d \to \R$ in \cref{thm_intro} is the realization function associated to the \ANN\ $\Phi$. 
Furthermore, we observe that for all $d\in\N$ and every $\Phi \in \ANNsm_d$ we have that the natural number $\paramANN(\Phi)$ specifies the number of scalar real parameters used to describe the \ANN\ $\Phi$. 
In particular, we note that for all $d\in\N$ and every $\Phi \in \ANNsm_d$ we have that $\paramANN(\Phi)$ is connected to the amount of memory (the amount of bits) needed to store $\Phi$ on a computer. 

In \cref{thm_intro} we also impose the assumption that the solutions $u_d\colon [0,T]\times \R^d \to \R$, $d\in\N$, of the \PDEs\ in \cref{eq:thm_intro_heateq} grow at most polynomially in the \PDE\ dimension and the spatial variable. 
This growth assumption is the subject of the regularity and approximation assumption in~\cref{eq:thm_intro_assump_G} in \cref{thm_intro}. 
More formally, observe that~\cref{eq:thm_intro_assump_G} implies that for all $d \in \N$, $t \in [0,T]$, 
$x = (x_1,\dots,x_d) \in \R^d$ 
we have that
\begin{equation}
  \textstyle{\abs{ u_d( t, x ) } \leq \constantAssumpMainThm d^{ \constantAssumpMainThm } ( 1+\sum_{k=1}^d \abs{x_k}^\constantAssumpMainThm )} .
\end{equation}
The assumption in~\cref{eq:thm_intro_assump_G} also ensures that the gradients of the initial values of the \PDE\ solutions $u_d\colon [0,T]\times \R^d \to \R$, $d\in\N$, grow at most polynomially in the \PDE\ dimension and the spatial variable. 
More formally, we note that~\cref{eq:thm_intro_assump_G} implies that for all $d \in \N$, 
$x = (x_1,\dots,x_d) \in \R^d$ 
we have that
\begin{equation}
  \textstyle{
  \sum_{k=1}^d \abs{
    \tfrac{\partial}{\partial x_k} u_d(0,x)
    }
    \leq \constantAssumpMainThm d^{ \constantAssumpMainThm } ( 1+\sum_{k=1}^d \abs{x_k}^\constantAssumpMainThm )} .
\end{equation}
In addition, in \cref{thm_intro} we also assume that the initial value functions $\R^d \ni x \mapsto u_d( 0, x ) \in \R$, $d\in\N$, of the \PDE\ solutions $u_d\colon [0,T]\times \R^d \to \R$, $d\in\N$, can be approximated by \ANNs\ without the \COD\ in the sense of~\cref{eq:thm_intro_assump_G}. 
More specifically, we observe that~\cref{eq:thm_intro_assump_G} ensures that for every arbitrarily large \PDE\ dimension $d \in \N$ and every arbitrarily small prescribed approximation accuracy $\varepsilon \in (0,1]$ we have that there exists an \ANN\ $\G \in \ANNsm_d$ 
such that for all 
$ x = ( x_1,\dots,x_d ) \in \R^d$ 
we have that the approximation error 
\begin{equation}
    \abs{ u_d( 0, x ) - ( \functionANN{}(\G) )( x ) }
\end{equation}
between the initial value function $u_d(0,\cdot)$ evaluated at~$x$ and the realization $\functionANN{}(\G)$ of the ANN~$\G$ evaluated at~$x$ is bounded by 
$\varepsilon  \constantAssumpMainThm d^{ \constantAssumpMainThm } ( 1+\sum_{k=1}^d \abs{x_k}^\constantAssumpMainThm )$ 
and such that the number of parameters~$\paramANN(\G)$ of the approximating ANN~$\G$ is bounded by 
$\constantAssumpMainThm d^\constantAssumpMainThm \varepsilon^{-\constantAssumpMainThm}$.

In the above described setup \cref{thm_intro} concludes in~\cref{eq:thm_intro_Lpconv} that there exists a constant $c \in \R$ which is independent of the \PDE\ dimension and the approximation accuracy such that for every arbitrarily large \PDE\ dimension $ d \in \N $ and every arbitrarily small prescribed approximation accuracy $\varepsilon \in (0,1]$ we have that there must exist an \ANN\ $\U \in \ANNsm_{d+1}$ 
such that the $L^p$-approximation error 
\begin{equation}
    \textstyle{
    [\int_{[0,T]\times [0,1]^d} \, \abs{u_d(y)-\pr{\functionANN{}(\U)}(y)}^{p} \, \dxx y ]^{ 1 / p }
    }
\end{equation} 
is smaller than or equal to the prescribed approximation accuracy $\varepsilon$ and such that the number of parameters $\paramANN(\U)$ of the approximating \ANN\ $\U$ (connected to the amount of memory to store $\U$) is bounded by $c d^c \eps^{-c}$ and thus grows at most polynomially, in both, the reciprocal~$\nicefrac{1}{\varepsilon}$ of the prescribed approximation accuracy~$\varepsilon > 0$ and the \PDE\ dimension $d \in \N$.

The arguments in our proof of \cref{thm_intro} are based on so-called 
\MLP\ 
approximation methods 
(see Hutzenthaler et al.~\cite{HutzenthalerJentzenKruse2018})  
and on the \ANN\ representations for \MLP\ methods in our preliminary article \cite{ackermann2023deep}. 
\MLP\ methods are certain nonlinear Monte Carlo methods 
(see Hutzenthaler et al.~\cite{HutzenthalerJentzenKruse2018}  
and E et al.~\cite{EHutzenthaler2021})  
that have been shown to overcome the \COD\ for certain classes of semilinear \PDEs\ 
(see, e.g., \cite{EHutzenthaler2021,HutzenthalerJentzenKruse2018,PadgettJentzen2021,GilesJentzenWelti2019,BeckGononJentzen2020,BeckHornungEtAl2019,HutzenthalerJentzenKruse201912,HutzenthalerJentzenKruseNguyen2020,HutzenthalerPricing2019,HutzenthalerKruse2020,HutzenthalerNguyen2022,neufeld2023multilevelPIDE,neufeld2023multilevelgradientdepCOD}) 
and related problems 
(see, e.g., \cite{HutzenthalerKruseNguyen2022,beck2023nonlinear}). 
In our proof of \cref{thm_intro} we employ that suitable \MLP\ methods provably overcome the \COD\ in the $L^p$-approximation of \PDEs\ of the form \cref{eq:thm_intro_heateq} 
(see our preliminary work Hutzenthaler et al.~\cite{PadgettJentzen2021}) 
and we design suitable deep \ANNs\ that appropriately approximate temporal linear interpolations of such \MLP\ approximations 
(see \cref{ANN_for_MLP2} in \cref{subsec:ANN_approx_MLP} below).

The remainder of this work is structured in the following way. 
In \cref{sec:PropPDEs} we establish appropriate elementary perturbation and regularity estimates for solutions of \PDEs. 
In our proofs of the \ANN\ approximation results for \PDEs\ in this work (such as \cref{thm_intro} above) we employ certain ingredients of a suitable calculus for \ANNs\ from the literature and in \cref{subsec:structured_description} we recall such ingredients of this \ANN\ calculus. 
One of our main goals in \cref{sec:ANN_approx_lin_interpolation_MLP} is to construct and study suitable \ANNs\ (with general/abstract activations) which approximate linear interpolations of appropriate \MLP\ approximations (see \cref{ANN_for_MLP2} in \cref{sec:ANN_approx_lin_interpolation_MLP} for details). 
In \cref{sec:ANN_approx_PDE} we employ some of the findings of 
\cref{sec:PropPDEs,subsec:structured_description,sec:ANN_approx_lin_interpolation_MLP}
to prove the space-time \ANN\ approximation results for semilinear heat \PDEs\ in \cref{theorem:main}, \cref{cor_of_mainthm1}, and \cref{cor_of_mainthm2}. 
\cref{thm_intro} in this introductory section is an immediate consequence of \cref{cor_of_mainthm2}.

%
%
%


\section{Properties of solutions of partial differential equations (PDEs)}
\label{sec:PropPDEs}

The \PDEs\ in the \ANN\ approximation results for \PDEs\ in this work (see \cref{sec:ANN_approx_PDE} and \cref{thm_intro} in the introduction) can be reformulated as suitable 
\SFPEs\  
and the resulting \SFPEs\ can then be solved approximately without the \COD\ by means of certain nonlinear Monte Carlo methods, specifically, by means of \MLP\ methods. 
Our proofs of the \ANN\ approximation results for \PDEs\ in this work (see \cref{sec:ANN_approx_PDE} and \cref{thm_intro} in the introduction) exploit this reformulation of the \PDEs\ as \SFPEs. 
In this section we establish certain elementary perturbation and regularity estimates for solutions of such \SFPEs. 
We employ those perturbation and regularity estimates for solutions of \SFPEs\ and \PDEs, respectively, in the proofs of our \ANN\ approximation results in \cref{sec:ANN_approx_PDE} and \cref{thm_intro}. 

In particular, in \cref{sol_stab2} we provide an elementary upper bound for the absolute value of the difference of two solutions~$u_1$ and~$u_2$ of \SFPEs\ at the same space-time evaluation point but with different (perturbed) nonlinearities~$f_1$ and~$f_2$ and different (perturbed) terminal/initial value functions~$g_1$ and~$g_2$. 
In this aspect we note that \SFPEs\ are usually formulated as terminal value problems and in this aspect we also note the elementary fact that initial value \PDE\ problems can be reformulated as terminal value \PDE\ problems and vice versa (see, for example, \cite[Remark~3.3]{BeckJentzenE2019arxiv}). 
Furthermore, in \cref{temp_u_reg1} we establish elementary temporal $\nicefrac{1}{2}$-Hölder continuity properties for solutions of \SFPEs. 
In particular, \cref{temp_u_reg1} provides an upper bound for the absolute value of the difference of the solution~$u$ of an \SFPE\ evaluated at the same spatial point but at different time points. 
We employ \cref{sol_stab2} and \cref{temp_u_reg1} in our proof of the \ANN\ approximation result in \cref{theorem:main} in \cref{sec:ANN_approx_PDE}.

\subsection{Perturbation estimates for solutions of PDEs}
\label{subsec:stability}

\begin{definition}[Standard and maximum norms]\label{DNN_norm}
	We denote by 
	$\norm{\cdot} \colon \pr{\cup_{d\in\N} \R^d} \to \R$
	and
	$\normmm{\cdot} \colon \pr{\cup_{d\in\N} \R^d} \to \R$ the functions which satisfy for all 
	$d\in\N$, 
	$x = (x_1,\dots,x_d)\in\R^d$ 
	that 
	$\norm{x} = [\sum_{i=1}^d \abs{x_i}^2]^{\nicefrac{1}{2}}$
	and 
	$\normmm{x} = \max_{i\in\{1,2,\dots,d\}} \abs{x_i}$.
\end{definition}

\cfclear
\begin{athm}{lemma}{shifted_time}
    Let $d\in\N$, $T,\LipConstF,\boundFG \in(0,\infty)$, $p\in[1,\infty)$, 
	let $f_1,f_2 \in C([0,T]\times \R^d \times \R,\R)$ satisfy for all $i\in\{1,2\}$, $s,t\in[0,T],$ $x\in\R^d$, $v,w\in\R$ that 
	\begin{equation}\label{shifted_time_eq1}
    \abs{ f_i(t,x,0) } \le \boundFG (1 + \norm{x})^p, \qquad 
	\abs{f_1(t,x,v) - f_1(t,x,w)} \le \LipConstF \abs{v-w},
	\end{equation}
 	\begin{equation}\label{shifted_time_eq2}
    \text{and} \qquad 
	\abs{f_2(s,x,v) - f_2(t,x,w)} \le \LipConstF \bpr{\abs{s-t} + \abs{v-w}},
	\end{equation}
	let $F_i\colon C([0,T]\times \R^d,\R) \to C([0,T]\times \R^d,\R)$, $i\in\{1,2\}$, satisfy for all $i\in\{1,2\}$, $t\in[0,T]$, $x\in\R^d$, $v \in C([0,T]\times \R^d,\R)$ that 
	\begin{equation}\label{shifted_time_eq4}
	\pr{F_i(v)}(t,x) = f_i(t,x,v(t,x)), 
	\end{equation}
    let $g\in C(\R^d,\R)$, 
	let $(\Omega,\cF,\mathbb{P})$ be a probability space, let $\fwpr \colon [0,T] \times \Omega \to \R^d$ be a standard Brownian motion, 
	let $u_1,u_2 \in C([0,T]\times \R^d, \R)$ satisfy for all  $i\in\{1,2\}$, $t\in[0,T]$, $x\in\R^d$ that
	\begin{equation}\label{sol_stab_int_ass}
	\E\biggl[\abs{g(x + \fwpr_{T-t})} + \int_t^T \abs{(F_i(u_i))(s,x+\fwpr_{s-t})} \dx s\biggr] < \infty
	\end{equation}
	\begin{equation}\label{sol_stab_f2}
 \text{and} \qquad
	u_i(t,x) = \E[g(x+\fwpr_{T-t})] + \int_t^T \E[(F_i(u_i))(s,x+\fwpr_{s-t})] \dx s,
	\end{equation}
    and let $\fu_{i,\ft} \colon [0,T-\ft] \times \R^d \to \R$, $\ft\in[0,T]$, $i\in\{1,2\}$, and $\ff_{i,\ft}\colon [0,T-\ft]\times\R^d\times \R \to \R$, $\ft \in [0,T]$, $i\in\{1,2\}$, be the functions which satisfy for all $i\in\{1,2\}$, $\ft \in [0,T]$, $t\in[0,T-\ft]$, $x\in\R^d$, $v\in\R$ that 
    \begin{equation}\label{eq:1015}
        \fu_{i,\ft}(t,x) 
        = u_i(t+\ft,x) \quad \text{ and } \quad \ff_{i,\ft}(t,x,v) = f_i(t+\ft,x,v)
    \end{equation}
	\cfload.
    Then 
    \begin{enumerate}[label=(\roman{*})]
        \item\label{shifted_time_item1}
            it holds for all $i\in\{1,2\}$, $\ft\in[0,T]$ that 
            $\ff_{i,\ft} \in C([0,T-\ft]\times \R^d \times \R,\R)$
            and $\fu_{i,\ft} \in C([0,T-\ft]\times \R^d,\R)$, 
            
        \item\label{shifted_time_item2}
            it holds for all  $i\in\{1,2\}$, $\ft\in[0,T]$, $t \in [0,T-\ft]$, $x\in\R^d$ that 
            \begin{equation}
        	\E\biggl[\abs{g(x + \fwpr_{(T-\ft)-t})} 
            + \int_t^{T-\ft} \abs{\ff_{i,\ft}\bpr{s,x+\fwpr_{s-t},\fu_{i,\ft}(s,x+\fwpr_{s-t})}} \dx s\biggr] < \infty,  
        	\end{equation}
            
        \item\label{shifted_time_item3}
            it holds for all $i\in\{1,2\}$, $\ft\in[0,T]$, $t \in [0,T-\ft]$, $x\in\R^d$ that 
            \begin{equation}
        	\fu_{i,\ft}(t,x) = \E[g(x+\fwpr_{(T-\ft)-t})] + \int_t^{T-\ft} \E\bbr{\ff_{i,\ft}\bpr{s,x+\fwpr_{s-t},\fu_{i,\ft}(s,x+\fwpr_{s-t})}} \dx s,
        	\end{equation}
            and 

            \item\label{shifted_time_itemlast}
            it holds for all $i\in\{1,2\}$, $\ft\in[0,T]$, $s, t \in [0,T-\ft]$, $x\in\R^d$, $v,w \in \R$ that 
            \begin{equation}
            \abs{ \ff_{i,\ft}(t,x,0) } \le \boundFG (1 + \norm{x})^p, 
            \qquad 
            \abs{\ff_{1,\ft}(t,x,v) - \ff_{1,\ft}(t,x,w)} \le \LipConstF \abs{v-w},
            \end{equation}
            \begin{equation}
            \text{and} \qquad 
        	\abs{\ff_{2,\ft}(s,x,v) - \ff_{2,\ft}(t,x,w)} \le \LipConstF \bpr{\abs{s-t} + \abs{v-w}}
            \dott 
        	\end{equation}
    \end{enumerate}
\end{athm}

\cfclear
\begin{aproof}
    \Nobs that the fact that $u_1,u_2 \in C([0,T]\times \R^d,\R)$, the fact that $f_1,f_2 \in C([0,T]\times \R^d \times \R,\R)$, and \cref{eq:1015} 
    establish \cref{shifted_time_item1}. 
    \startnewargseq
    \argument{\cref{shifted_time_eq4}; \cref{sol_stab_int_ass}; \cref{sol_stab_f2}; \cref{eq:1015}; Fubini's theorem; a change of variables}{
    that for all $\ft \in [0,T]$, $t\in[0,T-\ft]$, $x\in\R^d$, $i\in\{1,2\}$ it holds that 
		\begin{equation}
		\begin{split}
		& \infty > 
		\E\biggl[\abs{g(x + \fwpr_{T-(t+\ft)})} + \int_{(t+\ft)}^T \abs{(F_i(u_i))(s,x+\fwpr_{s-(t+\ft)})}\dx s\biggr] \\
		& \quad = \E\biggl[\abs{g(x + \fwpr_{T-t-\ft})} + \int_{t}^{(T-\ft)} \abs{(F_i(u_i))(s+\ft,x+\fwpr_{s-t})}\dx s\biggr] \\
		& \quad = \E\biggl[\abs{g(x + \fwpr_{(T-\ft)-t})} + \int_t^{(T-\ft)} \abs{\ff_{i,\ft}(s,x+\fwpr_{s-t},\fu_{i,\ft}(s,x+\fwpr_{s-t}))}\dx s\biggr] 
		\end{split}
		\end{equation}
		and 
		\begin{equation}
		\begin{split}
		\fu_{i,\ft}(t,x) & = u_i(t + \ft, x) \\
		& = \E[g(x+\fwpr_{T-(t+\ft)})] + \int_{(t+\ft)}^T \E[(F_i(u_i))(s,x+\fwpr_{s-(t+\ft)})]\dx s\\
		& = \E[g(x+\fwpr_{T-(t+\ft)})] + \E\biggl[ \int_{(t+\ft)}^T (F_i(u_i))(s,x+\fwpr_{s-(t+\ft)}) \dx s \biggr]\\
		& = \E[g(x+\fwpr_{(T-\ft)-t})] + \E\biggl[ \int_{t}^{(T-\ft)} \ff_{i,\ft}(s,x+\fwpr_{s-t},\fu_{i,\ft}(s,x+\fwpr_{s-t})) \dx s \biggr] \\
		& = \E[g(x+\fwpr_{(T-\ft)-t})] + \int_{t}^{(T-\ft)} \E[\ff_{i,\ft}(s,x+\fwpr_{s-t},\fu_{i,\ft}(s,x+\fwpr_{s-t}))]\dx s 
		\dott 
		\end{split}
		\end{equation}
  }
  This proves \cref{shifted_time_item2,shifted_time_item3}. 
  Combining \cref{shifted_time_eq1}, \cref{shifted_time_eq2}, and \cref{eq:1015} establishes \cref{shifted_time_itemlast}. 
\end{aproof}

\cfclear
\begin{athm}{corollary}{sol_stab2}
	Let $d\in\N$, $T,\LipConstF,\boundFG,\diffBd \in(0,\infty)$, $p,q\in[1,\infty)$, 
	$f_1,f_2 \in C([0,T]\times \R^d \times \R,\R)$, $g_1,g_2\in C(\R^d,\R)$ satisfy for all $i\in\{1,2\}$, $t\in[0,T],$ $x\in\R^d$, $v,w\in\R$ that 
	\begin{equation}
	\abs{f_i(t,x,v) - f_i(t,x,w)} \le \LipConstF \abs{v-w}, 
    \qquad 
	\max\{\abs{ f_i(t,x,0) },\abs{g_i(x)} \} \le \boundFG (1 + \norm{x})^p,
	\end{equation} 
	\begin{equation}\label{sol_stab2_assump_diff}
 \text{and} \qquad 
	\max\{\abs{f_1(t,x,v) - f_2(t,x,v)},\abs{g_1(x) - g_2(x)}\} \le \diffBd \bpr{ (1 + \norm{x})^{pq} + \abs{v}^q }, 
	\end{equation}
	let $(\Omega,\cF,\mathbb{P})$ be a probability space, let $\fwpr \colon [0,T] \times \Omega \to \R^d$ be a standard Brownian motion, 
	and let $u_1,u_2 \in C([0,T]\times \R^d, \R)$ satisfy for all $i\in\{1,2\}$, $t\in[0,T]$, $x\in\R^d$ that
	\begin{equation}\label{sol_stab2_int_ass}
	\E\biggl[\abs{g_i(x + \fwpr_{T-t})} + \int_t^T \abs{f_i\bpr{s,x+\fwpr_{s-t},u_i(s,x+\fwpr_{s-t})}} \dx s\biggr] < \infty
	\end{equation}
	\begin{equation}\label{sol_stab2_f2}
 \text{and} \qquad 
	u_i(t,x) = \E[g_i(x+\fwpr_{T-t})] + \int_t^T \E\bbr{f_i\bpr{s,x+\fwpr_{s-t},u_i(s,x+\fwpr_{s-t})}} \dx s
	\end{equation}
	\cfload.
	Then it holds for all $t\in[0,T]$, $x\in\R^d$ that
	\begin{equation}
	\abs{ u_1(t,x) - u_2(t,x)} \le 
	\diffBd\bigl(e^{\LipConstF T}(T+1)\bigr)^{q+1}\bigl(\boundFG^q+1\bigr) 3^{pq-1} 
	\bpr{ 1 + \norm{ x }^{pq} + \sup\nolimits_{s\in[0,T]} \E[ \norm{ \fwpr_{s} }^{pq} ] } . 
	\end{equation}
\end{athm}

\cfclear
\begin{aproof}
	Throughout this proof let $\fu_{i,\ft} \colon [0,T-\ft] \times \R^d \to \R$, $\ft\in[0,T]$, $i\in\{1,2\}$, and $\ff_{i,\ft}\colon [0,T-\ft]\times\R^d\times \R \to \R$, $\ft \in [0,T]$, $i\in\{1,2\}$, be the functions which satisfy for all $\ft \in [0,T]$, $t\in[0,T-\ft]$, $x\in\R^d$, $v\in\R$, $i\in\{1,2\}$ that $\fu_{i,\ft}(t,x) = u_i(t+\ft,x)$ and $\ff_{i,\ft}(t,x,v) = f_i(t+\ft,x,v)$.  
 \startnewargseq
    \argument{\cref{sol_stab2_assump_diff}}{
    that for all $\ft\in[0,T]$, $t\in[0,T-\ft]$, $x\in\R^d$, $v\in\R$ it holds that
    \begin{equation}\llabel{sol_stab2_4}
	\abs{\ff_{1,\ft}(t,x,v) - \ff_{2,\ft}(t,x,v)} 
    = \abs{f_1(t+\ft,x,v) - f_2(t+\ft,x,v)} 
    \le \diffBd\bpr{(1 + \norm{x})^{pq} + \abs{v}^q} 
    \dott
	\end{equation}
    }
    \argument{\lref{sol_stab2_4}; \cref{shifted_time}; Hutzenthaler et al.~\cite[Lemma~2.3]{HutzenthalerJentzenKruse2019}
	(applied for every $\ft \in [0,T]$ with $T\with T-\ft$, $L \with \LipConstF$, $B\with \boundFG$, $\delta \with \diffBd$, $u_1 \with \fu_{1,\ft}$, $u_2 \with \fu_{2,\ft}$, $f_1 \with \ff_{1,\ft}$, $f_2 \with \ff_{2,\ft}$, $g_1 \with g_1$, $g_2 \with g_2$, $p\with p$, $q\with q$ in the notation of Hutzenthaler et al.~\cite[Lemma~2.3]{HutzenthalerJentzenKruse2019})}{
    that for all $\ft\in[0,T]$, $t\in[0,T-\ft]$, $x\in\R^d$ it holds that
	\begin{equation}\llabel{sol_stab2_5}
	\E\Bigl[ \bigl\lvert \fu_{1,\ft}(t,x+\fwpr_t) - \fu_{2,\ft}(t,x+\fwpr_t) \bigr\rvert \Bigr] 
	\le \diffBd\bigl(e^{\LipConstF T}(T+1)\bigr)^{q+1}\bigl(\boundFG^q+1\bigr)
	\Bigl( 1 + \lVert x \rVert + \bigl( \E[ \lVert \fwpr_{T-\ft} \rVert^{pq} ] \bigr)^{\frac{1}{pq}} \Bigr)^{pq} 
    \dott
	\end{equation}
    }
    \argument{\lref{sol_stab2_5}; 
	Jensen's inequality}{ 
    that for all $\ft \in [0,T]$, $x\in\R^d$ it holds that
	\begin{equation}
	\begin{split}
	& \abs{u_1(\ft,x) - u_2(\ft,x)} = 
	\lvert \fu_{1,\ft}(0,x) - \fu_{2,\ft}(0,x) \rvert
	= \E\bbr{ \abs{ \fu_{1,\ft}(0,x+\fwpr_0) - \fu_{2,\ft}(0,x+\fwpr_0) } } \\
	& \quad \le \diffBd\bigl(e^{\LipConstF T}(T+1)\bigr)^{q+1}\bigl(\boundFG^q+1\bigr) 3^{pq-1}
	\bpr{ 1 + \lVert x \rVert^{pq} + \sup\nolimits_{s\in[0,T]} \E[ \lVert \fwpr_{s} \rVert^{pq} ] }
	\dott 
	\end{split}
	\end{equation}
    }
\end{aproof}

\subsection{Temporal regularity estimates for solutions of PDEs}
\label{subsec:temporal_regularity}

\cfclear
\begin{athm}{lemma}{g_der_bd}
Let $d\in\N$, $T, \boundFG \in(0,\infty)$, $p\in[1,\infty)$, 
let $g\in C^1(\R^d,\R)$ satisfy for all $x\in\R^d$ that $\norm{\nabla g(x)} \le \boundFG (1 + \norm{x})^p$, 
let $(\Omega,\cF,\mathbb{P})$ be a probability space, and let $\fwpr \colon [0,T] \times \Omega \to \R^d$ be a standard Brownian motion \cfload.
Then it holds for all $t,\ft \in [0,T]$, $x\in\R^d$ that
\begin{equation}
\E\bbr{ \abs{ g(x + \fwpr_{\ft}) - g(x + \fwpr_t) }} 
\le 8^{p+1} \boundFG 
\bbpr{ 1  + \lVert x\rVert^{p} + \sup\nolimits_{s\in[0,T]} \bigl( \E[ \lVert \fwpr_s \rVert^{2p} ] \bigr)^{\frac12} }
\sqrt{\lvert \ft-t\rvert} \sqrt{d}
\dott
\end{equation}
\end{athm}

\cfclear
\begin{aproof}
\Nobs that the fundamental theorem of calculus, the multivariate chain rule, and the Cauchy--Schwarz inequality assure that for all $t,\ft \in [0,T]$, $x\in\R^d$ it holds that 
\begin{equation}\llabel{eq:1747a}
\begin{split}
\lvert g(x+\fwpr_{\ft}) - g(x+\fwpr_t) \rvert 
& \le \int_0^1 \bigl\lVert \nabla g\bigl( (\fwpr_{\ft} - \fwpr_t) z + x + \fwpr_t \bigr) \bigr\rVert 
\lVert \fwpr_{\ft} - \fwpr_t  \rVert \dx z .
\end{split}
\end{equation}
\argument{
the fact that 
for all $x\in\R^d$ it holds that $\lVert \nabla g(x) \rVert \le \boundFG (1+\lVert x\rVert)^p$; 
the triangle inequality;
Jensen's inequality}{
that for all $t,\ft \in [0,T]$, $x\in\R^d$, $z\in[0,1]$ it holds that 
\begin{equation}\llabel{eq:1747b}
\begin{split}
\bigl\lVert \nabla g\bigl( (\fwpr_{\ft} - \fwpr_t) z + x + \fwpr_t \bigr) \bigr\rVert
& \le \boundFG \bigl( 1 + \lVert (\fwpr_{\ft} - \fwpr_t) z + x + \fwpr_t \rVert \bigr)^p \\
& \le  \boundFG \bigl( 1 + \lVert \fwpr_{\ft} \rVert  + \lVert x \rVert + 2 \lVert \fwpr_t \rVert \bigr)^p \\
& \le 4^{p-1} \boundFG \bigl( 1 + \lVert \fwpr_{\ft} \rVert^p  + \lVert x \rVert^p + 2^p \lVert \fwpr_t \rVert^p \bigr) 
\dott
\end{split}
\end{equation}
}
\argument{\lref{eq:1747b};
\lref{eq:1747a}}{
that for all $t,\ft \in [0,T]$, $x\in\R^d$ it holds that 
\begin{equation}\llabel{eq:1747c}
\begin{split}
\lvert g(x+\fwpr_{\ft}) - g(x+\fwpr_t) \rvert
& \le 4^{p-1} \boundFG \bigl( 1 + \lVert \fwpr_{\ft} \rVert^p  + \lVert x \rVert^p + 2^p \lVert \fwpr_t \rVert^p \bigr) 
\lVert \fwpr_{\ft} - \fwpr_t  \rVert  
\dott
\end{split}
\end{equation}
}
\argument{\lref{eq:1747c}; the Cauchy--Schwarz inequality; Jensen's inequality}{
that for all $t,\ft \in [0,T]$, $x\in\R^d$ it holds that 
\begin{equation}\llabel{eq:1747d}
\begin{split}
& \E\bigl[ \lvert g(x+\fwpr_{\ft}) - g(x+\fwpr_t) \rvert \bigr] \\
& \le  4^{p-1} \boundFG \bpr{\E\bbr{\pr{ 1 + \lVert \fwpr_{\ft} \rVert^p  + \lVert x \rVert^p + 2^p \lVert \fwpr_t \rVert^p }^2 }}^{\frac{1}{2}} 
\bigl(\E\bigl[ \lVert \fwpr_{\ft} - \fwpr_t  \rVert^2 \bigr]\bigr)^{\frac{1}{2}} \\
& \le 4^{p} \boundFG \bigl( 1  + \lVert x\rVert^{2p} +  \E\bigl[ \lVert \fwpr_{\ft} \rVert^{2p} \bigr] + 2^{2p} \E\bigl[ \lVert \fwpr_t \rVert^{2p} \bigr]\bigr)^{\frac{1}{2}} 
\bigl(\E\bigl[ \lVert \fwpr_{\lvert \ft - t \rvert} \rVert^2 \bigr]\bigr)^{\frac{1}{2}} \\
& \le 4^{p} \boundFG \bpr{ 1  + \lVert x\rVert^{2p} + (1+2^{2p}) \sup\nolimits_{s\in[0,T]} \E\bigl[ \lVert \fwpr_{s} \rVert^{2p} \bigr] }^{\frac{1}{2}} 
\bigl( \lvert \ft-t\rvert d \bigr)^{\frac{1}{2}} \\
& \le 4^{p} (1+2^p) \boundFG \Bigl( 1  + \lVert x\rVert^{p} + \sup\nolimits_{s\in[0,T]} \bigl( \E\bigl[ \lVert \fwpr_{s} \rVert^{2p} \bigr] \bigr)^{\frac{1}{2}} \Bigr)
\bigl( \lvert \ft-t\rvert d \bigr)^{\frac{1}{2}} 
\dott
\end{split}
\end{equation}
}
\end{aproof}

\cfclear
\begin{athm}{lemma}{lem:BM_power_bound}
Let $T, \constantAssumpMainThm \in(0,\infty)$, $p,r,q,\fq \in[1,\infty)$,
let $(\Omega,\cF,\mathbb{P})$ be a probability space, 
for every $d \in \N$ let $\fwpr^d \colon [0,T] \times \Omega \to \R^d$ be a standard Brownian motion,
and for every $d \in \N$ let $\nu_d \colon \mathcal{B}(\R^{d+1})\to[0,\infty)$ be a measure with 
\begin{equation}\label{eq:BM_power_bound_assump_meas}
    \int_{\R^{d+1}} \pr{1+\norm{y}^{p^2q\fq}}\, \nu_d (\dxx y) \le \constantAssumpMainThm d^{r p^2 q\fq} 
\end{equation}
\cfload. 
Then 
\begin{enumerate}[label=(\roman *)]
\item 
\label{BMpowerbound_item1}
it holds for all $d\in\N$, $\fp \in [1,\infty)$, $s\in[0,T]$ that 
\begin{equation}
\E[\norm{\fwpr^d_s}^{\fp}] \le 1 + (1+2T)^{\fp} \bbpr{ \frac{d}{2} + \fp }^{\fp} 
\end{equation}
and
\item 
\label{BMpowerbound_item2}
there exists $\fC \in [1,\infty)$ such that for all $d \in \N$ it holds that  
\begin{equation}\label{eq:proofmainthm_bound_BM}
    \begin{split}
        & \int_{[0,T]\times \R^d} \bpr{ 1 + \norm{x}^{p^2 q\fq} + \sup\nolimits_{s \in [0,T]} \E\bbr{ \norm{\fwpr^d_s}^{2p^2q\fq} } } \, \nu_d(\dxx t, \dxx x) 
        \le \fC d^{(r+2) p^2 q \fq}  
        \cfout.
    \end{split}
\end{equation}
\end{enumerate}
\end{athm}

\cfclear
\begin{aproof}
\Nobs that the fact that for all $d\in\N$, $s \in (0,T]$ the random variable $\lVert \fwpr^d_s/\sqrt{s} \rVert^2$ is chi-square distributed with $d$ degrees of freedom, Jensen's inequality, and \eg (2.35) in~\cite{simon2007probability},
show that for all $d \in \N$, $k\in\N$, $s\in[0,T]$ it holds that 
\begin{equation}\llabel{eq:1747e}
\begin{split}
\bigl( \E[\lVert \fwpr^d_s\rVert^k ] \bigr)^2 
& \le \E\bigl[ \lVert \fwpr^d_s\rVert^{2k} \bigr]
\le \frac{(2s)^{k} \Gamma(\frac{d}{2}+k)}{\Gamma(\frac{d}{2})}  
= (2s)^k \prod_{j=0}^{k-1} \Bigl( \frac{d}{2} + j \Bigr)
\le (2s)^k  \Bigl( \frac{d}{2} + k-1 \Bigr)^k . 
\end{split}
\end{equation}
This ensures that for all $d \in \N$, $\fp \in [1,\infty)$, $s\in[0,T]$ it holds that 
\begin{equation}\llabel{eq:1747f}
    \begin{split}
    \E[\lVert \fwpr^d_s\rVert^{\fp} ]
    & \le 1 + \E\bigl[ \lVert \fwpr^d_s \rVert^{\lceil p \rceil} \bigr]
    \le 1 + (2s)^{\frac{\lceil \fp \rceil}{2}}  \Bigl( \frac{d}{2} + \lceil \fp \rceil -1 \Bigr)^{\frac{\lceil \fp \rceil}{2}} 
    \le 1 + (1+2T)^{\fp} \Bigl( \frac{d}{2} + \fp \Bigr)^{\fp} .
    \end{split}
\end{equation}
\argument{\lref{eq:1747f}; 
\cref{eq:BM_power_bound_assump_meas}}{
that for all $d \in \N$ it holds that 
\begin{equation}
    \begin{split}
        & \int_{[0,T]\times \R^d} \bpr{ 1 + \norm{x}^{p^2 q\fq} + \sup\nolimits_{s \in [0,T]} \E\bbr{ \norm{\fwpr^d_s}^{2p^2q\fq} } } \, \nu_d(\dxx t, \dxx x) \\
        & \le \constantAssumpMainThm d^{rp^2q\fq}
        +  \constantAssumpMainThm  d^{rp^2q\fq} \bbbpr{ 1 + (1+2T)^{2p^2 q \fq} \bbpr{ \frac{d}{2} + 2 p^2 q\fq }^{2p^2 q \fq} } \\
        & \le 2 \constantAssumpMainThm d^{rp^2q\fq}
        + \constantAssumpMainThm (1+2T)^{2p^2 q \fq} d^{(r+2)p^2q\fq} 
        \bbpr{\frac12 + 2p^2q\fq}^{2p^2 q \fq} \\
        & \le
        \constantAssumpMainThm 
        \bbbpr{2 + (1+2T)^{2p^2 q \fq} 
        \bbpr{\frac12 + 2p^2q\fq}^{2p^2 q \fq} }
        d^{(r+2)p^2 q \fq}
        \dott
    \end{split}
\end{equation}
}
\end{aproof}

\cfclear
\begin{athm}{lemma}{temp_u_reg1a}
Let $d\in\N$, $T,\LipConstF,\boundFG \in(0,\infty)$, $p\in[1,\infty)$, 
let $f \in C([0,T]\times \R^d \times \R,\R)$ and $g\in C^1(\R^d,\R)$ satisfy for all $s,t\in[0,T],$ $x\in\R^d$, $v,w\in\R$ that 
\begin{equation}\label{temp_u_reg1_0a}
    \abs{f(s,x,v) - f(t,x,w)} \le \LipConstF (\abs{s-t} + \abs{v-w})
\end{equation} 
\begin{equation}\label{temp_u_reg1_0b}
\text{and} \qquad 
    \max\{\abs{f(t,x,0)},\lvert g(x)\rvert,\norm{\nabla g(x)} \} \le \boundFG (1 + \norm{x} )^p ,
\end{equation}
let $F\colon C([0,T]\times \R^d,\R) \to C([0,T]\times \R^d,\R)$ satisfy for all $t\in[0,T]$, $x\in\R^d$, $v \in C([0,T]\times \R^d,\R)$ that 
	\begin{equation}\label{temp_u_reg1_0}
	\pr{F(v)}(t,x) = f(t,x,v(t,x)), 
	\end{equation}
let $(\Omega,\cF,\mathbb{P})$ be a probability space, let $\fwpr \colon [0,T] \times \Omega \to \R^d$ be a standard Brownian motion, and let $u \in C([0,T]\times \R^d, \R)$ satisfy for all $t\in[0,T]$, $x\in\R^d$ that
\begin{equation}\label{temp_u_reg1_f0}
    \E\biggl[\abs{g(x + \fwpr_{T-t})} + \int_t^T \abs{(F(u))(s,x+\fwpr_{s-t})} \dx s\biggr] < \infty
\end{equation}
\begin{equation}\label{temp_u_reg1_f1}
\text{and} \qquad 
u(t,x) = \E[g(x+\fwpr_{T-t})] + \int_t^T \E[(F(u))(s,x+\fwpr_{s-t})]\dx s
\end{equation}
\cfload.
Then it holds for all $\ft \in [0,T]$, $t \in [0,\ft]$, $x\in\R^d$ 
that
\begin{equation}
\begin{split}  
& \E\bigl[ \lvert u(\ft,x+\fwpr_t) - u(t,x+\fwpr_t) \rvert \bigr] \\
& \le 
e^{2 \LipConstF T} (T+1)^2 (\LipConstF +1) (\boundFG+1) 8^{p+2} 
\bpr{ 1 + \lVert x \rVert^p + \sup\nolimits_{s\in[0,T]}\E\bigl[ \lVert \fwpr_{s} \rVert^{2p} \bigr] } 
\sqrt{\lvert \ft-t \rvert} \sqrt{d}
\dott
\end{split}
\end{equation}
\end{athm}

\cfclear
\begin{aproof}
\Nobs that \cref{temp_u_reg1_f1} and the triangle inequality show that for all $\delta \in (0,T]$, $x\in\R^d$, $t\in[0,T-\delta]$ it holds that 
\begin{equation}\llabel{temp_u_reg1_p1}
    \begin{split}
    & \lvert u(t+\delta,x) - u(t,x) \rvert \\
    & \le \E\bigl[ \lvert g(x+\fwpr_{T-(t+\delta)}) - g(x+\fwpr_{T-t}) \rvert \bigr] \\
    & \quad + \biggl\lvert \int_{t+\delta}^T \E\bigl[ (F(u))(s,x+\fwpr_{s-(t+\delta)}) \bigr] \dx s  
    - \int_{t}^T \E\bigl[ (F(u))(s,x+\fwpr_{s-t}) \bigr] \dx s
    \biggr\rvert 
    \dott
    \end{split}
\end{equation}
\argument{\cref{temp_u_reg1_f0};Fubini's theorem;a change of variables}{ that for all $\delta \in (0,T]$, $x\in\R^d$, $t\in[0,T-\delta]$ it holds that
\begin{equation}\llabel{temp_u_reg1_p2}
\begin{split}
\int_{t+\delta}^T \E\bigl[ (F(u))(s,x+\fwpr_{s-(t+\delta)}) \bigr] \dx s 
& = \E\biggl[ \int_{t+\delta}^T (F(u))(s,x+\fwpr_{s-(t+\delta)}) \dx s  \biggr] \\
& = \E\biggl[ \int_{t}^{T-\delta} (F(u))(s+\delta,x+\fwpr_{s-t}) \dx s  \biggr] \\
& =  \int_{t}^{T-\delta} \E\bigl[ (F(u))(s+\delta,x+\fwpr_{s-t}) \bigr] \dx s  \dott
\end{split}
\end{equation}
}
\argument{\lref{temp_u_reg1_p2};\cref{temp_u_reg1_0};\cref{temp_u_reg1_f0};the triangle inequality}{ that for all $\delta\in(0,T]$, $x\in\R^d$, $t\in[0,T-\delta]$ it holds that 
\begin{equation}\llabel{temp_u_reg1_p3}
\begin{split}
& \biggl\lvert \int_{t+\delta}^T \E\bigl[ (F(u))(s,x+\fwpr_{s-(t+\delta)}) \bigr] \dx s  
- \int_{t}^T \E\bigl[ (F(u))(s,x+\fwpr_{s-t}) \bigr] \dx s 
    \biggr\rvert \\
& = \bigg\lvert \int_{t}^{T-\delta} \E\big[ (F(u))(s+\delta,x+\fwpr_{s-t}) 
\big] 
\dx s  
- \int_{t}^{T-\delta} \E\bigl[ (F(u))(s,x+\fwpr_{s-t}) \bigr] \dx s \\
& \quad \;
- \int_{T-\delta}^{T} \E\bigl[ (F(u))(s,x+\fwpr_{s-t}) \bigr] \dx s
\bigg\rvert \\
& \le 
\int_t^{T-\delta} \E\bbr{ \babs{ f\bigl(s+\delta,x+\fwpr_{s-t}, u(s+\delta,x+\fwpr_{s-t}) \bigr) -  f\bigl(s,x+\fwpr_{s-t}, u(s,x+\fwpr_{s-t}) \bigr)  } } \dx s  \\
& \quad +  \int_{T-\delta}^{T} \E\bbr{ \babs{ f\bigl(s,x+\fwpr_{s-t}, u(s,x+\fwpr_{s-t}) \bigr) }} \dx s
\dott
\end{split}
\end{equation}
}
\argument{the fact that for all $s,t\in[0,T]$, $x\in\R^d$, $v,w\in\R$ it holds that $\lvert f(s,x,v) - f(t,x,w) \rvert \le \LipConstF (\lvert s-t\rvert + \lvert v-w\rvert)$}{ that for all $\delta\in(0,T]$, $x\in\R^d$, $t\in[0,T-\delta]$ it holds that 
\begin{equation}\llabel{temp_u_reg1_p4}
\begin{split}
& \int_t^{T-\delta} \E\bbr{ \babs{ f\bigl(s+\delta,x+\fwpr_{s-t}, u(s+\delta,x+\fwpr_{s-t}) \bigr) -  f\bigl(s,x+\fwpr_{s-t}, u(s,x+\fwpr_{s-t}) \bigr)  } } \dx s  \\
& \le 
\LipConstF \delta T  + 
\LipConstF \int_t^{T-\delta} \E\bbr{ \babs{  u(s+\delta,x+\fwpr_{s-t}) - u(s,x+\fwpr_{s-t}) }} \dx s 
\dott
\end{split}
\end{equation}
}
\argument{the fact that for all $s,t\in[0,T]$, $x\in\R^d$, $v,w\in\R$ it holds that $\lvert f(s,x,v) - f(t,x,w) \rvert \le \LipConstF (\lvert s-t\rvert + \lvert v-w\rvert)$;the fact that for all $t\in[0,T]$, $x\in\R^d$ it holds that $\lvert f(t,x,0)\rvert \le \boundFG (1+\lVert x\rVert)^p$;the triangle inequality}{ that for all $\delta \in (0,T]$, $x\in\R^d$, $t\in[0,T-\delta]$, $s\in[T-\delta,T]$ it holds that 
\begin{equation}\llabel{temp_u_reg1_p5}
\begin{split}
& \E\bbr{ \babs{ f\bigl( s,x+\fwpr_{s-t}, u(s,x+\fwpr_{s-t}) \bigr) }} \\
& \le \E\bbr{ \babs{ f\bigl( s,x+\fwpr_{s-t}, u(s,x+\fwpr_{s-t}) \bigr) -  f\bigl( s,x+\fwpr_{s-t}, 0 \bigr) }} 
+ \E\bbr{ \babs{ f\bigl( s,x+\fwpr_{s-t}, 0 \bigr) }} \\
& \le \LipConstF \E\bbr{ \babs{ \lvert u(s,x+\fwpr_{s-t}) }} 
+ \boundFG \E\bbr{ (1+\lVert x + \fwpr_{s-t} \rVert)^p } \dott
\end{split}
\end{equation}
}
\argument{\lref{temp_u_reg1_p5};\lref{temp_u_reg1_p1};\lref{temp_u_reg1_p3};\lref{temp_u_reg1_p4}}{ that for all $\delta \in (0,T]$, $x\in\R^d$, $t\in[0,T-\delta]$ it holds that 
\begin{equation}\llabel{temp_u_reg1_p6}
\begin{split}
& \lvert u(t+\delta,x) - u(t,x) \rvert \\
& \le \E\bigl[ \lvert g(x+\fwpr_{T-(t+\delta)}) - g(x+\fwpr_{T-t}) \rvert \bigr] 
+ \LipConstF \delta T \\
& \quad + \LipConstF \int_t^{T-\delta} \E\bigl[ \lvert u(s+\delta,x+\fwpr_{s-t}) - u(s,x+\fwpr_{s-t}) \rvert \bigr] \dx s  \\
& \quad + \LipConstF \int_{T-\delta}^T \E[ \lvert u(s,x+\fwpr_{s-t}) \rvert ] \dx s
+ \boundFG \int_{T-\delta}^T \E[(1+\lVert x + \fwpr_{s-t} \rVert)^p] \dx s 
\dott
\end{split}
\end{equation}
}
\argument{\lref{temp_u_reg1_p6};Fubini's theorem;the fact that $\fwpr$ has independent and stationary increments}{ that for all $\delta \in (0,T]$, $x\in\R^d$, $t\in[0,T-\delta]$ it holds that  
\begin{equation}\llabel{temp_u_reg1_p7}
\begin{split}
& \E\bigl[ \lvert u(t+\delta,x+\fwpr_{t}) - u(t,x+\fwpr_{t}) \rvert \bigr] \\
& \le 
\E\bigl[ \lvert g(x+\fwpr_{T-\delta}) - g(x+\fwpr_{T}) \rvert \bigr] 
+ \LipConstF \delta T \\
& \quad + \LipConstF \int_t^{T-\delta} \E\bigl[ \lvert u(s+\delta,x+\fwpr_{s}) - u(s,x+\fwpr_{s}) \rvert \bigr] \dx s  \\
& \quad + \LipConstF \int_{T-\delta}^T \E[ \lvert u(s,x+\fwpr_{s}) \rvert ] \dx s
+ \boundFG \int_{T-\delta}^T \E[(1+\lVert x + \fwpr_{s} \rVert)^p] \dx s
\dott 
\end{split}
\end{equation}
}
\argument{ 
Hutzenthaler et al.~\cite[Lemma 2.2]{HutzenthalerJentzenKruse2019} (applied with $d\with d$, $T\with T$, $L \with \LipConstF$, $B \with \boundFG$, $p\with p$, $q\with 1$, 
$f_1\with f$, $g_1\with g$, $u_1\with u$ 
in the notation of Hutzenthaler et al.~\cite[Lemma 2.2]{HutzenthalerJentzenKruse2019}); 
the triangle inequality; 
Jensen's inequality}{ that for all $x\in\R^d$, $s\in[0,T]$ it holds that 
\begin{equation}\llabel{temp_u_reg1_p8}
\begin{split}
\E\bigl[ \lvert u(s,x+\fwpr_{s}) \rvert \bigr] 
& \le 
e^{\LipConstF T}
(T+1) \boundFG 
3^{p-1} 
\bpr{ 1 + \lVert x \rVert^{p}  +  \sup\nolimits_{r\in[0,T]} \E[ \lVert \fwpr_{r} \rVert^{p} ] } 
\dott
\end{split}
\end{equation}
}
\argument{\lref{temp_u_reg1_p8}; the triangle inequality; Jensen's inequality}{ that for all $\delta \in (0,T]$, $x\in\R^d$, $t\in[0,T-\delta]$ it holds that 
\begin{equation}\llabel{temp_u_reg1_p9}
\begin{split}
& \LipConstF \delta T + \LipConstF \int_{T-\delta}^T \E[ \lvert u(s,x+\fwpr_{s}) \rvert ] \dx s
+ \boundFG \int_{T-\delta}^T \E[(1+\lVert x + \fwpr_{s} \rVert)^p] \dx s \\
& \le 
\LipConstF \delta T 
+ \LipConstF \delta 
e^{\LipConstF T}
(T+1) \boundFG 
3^{p-1} 
\bpr{ 1 + \lVert x \rVert^{p}  +  \sup\nolimits_{r\in[0,T]} \E[ \lVert \fwpr_{r} \rVert^{p} ] }
\\
& \quad + 
\boundFG \delta 3^{p-1} \bpr{ 1 + \lVert x \rVert^p + \sup\nolimits_{s\in[0,T]}\E\bigl[ \lVert \fwpr_{s} \rVert^{p} \bigr] }
\\
& \le
\delta 
e^{\LipConstF T} (T+1) (\LipConstF +1) (\boundFG+1) 3^{p}
\bpr{ 1 + \lVert x \rVert^p + \sup\nolimits_{s\in[0,T]}\E\bigl[ \lVert \fwpr_{s} \rVert^{p} \bigr] } \\
& \le \sqrt{\delta} 
e^{\LipConstF T} (T+1)^2 (\LipConstF +1) (\boundFG+1)3^{p+1} 
\bpr{ 1 + \lVert x \rVert^p + \sup\nolimits_{s\in[0,T]}\E\bigl[ \lVert \fwpr_{s} \rVert^{2p} \bigr] } 
\dott
\end{split}
\end{equation}
}
\argument{\lref{temp_u_reg1_p9};\lref{temp_u_reg1_p7};\cref{g_der_bd}}{ that for all $\delta \in (0,T]$, $x\in\R^d$, $t\in[0,T-\delta]$ it holds that 
\begin{equation}\llabel{temp_u_reg1_p10}
\begin{split}
& \E\bigl[ \lvert u(t+\delta,x+\fwpr_{t}) - u(t,x+\fwpr_{t}) \rvert \bigr] \\
& \le 
8^{p+1} \boundFG \bpr{ 1 + \norm{ x }^p + 1 + \sup\nolimits_{s\in[0,T]} \E\bigl[ \lVert \fwpr_{s} \rVert^{2p} \bigr] } 
\sqrt{\delta} \sqrt{d} \\
& \quad + \LipConstF \int_t^{T-\delta} \E\bigl[ \lvert u(s+\delta,x+\fwpr_{s}) - u(s,x+\fwpr_{s}) \rvert \bigr] \dx s  \\
& \quad + 
\sqrt{\delta} 
e^{\LipConstF T} (T+1)^2 (\LipConstF +1) (\boundFG+1)3^{p+1} 
\bpr{ 1 + \lVert x \rVert^p + \sup\nolimits_{s\in[0,T]}\E\bigl[ \lVert \fwpr_{s} \rVert^{2p} \bigr] } 
\\
& \le 
e^{\LipConstF T} (T+1)^2 (\LipConstF +1) (\boundFG+1) 8^{p+2} 
\bpr{ 1 + \lVert x \rVert^p + \sup\nolimits_{s\in[0,T]}\E\bigl[ \lVert \fwpr_{s} \rVert^{2p} \bigr] } 
\sqrt{\delta} \sqrt{d} 
\\
& \quad + \LipConstF \int_t^{T-\delta} \E\bigl[ \lvert u(s+\delta,x+\fwpr_{s}) - u(s,x+\fwpr_{s}) \rvert \bigr] \dx s  
\dott 
\end{split}
\end{equation}
}
\argument{\lref{temp_u_reg1_p10};\lref{temp_u_reg1_p8};\cref{lem:BM_power_bound};Gronwall's integral inequality (see \eg \cite[Corollary~2.2]{PadgettJentzen2021})}{ that for all $\delta \in (0,T]$, $x\in\R^d$, $t\in[0,T-\delta]$ it holds that 
\begin{equation}
\begin{split}
& \E\bigl[ \lvert u(t+\delta,x+\fwpr_{t}) - u(t,x+\fwpr_{t}) \rvert \bigr] \\
& \le 
e^{2 \LipConstF T} (T+1)^2 (\LipConstF +1) (\boundFG+1) 8^{p+2} 
\bpr{ 1 + \lVert x \rVert^p + \sup\nolimits_{s\in[0,T]}\E\bigl[ \lVert \fwpr_{s} \rVert^{2p} \bigr] } 
\sqrt{\delta} \sqrt{d} 
\dott
\end{split}
\end{equation}
}
\end{aproof}

\cfclear
\begin{athm}{corollary}{temp_u_reg1}
Let $d\in\N$, $T,\LipConstF,\boundFG \in(0,\infty)$, $p\in[1,\infty)$, 
let $f \in C([0,T]\times \R^d \times \R,\R)$ and $g\in C^1(\R^d,\R)$ satisfy for all $s,t\in[0,T],$ $x\in\R^d$, $v,w\in\R$ that 
\begin{equation}\label{temp_u_reg11_0a}
    \abs{f(s,x,v) - f(t,x,w)} \le \LipConstF (\abs{s-t} + \abs{v-w})
\end{equation}
\begin{equation}\label{temp_u_reg11_0b}
    \text{and} \qquad 
    \max\{\abs{f(t,x,0)},\lvert g(x)\rvert,\norm{\nabla g(x)} \} \le \boundFG (1 + \norm{x} )^p ,
\end{equation} 
let $(\Omega,\cF,\mathbb{P})$ be a probability space, let $\fwpr \colon [0,T] \times \Omega \to \R^d$ be a standard Brownian motion, and let $u \in C([0,T]\times \R^d, \R)$ satisfy for all $t\in[0,T]$, $x\in\R^d$ that
\begin{equation}\label{temp_u_reg11_f0}
    \E\biggl[\abs{g(x + \fwpr_{T-t})} + \int_t^T \abs{f\bpr{s,x+\fwpr_{s-t},u(s,x+\fwpr_{s-t})}} \dx s\biggr] < \infty
\end{equation}
\begin{equation}\label{temp_u_reg11_f1}
\text{and} \qquad 
u(t,x) = \E[g(x+\fwpr_{T-t})] + \int_t^T \E\bbr{f\bpr{s,x+\fwpr_{s-t},u(s,x+\fwpr_{s-t})}}\dx s
\end{equation}
\cfload.
Then it holds for all $\ft, t \in [0,T]$, $x\in\R^d$ 
that
\begin{equation}
\begin{split}  
& \abs{ u(\ft,x) - u(t,x) } \\
& \le 
e^{2 \LipConstF T} (T+1)^2 (\LipConstF +1) (\boundFG+1) 8^{p+2} 
\bpr{ 1 + \lVert x \rVert^p + \sup\nolimits_{s\in[0,T]}\E\bbr{ \lVert \fwpr_{s} \rVert^{2p} } }
\sqrt{\lvert \ft-t \rvert} \sqrt{d} 
\dott
\end{split}
\end{equation}
\end{athm}

\cfclear
\begin{aproof}
    Let $\fu_{\ft} \colon [0,T-\ft] \times \R^d \to \R$, $\ft\in[0,T]$, and $\ff_{\ft}\colon [0,T-\ft]\times\R^d\times \R \to \R$, $\ft \in [0,T]$, be the functions which satisfy for all $\ft \in [0,T]$, $t\in[0,T-\ft]$, $x\in\R^d$, $v\in\R$ that $\fu_{\ft}(t,x) = u(t+\ft,x)$ and $\ff_{\ft}(t,x,v) = f(t+\ft,x,v)$. 
    \startnewargseq
    \argument{\cref{shifted_time}; \cref{temp_u_reg1a}
    (applied for every $\ft \in [0,T]$ with 
    $d\with d$, $T \with T-\ft$, $\LipConstF \with \LipConstF$, $\boundFG \with \boundFG$, $p \with p$, $f \with \ff_{\ft}$, $u\with \fu_{\ft}$, $g\with g$ 
    in the notation of  \cref{temp_u_reg1a})}{
    that 
    for all $\ft \in[0,T]$, $t \in[0,T-\ft]$, $x \in \R^d$ it holds that
    \begin{equation}\llabel{eq:1652}
        \begin{split}
            & \abs{ u(t+\ft,x) - u(\ft,x) }
            = \abs{ \fu_{\ft}(t,x) - \fu_{\ft}(0,x) }
            = \E[ \abs{ \fu_{\ft}(t,x+\fwpr_0) - \fu_{\ft}(0,x+\fwpr_0) } ]
            \\
            & \le 
            e^{2 \LipConstF T} (T+1)^2 (\LipConstF +1) (\boundFG+1) 8^{p+2} 
            \bpr{ 1 + \lVert x \rVert^p + \sup\nolimits_{s\in[0,T]}\E\bigl[ \lVert \fwpr_{s} \rVert^{2p} \bigr] } 
            \sqrt{t} \sqrt{d} 
            \dott
        \end{split}
    \end{equation}
    }
    \argument{\lref{eq:1652}}{
    that for all $\ft \in[0,T]$, $t \in[\ft,T]$, $x \in \R^d$ it holds that
    \begin{equation}
        \begin{split}
            & \abs{ u(t,x) - u(\ft,x) } \\
            & \le 
            e^{2 \LipConstF T} (T+1)^2 (\LipConstF +1) (\boundFG+1) 8^{p+2} 
            \bpr{ 1 + \lVert x \rVert^p + \sup\nolimits_{s\in[0,T]}\E\bigl[ \lVert \fwpr_{s} \rVert^{2p} \bigr] } 
            \sqrt{\abs{t-\ft}} \sqrt{d} 
            \dott
        \end{split}
    \end{equation}
    }
\end{aproof}

\section{Artificial neural network (ANN) calculus}
\label{subsec:structured_description}

In our proofs of the \ANN\ approximation results in \cref{sec:ANN_approx_lin_interpolation_MLP,sec:ANN_approx_PDE,thm_intro} in the introduction we make use of a suitable calculus for \ANNs\ from the literature 
(cf., for example, \cite[Section~2]{GrohsHornungJentzen2019},  \cite[Section~1.3 and Chapter~2]{jentzen2023mathematical}, and the references therein). 
In this section we recall the ingredients of this \ANN\ calculus that we need in the later \ANN\ approximation results of this work. 
The notions in this section can -- often in a slightly modified form -- be found in 
\cite[Section~1.3 and Chapter~2]{jentzen2023mathematical} and \cite[Section~2]{ackermann2023deep}, for instance.

\subsection{ANNs}
\label{subsec:ANNs}

\begin{definition}[ANNs]\label{def:ANN}
We denote by $\ANNsm$ the set given by 
\begin{equation}\label{eq:defANN}
    \textstyle{
    \ANNsm =   \cup_{ L \in \N } 
  \cup_{ l_0, l_1, \dots, l_L \in \N } 
  \pr{ 
    \times_{ k = 1 }^L 
    \pr{
      \R^{ l_k \times l_{k-1} } \times \R^{ l_k } 
    }
  } }, 
\end{equation}
for every  
$L \in\N$, 
$l_0,l_1,\dots, l_L \in \N$, 
$
\Phi 
\in  \allowbreak
( \times_{k = 1}^L\allowbreak(\R^{l_k \times l_{k-1}} \times \R^{l_k}) ) \subseteq \ANNsm$
we denote by 
$\paramANN(\Phi),\allowbreak \lengthANN(\Phi),\allowbreak \inDimANN(\Phi),\allowbreak \outDimANN(\Phi), \allowbreak \hiddenLength(\Phi) \in \R$ 
the numbers given by 
\begin{equation}\label{eq:param_length_indim_outdim}
    \paramANN(\Phi) = \smallsum_{k = 1}^L l_k(l_{k-1} + 1),
    \qquad 
    \lengthANN(\Phi) = L, 
    \qquad 
    \inDimANN(\Phi) = l_0, 
    \qquad 
    \outDimANN(\Phi) = l_L, 
\end{equation}
and $\hiddenLength(\Phi) = L-1$, 
for every $n \in \N_0$, $L\in\N$, 
$l_0,l_1,\dots, l_L \in \N$, 
$
\Phi 
\in  \allowbreak
( \times_{k = 1}^L\allowbreak(\R^{l_k \times l_{k-1}} \times \R^{l_k}) ) \subseteq \ANNsm$
we denote by
$\dimANNlevel_n(\Phi) \in \R$ 
the number given by 
\begin{align}\label{eq:dimlevel}
\begin{split}
\dimANNlevel_n (\Phi) =
\begin{cases}
l_n &\colon n \leq L \\
0 &\colon n > L , 
\end{cases}
\end{split}
\end{align}
for every $\Phi \in \ANNsm$ we denote by 
$\dims(\Phi) \in \R^{\lengthANN(\Phi)+1}$ 
the vector given by 
\begin{equation}\label{eq:dimsvector}
    \dims(\Phi)
    = \pr{
    \dimANNlevel_0(\Phi), 
    \dimANNlevel_1(\Phi), 
    \dots,
    \dimANNlevel_{\lengthANN(\Phi)}(\Phi)
    },
\end{equation}
and for every 
$L \in\N$, 
$l_0,l_1,\dots, l_L \in \N$, 
$
\Phi 
=
((W_1, B_1),\allowbreak \dots, (W_L,\allowbreak B_L))
\in  \allowbreak
( \times_{k = 1}^L\allowbreak(\R^{l_k \times l_{k-1}} \times \R^{l_k}))
\subseteq \ANNsm$, 
$n\in\{1,2,\dots,L\}$
we denote by 
$\WNN_{n,\Phi} \in \R^{l_n\times l_{n-1}}$, 
$\BNN_{n,\Phi}\in\R^{l_n}$ 
the matrix and the vector given by 
\begin{equation}
    \WNN_{n,\Phi} = W_n \qquad \text{and} \qquad 
    \BNN_{n,\Phi} = B_n
    .
\end{equation}
\end{definition}

\cfclear
\begin{definition}[ANN]\label{def:ANN2}
\cfconsiderloaded{def:ANN2}
We say that $\Phi$ is an \ANN\ if and only if it holds that $\Phi \in \ANNsm$ \cfload.
\end{definition}

\subsection{Realizations of ANNs}
\label{subsec:realizations_of_ANNs}

\cfclear
\begin{definition}[Multidimensional version]\label{multi}
\cfconsiderloaded{multi}
Let $a\colon \R \to \R$ be a function and let $d\in\N$.  
Then we denote by $\Mmulti_{a,d}\colon\R^d\to\R^d$ the function which satisfies for all $x=(x_1,\dots,x_d)\in\R^d$ that 
\begin{equation}\label{multidim_version:Equation}
\Mmulti_{a,d}(x) = (a(x_1),\dots,a(x_d))
\dott
\end{equation}
\end{definition}

\cfclear
\begin{definition}[Realization associated to an ANN]
\label{def:ANNrealization}
\cfconsiderloaded{def:ANNrealization}
Let $a\colon \R \to \R$ be a function and let $\Phi \in \ANNsm$ \cfload. 
Then we denote by 
$\Ra(\Phi) \colon \R^{\inDimANN(\Phi)} \to \R^{\outDimANN(\Phi)}$ 
the function which satisfies for all $x_0 \in \R^{\dimANNlevel_0(\Phi)}$, $x_1 \in \R^{\dimANNlevel_1(\Phi)}$, $\dots$, $x_{\lengthANN(\Phi)} \in \R^{\dimANNlevel_{\lengthANN(\Phi)}(\Phi)}$ with 
$\forall\, k \in \{1,2,\dots,\lengthANN(\Phi)\}\colon x_k = \Mmulti_{a\1_{(0,\lengthANN(\Phi))}(k)+\id_{\R} \1_{\{\lengthANN(\Phi)\}}(k),\dimANNlevel_k(\Phi)}\pr{\WNN_{k,\Phi}x_{k-1}+\BNN_{k,\Phi}} $
that 
\begin{equation}
    \pr{\Ra(\Phi)}(x_0) = x_{\lengthANN(\Phi)}
\end{equation}
\cfout.
\end{definition}

\subsection{Activation ANNs}
\label{subsec:activation}

\cfclear
\begin{definition}[Identity matrices]\label{def:identityMatrix}
	Let $d\in\N$. Then we denote by $\idMatrix_{d}\in \R^{d\times d}$ the identity matrix in $\R^{d\times d}$.
\end{definition}

\cfclear
\begin{definition}[Activation ANNs]\label{padding}
\cfconsiderloaded{padding}
Let $d\in\N$.
Then we denote by
$\ii_d \in ((\R^{d\times d}\times \R^d)\times (\R^{d\times d}\times \R^d)) \subseteq \ANNsm$
the \ANN\ given by
$\ii_d = ((\idMatrix_d,0),(\idMatrix_d,0))$
\cfout.
\end{definition}

In \cref{ANN_for_hatfct1} in \cref{sec:ANN_approx_lin_interpolation_MLP} below we establish an elementary representation result for hat functions in terms of \ANNs\ with the leaky \ReLU\ activation functions. 
We employ \cref{ANN_for_hatfct1} in our proofs of the \ANN\ approximation results for \PDEs\ in \cref{sec:ANN_approx_PDE} and \cref{thm_intro} in the introduction, respectively. 
Our proof of \cref{ANN_for_hatfct1}, in turn, is based on the elementary fact that hat functions (and other piecewise linear functions) can be exactly represented by \ANNs\ with the \ReLU\ activation function (cf., for example, \cref{eq:1005a} 
and \cite[Lemma~4.10]{ackermann2023deep}) and the elementary fact that 
multidimensional versions of the \ReLU\ activation function can be exactly represented by leaky \ReLU\ \ANNs.  
These representations of multidimensional versions of the \ReLU\ activation through leaky \ReLU\ \ANNs\ are precisely the subject of the following elementary result, \cref{lemma_relation_leakyReLU_ReLU} below. 
Our statement of \cref{lemma_relation_leakyReLU_ReLU} employs the notion of the activation \ANNs\ in \cref{padding} above. 
In this context we note that the realization functions of the activation \ANNs\ exactly coincide with the multidimensional versions of the activation function under consideration (cf., for example, \cite[Lemma~3.2.2]{jentzen2023mathematical}).

\cfclear
\begin{athm}{lemma}{lemma_relation_leakyReLU_ReLU}
    Let $d\in\N$, $\leaky \in \R\backslash\{-1,1\}$ and let 
    $a\colon \R\to\R$ and $\fr\colon \R\to\R$ 
    satisfy for all $x \in \R$ that 
    $a(x)=\max\{x,\leaky x\}$ and $\fr(x)=\max\{x,0\}$
    \cfload.
    Then it holds for all $x \in \R^d$ that 
    \begin{equation}\label{eq:relation_leakyReLU_ReLU}
        \begin{split}
            (\Rr(\ii_d))(x) 
            & = 
            \frac{\abs{1-\leaky} }{(1-\leaky)(1-\leaky^2)} 
            \bbbbr{
            \leaky 
            \bpr{\Ra(\ii_d)} \bbbpr{\frac{-\abs{1-\leaky}x}{1-\leaky}} 
            +
            \bpr{\Ra(\ii_d)} \bbbpr{\frac{\abs{1-\leaky}x}{1-\leaky}}
            }
        \end{split}
    \end{equation}
    \cfout.
\end{athm}

\cfclear
\begin{aproof}
    \Nobs that \eg
    item~(iii) in Lemma~3.2 in~\cite{ackermann2023deep} establishes 
    that $\Rr(\ii_d)=\Mmulti_{\fr,d}$ and $\Ra(\ii_d)=\Mmulti_{a,d}$ \cfload. 
    \argument{the fact that for all $x,y\in\R$ it holds that $\max\{x,y\}=\tfrac12(x+y+\abs{x-y})$}{
    that for all $x\in\R$ it holds that 
    \begin{equation}\llabel{eq:1406}
        \begin{split}
            2 a\bbbpr{\frac{\abs{1-\leaky}x}{1-\leaky}} 
            & = \frac{\abs{1-\leaky}x}{1-\leaky} + \frac{\abs{1-\leaky}\leaky x}{1-\leaky}
            + \bbbabs{ \frac{\abs{1-\leaky}x}{1-\leaky} - \frac{\abs{1-\leaky}\leaky x}{1-\leaky} } \\
            & = \frac{\abs{1-\leaky}(1+\leaky)x}{1-\leaky}
            + \abs{1-\leaky}\abs{x}
            \dott 
        \end{split}
    \end{equation}
    }
	\Hence for all $x \in \R$ that 
    \begin{equation}\llabel{eq:1418a}
        \begin{split}
            & \leaky 
            a\bbbpr{\frac{-\abs{1-\leaky}x}{1-\leaky}} 
            +
            a\bbbpr{\frac{\abs{1-\leaky}x}{1-\leaky}} \\
            & = 
            \frac12 \bbbpr{
            \frac{-\leaky \abs{1-\leaky}(1+\leaky) x}{1-\leaky}
            + \leaky  \abs{1-\leaky}\abs{-x}
            + \frac{\abs{1-\leaky}(1+\leaky)x}{1-\leaky}
            + \abs{1-\leaky}\abs{x}
            } \\
            & = \frac12 \bbbpr{
            \frac{(1-\leaky) \abs{1-\leaky}(1+\leaky) x}{1-\leaky}
            + (1+\leaky)  \abs{1-\leaky}\abs{x}
            } \\
            & = \frac{\abs{1-\leaky}(1+\leaky)(x+\abs{x})}{2}
            \dott 
        \end{split}
    \end{equation}
	\Hence for all $x \in \R$ that 
    \begin{equation}\llabel{eq:1418b}
        \begin{split}
            & \frac{\abs{1-\leaky} }{(1-\leaky)(1-\leaky^2)} 
            \bbbbr{
            \leaky 
            a\bbbpr{\frac{-\abs{1-\leaky}x}{1-\leaky}} 
            +
            a\bbbpr{\frac{\abs{1-\leaky}x}{1-\leaky}}
            } \\
            & = \frac{\abs{1-\leaky} }{(1-\leaky)(1-\leaky)(1+\leaky)} 
            \frac{\abs{1-\leaky}(1+\leaky)(x+\abs{x})}{2} 
            = \frac{x+\abs{x}}{2}
            = \max\{x,0\} = \fr(x)
            \dott 
        \end{split}
    \end{equation}
    Combining this with the fact that $\Rr(\ii_d)=\Mmulti_{\fr,d}$ and the fact that $\Ra(\ii_d)=\Mmulti_{a,d}$
    establishes~\cref{eq:relation_leakyReLU_ReLU}. 
\end{aproof}

\subsection{Compositions, powers, and extensions of ANNs}
\label{subsec:compositions_of_dnns}

\cfclear
\begin{definition}[Composition of ANNs]
\label{def:ANNcomposition}
\cfconsiderloaded{def:ANNcomposition}
Let $\Phi,\Psi \in\ANNsm$ satisfy 
$\inDimANN(\Phi)=\outDimANN(\Psi)$ \cfload. 
Then we denote by $\compANN{\Phi}{\Psi} \in \ANNsm$ 
the \ANN\ which satisfies for all 
$k\in\N\cap (0,\lengthANN(\Phi)+\lengthANN(\Psi))$ 
that $\lengthANN(\compANN{\Phi}{\Psi})=\lengthANN(\Phi)+\lengthANN(\Psi)-1$ 
and 
\begin{equation}\label{ANNoperations:Composition}
\begin{split}
& (\WNN_{k,\compANN{\Phi}{\Psi}},\BNN_{k,\compANN{\Phi}{\Psi}})
=
\begin{cases} 
(\WNN_{k,\Psi},\BNN_{k,\Psi})
& 
\colon  
k<\lengthANN(\Psi) 
\\
(\WNN_{1,\Phi}\WNN_{\lengthANN(\Psi),\Psi},\WNN_{1,\Phi}\BNN_{\lengthANN(\Psi),\Psi}+\BNN_{1,\Phi})
&
\colon 
k=\lengthANN(\Psi)
\\
(\WNN_{k-\lengthANN(\Psi)+1,\Phi},\BNN_{k-\lengthANN(\Psi)+1,\Phi})
&
\colon 
k> \lengthANN(\Psi)
\dott 
\end{cases}
\end{split}
\end{equation}
\end{definition}

\cfclear
\begin{definition}[Affine transformation ANNs]\label{linear}
	\cfconsiderloaded{linear}
	Let 
	$m,n\in\N$, 
	$\weight\in\R^{m\times n}$, 
	$\bias\in\R^m$ \cfload. 
	Then we denote by 
	$\A_{\weight,\bias} \in (\R^{m\times n}\times \R^m) \subseteq \ANNsm$ the ANN
	given by $\A_{\weight,\bias} = (\weight,\bias)$
	\cfout.
\end{definition}

\cfclear
\begin{definition}[Powers of ANNs]\label{def:iteratedANNcomposition}
\cfconsiderloaded{def:iteratedANNcomposition}
Let $\Phi \in\ANNsm$ satisfy 
$\inDimANN(\Phi)=\outDimANN(\Phi)$ \cfload. 
Then we denote by 
$\power{\Phi}{n} \in \cu{ \Psi \in \ANNsm \colon \inDimANN(\Psi)=\outDimANN(\Psi)=\inDimANN(\Phi) }$, 
$n \in \N_0$, 
the \ANNs\ which satisfy 
for all $n \in \N_0$ that  
\begin{equation}\label{iteratedANNcomposition:equation}
\begin{split}
\power{\Phi}{n} 
=
\begin{cases} 
\A_{\idMatrix_{\outDimANN(\Phi)},0} 
&
\colon 
n=0 
\\
\compANN{\Phi}{ \pr{ \power{\Phi}{(n-1)} } } 
&
\colon 
n>0 
\end{cases}
\end{split}
\end{equation}	
\cfload.
\end{definition}

\cfclear
\begin{definition}[Extensions of ANNs]\label{def:ANNenlargement}
\cfconsiderloaded{def:ANNenlargement}
Let $L\in\N$, $\Phi, \Psi \in\ANNsm$ satisfy 
$\lengthANN(\Phi)\le L$ and 
$\outDimANN(\Phi)=\inDimANN(\Psi)=\outDimANN(\Psi)$ \cfload. 
Then we denote by $\longerANN{L,\Psi}(\Phi) \in\ANNsm$ the \ANN\ 
given by 
\begin{equation}\label{ANNenlargement:Equation}
\longerANN{L,\Psi}(\Phi)
=	 
\compANN{\pr{ \power{\Psi}{(L-\lengthANN(\Phi))} } }{\Phi}
\end{equation}
\cfout.
\end{definition}

\subsection{Parallelizations of ANNs}
\label{subsec:parallelizations_of_dnns}

\cfclear
\begin{definition}[Parallelization of \ANNs\ with the same length]
\label{def:simpleParallelization}
\cfconsiderloaded{def:simpleParallelization}
Let $n \in \N$, $\Phi=(\Phi_1,\dots,\Phi_n) \allowbreak \in \ANNsm^n$ 
satisfy $\lengthANN(\Phi_1)=\lengthANN(\Phi_2)=\ldots=\lengthANN(\Phi_n)$ \cfload. 
Then we denote by $\parallelizationSpecial_n(\Phi) \in \ANNsm$ the \ANN\ which satisfies 
that 
$\lengthANN(\parallelizationSpecial_{n}(\Phi)) = \lengthANN(\Phi_1)$
and that 
for all  
$k\in\{1,2,\dots,\lengthANN(\Phi_1)\}$ 
it holds 
that 
\begin{equation}
    \WNN_{k,\parallelizationSpecial_n(\Phi)}
    = 
    \begin{pmatrix}
    \WNN_{k,\Phi_1}& 0& 0& \cdots& 0\\
    0& \WNN_{k,\Phi_2}& 0&\cdots& 0\\
    0& 0& \WNN_{k,\Phi_3}&\cdots& 0\\
    \vdots& \vdots&\vdots& \ddots& \vdots\\
    0& 0& 0&\cdots& \WNN_{k,\Phi_n}
    \end{pmatrix} 
    \quad \text{and} \quad 
    \BNN_{k,\parallelizationSpecial_n(\Phi)}
    = 
    \begin{pmatrix}
    \BNN_{k,\Phi_1}\\
    \BNN_{k,\Phi_2}\\
    \vdots\\ 
    \BNN_{k,\Phi_n}
    \end{pmatrix}
    \dott 
\end{equation} 
\end{definition}

\cfclear
\begin{definition}[Parallelization of \ANNs\ with different lengths]\label{def:generalParallelization}
\cfconsiderloaded{def:generalParallelization}
    Let $n\in\N$, $\Psi \in \ANNsm$, 
    $\Phi=(\Phi_1,\dots,\Phi_n)\in\ANNsm^n$ 
    satisfy 
    $\hiddenLength(\Psi)=1$
    and 
    $\outDimANN(\Phi_1)=\ldots=\outDimANN(\Phi_n)=\inDimANN(\Psi)=\outDimANN(\Psi)$ \cfload. 
    Then we denote by 
    $\parallelization_{n,\Psi}(\Phi) \in\ANNsm$ the \ANN\ 
    given by 
    \begin{equation}\label{generalParallelization:DefinitionFormula}
		\parallelization_{n,\Psi}(\Phi)
        =\parallelizationSpecial_{n}\bpr{ \longerANN{\max_{k\in\{1,2,\dots,n\}}\lengthANN(\Phi_k),\Psi}({\Phi_1}),\dots,\longerANN{\max_{k\in\{1,2,\dots,n\}}\lengthANN(\Phi_k),\Psi}({\Phi_n}) }
	\end{equation}
	\cfout.
\end{definition}

In our \ANN\ approximation results in \cref{sec:ANN_approx_PDE} we reformulate the \PDE\ under consideration from a terminal value \PDE\ problem (as \PDEs\ are often formulated in the finance/stochastic analysis literature) to an initial value \PDE\ problem (as \PDEs\ are often presented in the physics literature). 
In particular, in \cref{cor_of_mainthm1}  
in \cref{sec:ANN_approx_PDE} the \PDE\ approximation problem is formulated as a terminal value \PDE\ problem in which the terminal value function of the \PDE\ is considered to be (explicitly representable or) approximable by \ANNs\ without the \COD\ 
(see \cref{eq:cor1:heateq} and \cref{eq:assump_approx_cor1} in \cref{cor_of_mainthm1} for details) 
and in \cref{cor_of_mainthm2} in 
\cref{sec:ANN_approx_PDE} the \PDE\ approximation problem is formulated as an initial value \PDE\ problem in which the initial value function of the \PDE\ is considered to be (explicitly representable or) approximable by \ANNs\ without the \COD\ 
(see \cref{eq:cor2:heateq} and \cref{eq:2305} in \cref{cor_of_mainthm2} for details). 
In the next elementary result, \cref{lemma_time_reversal_ANN} below, we provide a suitable elementary transformation result for \ANN\ approximations that allows us to suitably shift/transform the temporal variable of the realization functions of space-time \ANN\ approximations. 
In our proof of \cref{cor_of_mainthm2} 
we apply \cref{lemma_time_reversal_ANN} in conjunction with the \ANN\ approximation result in \cref{cor_of_mainthm1}  
to establish the \ANN\ approximation result in \cref{cor_of_mainthm2}. 
Our proof of \cref{lemma_time_reversal_ANN}, in turn, is based on applications of appropriate \ANN\ calculus results in the literature (cf., for instance, \cite[Section~2.1]{ackermann2023deep}, \cite[Sections~2.2 and~2.3]{GrohsHornungJentzen2019}, and \cite[Section~2.2]{jentzen2023mathematical}).

\cfclear
\begin{athm}{lemma}{lemma_time_reversal_ANN}
    Let $T,c\in \R$, $d, \fd \in \N$, 
    $a\in C(\R,\R)$, 
    $\F, \fJ, \G \in \ANNsm$
    satisfy $\dims(\fJ)=(1,\fd,1)$, 
    $\Ra(\fJ) = \operatorname{id}_\R$, 
    $\Ra(\F) \in C(\R^{d+1},\R)$, 
    and 
    \begin{equation}
        \G = \compANN{\F}{ \parallelization_{d+1,\fJ}\bpr{
        \A_{c,T},\fJ,\fJ,\dots,\fJ} } 
    \end{equation}
    \cfload.
    Then 
    \begin{enumerate}[label=(\roman *)]
    \item
    \label{lemma_time_reversal_ANN_item0}
    it holds that $\Ra(\G) \in C(\R^{d+1},\R)$, 
    \item
    \label{lemma_time_reversal_ANN_item1}
        it holds for all $s \in \R$, $x \in \R^d$ that 
        $\pr{\Ra(\G)}(s,x) = \pr{\Ra(\F)}(T+cs,x)$, 
        and 
    \item
    \label{lemma_time_reversal_ANN_item2}
        it holds that 
        $\paramANN(\G) \le \paramANN(\F) \bpr{ 1+ 32\fd^2 d^2 + 2\fd d }
            \le  96 \fd^2 d^2 \paramANN(\F)$\dott
    \end{enumerate}
\end{athm}

\cfclear
\begin{aproof}
    Throughout this proof let $\Phi\in\ANNsm$ satisfy  
    \begin{equation}\llabel{eq:1711}
        \Phi 
        = \parallelization_{d+1,\fJ}\bpr{
        \A_{c,T},\fJ,\fJ,\dots,\fJ}  
    \end{equation}
    \cfload. 
    \Nobs that \cref{ANNoperations:Composition}, \cref{iteratedANNcomposition:equation}, \cref{ANNenlargement:Equation}, \cref{generalParallelization:DefinitionFormula}, and \lref{eq:1711} show that 
    \begin{equation}\llabel{eq:1054}
        \Phi 
        = \parallelizationSpecial_{d+1}\pr{
            \longerANN{2,\fJ}\pr{\A_{c,T}},\allowbreak\fJ,\fJ,\dots,\fJ}
        \dott 
    \end{equation}
    \cfload\dott 
    \Moreover \eg item~(i) in Corollary~2.23 in~\cite{GrohsHornungJentzen2019} and \eg 
    item~(vi) in Proposition~2.6 in~\cite{GrohsHornungJentzen2019} 
    establish \cref{lemma_time_reversal_ANN_item0}. 
    \startnewargseq
    \argument{\eg item~(vi) in Proposition~2.6 in~\cite{GrohsHornungJentzen2019}}{
    that for all $s \in \R$, $x \in \R^d$ it holds that 
    \begin{equation}\llabel{eq:2343}
        \begin{split}
            \bpr{\Ra(\G)}(s,x)
            & = \bpr{\bpr{\Ra(\F)}\circ \bpr{\Ra(\Phi)}}(s,x)
            \dott
        \end{split}
    \end{equation}
    }
    Furthermore, \nobs that the fact that $\Ra(\fJ)=\id_{\R}$ and \eg item~(ii) in Corollary~2.23 in~\cite{GrohsHornungJentzen2019}
    show 
    that for all $s \in \R$, $x \in \R^d$ it holds that 
    \begin{equation}\llabel{eq:1658}
        \begin{split}
            \bpr{\Ra(\Phi)}(s,x)
            & = 
            \bpr{  
            \bpr{\Ra\bpr{\A_{c,T}}}(s),
            x
            }
            =
            \pr{cs+T,x}
            \dott
        \end{split}
    \end{equation}
    This and \lref{eq:2343} prove 
    \cref{lemma_time_reversal_ANN_item1}. 
    \startnewargseq
    \argument{\eg item~(v) in Proposition~2.6 in~\cite{GrohsHornungJentzen2019}}{
    that 
    \begin{equation}\llabel{eq:2358a}
        \begin{split}
            \paramANN(\G)
            & \le
            \paramANN(\F)
            + \paramANN(\Phi)
            + \dimANNlevel_1(\F) \cdot 
            \dimANNlevel_{\lengthANN(\Phi)-1}(\Phi)
            \dott
        \end{split}
    \end{equation}
    }
    \argument{\lref{eq:1054}; \eg \cite[Proposition~2.20]{GrohsHornungJentzen2019}
    }{
    that 
    \begin{equation}\llabel{eq:2358b}
        \lengthANN(\Phi)-1=1, 
        \qquad 
        \dimANNlevel_1(\Phi)
            = \dimANNlevel_1\bpr{ \longerANN{2,\fJ}\bpr{\A_{c,T}} }
            + \sum_{k=1}^d \dimANNlevel_1(\fJ),
    \end{equation}
    \begin{equation}\llabel{eq:2358c}
        \text{and} \qquad 
        \paramANN(\Phi)
            \le 
            \frac12 
            \bbbbr{ 
            \paramANN\bpr{ \longerANN{2,\fJ}\bpr{\A_{c,T}} }
            + \sum_{k=1}^d \paramANN(\fJ) }^2
            \dott 
    \end{equation}   
    }
    \argument{
    \eg \cite[Lemma 2.2.11]{jentzen2023mathematical}}{
    that 
    \begin{equation}\llabel{eq:2358d}
        \dimANNlevel_1\bpr{ \longerANN{2,\fJ}\bpr{\A_{c,T}} }
        = \dimANNlevel_1(\fJ)
        \dott 
    \end{equation}
    }
    \argument{\lref{eq:2358d}; \lref{eq:2358b}; the fact that $\dims(\fJ)=(1,\fd,1)$}{
    that 
    \begin{equation}\llabel{eq:2358e}
        \dimANNlevel_1(\Phi)
        = (d+1) \fd
        \le 2\fd d 
        \dott 
    \end{equation}
    }
    \argument{\eg item~(ii) in Lemma~2.13 in~\cite{GrohsHornungJentzen2019}}{
    that 
    \begin{equation}\llabel{eq:2358f}
        \begin{split}
            \paramANN\bpr{ \longerANN{2,\fJ}\bpr{\A_{c,T}} }
            & \le \fd \paramANN\bpr{\A_{c,T}} 
            + \fd + 1
            = 2\fd + \fd + 1
            \le 4 \fd 
            \dott 
        \end{split}
    \end{equation}
    }
    \argument{the fact that $\dims(\fJ)=(1,\fd,1)$}{
    that 
    \begin{equation}\llabel{eq:1700}
        \paramANN(\fJ)=2\fd + \fd +1 \le 4 \fd
        \dott 
    \end{equation}
    }
    \argument{\lref{eq:1700}; \lref{eq:2358c}; \lref{eq:2358f}}{
    that 
    \begin{equation}\llabel{eq:1627}
        \begin{split}
            \paramANN(\Phi)
            & \le \tfrac12 \bpr{4\fd +4\fd d}^2
            \le 32 \fd^2 d^2 
            \dott 
        \end{split}
    \end{equation}
    }
    \argument{\lref{eq:1627}; \lref{eq:2358a}; \lref{eq:2358e};\eg \cite[Lemma 2.4]{ackermann2023deep}}{
    that 
    \begin{equation}\llabel{eq:0931}
        \begin{split}
            \paramANN(\G)
            & \le \paramANN(\F)
            + 32\fd^2 d^2
            + \dimANNlevel_1(\F) \cdot 
            2\fd d
            \le \paramANN(\F) \bpr{ 1+ 32\fd^2 d^2 + 2\fd d }
            \dott
        \end{split}
    \end{equation}
    }
    \argument{\lref{eq:0931}}{
    \cref{lemma_time_reversal_ANN_item2}\dott{}
    }
\end{aproof}

\subsection{Scalar multiplications and sums of ANNs}
\label{subsec:sums}

\cfclear
\begin{definition}[Scalar multiplications of ANNs]
\label{def:ANNscalar}
\cfconsiderloaded{def:ANNscalar}
Let $\lambda \in \R$, $\Phi \in \ANNsm$ \cfload. 
Then we denote by $\scalar{\lambda}{\Phi}\in\ANNsm$ the \ANN\ 
given by 
\begin{equation}\label{eq:ANNscalar}
    \scalar{\lambda}{\Phi} = \compANN{\A_{\lambda \idMatrix_{\outDimANN(\Phi)},0}}{\Phi}
\end{equation}
\cfout.
\end{definition}

\cfclear
\begin{definition}[Summation ANNs]\label{def:ANN:sum}
\cfconsiderloaded{def:ANN:sum}
Let $m, n \in \N$. 
Then we denote by $\sumANN_{m, n} \in (\R^{m \times (nm)} \times \R^m)$ the ANN
given by 
$\sumANN_{m, n} = \A_{(\idMatrix_m \,\,\,  \idMatrix_m \,\,\, \ldots \,\,\, \idMatrix_m), 0 }$
\cfout.
\end{definition}

\cfclear
\begin{definition}[Transpose of matrices]\label{def:Transpose}
\cfconsiderloaded{def:Transpose}
Let 
$m, n \in \N$, 
$A \in \R^{m \times n}$. 
Then we denote by $A^\transpose \in \R^{n \times m}$ the transpose of A
\cfout.
\end{definition}

\cfclear
\begin{definition}[Vectorization ANNs]\label{def:ANN:extension}
\cfconsiderloaded{def:ANN:extension}
Let $m, n \in \N$. 
Then we denote by $\extensionANN_{m, n} \in (\R^{(nm) \times m} \times \R^{nm})$ the \ANN\ given by 
$\extensionANN_{m, n} = \A_{(\idMatrix_m \,\,\,  \idMatrix_m \,\,\, \ldots \,\,\, \idMatrix_m)^\transpose , 0}$ 
\cfout.
\end{definition}

\cfclear
\begin{definition}[Sums of \ANNs\ with the same length]
\label{def:ANNsum:same}
\cfconsiderloaded{def:ANNsum:same}
Let 
$\lbd \in \Z$, 
$\ubd \in \Z \cap [\lbd,\infty)$, 
$\Phi_\lbd, \Phi_{\lbd+1}, \dots, \Phi_\ubd \in \ANNsm$ 
satisfy for all  
$k \in \Z \cap [\lbd,\ubd]$ 
that
$\lengthANN(\Phi_k) = \lengthANN(\Phi_\lbd)$,
$\inDimANN(\Phi_k) = \inDimANN(\Phi_\lbd)$, 
and 
$\outDimANN(\Phi_k) = \outDimANN(\Phi_\lbd)$ \cfload.
Then we denote by $\oSum_{k=\lbd}^\ubd \Phi_k$ (we denote by $\Phi_\lbd \oSum \Phi_{\lbd+1} \oSum \dots \oSum \Phi_\ubd$)
the \ANN\ given by
\begin{equation}
\OSum{k=\lbd}{\ubd} \Phi_k 
= 
\bbpr{ \compANN{\sumANN_{\outDimANN(\Phi_\lbd), \ubd-\lbd+1}}{{\compANN{\bbr{\parallelizationSpecial_{\ubd-\lbd+1}(\Phi_\lbd,\Phi_{\lbd+1},\dots, \Phi_\ubd)}}{\extensionANN_{\inDimANN(\Phi_\lbd), \ubd-\lbd+1}}}} } 
\in 
\ANNsm
\end{equation}
\cfout.
\end{definition}

\cfclear
\begin{definition}[Sums of \ANNs\ with different lengths]\label{def:ANN:sum_diff}
\cfconsiderloaded{def:ANN:sum_diff}
Let 
$\lbd\in\Z$, 
$\ubd \in \Z \cap [\lbd,\infty)$, 
$\Phi_\lbd, \allowbreak \Phi_{\lbd+1}, \allowbreak \dots, \allowbreak \Phi_\ubd, \Psi \in \ANNsm$ 
satisfy
for all $k \in \Z \cap [\lbd,\ubd]$ that
$\inDimANN(\Phi_k) = \inDimANN(\Phi_\lbd)$, 
$\outDimANN(\Phi_k) = \inDimANN(\Psi) = \outDimANN(\Psi)$, 
and 
$\hiddenLength(\Psi) = 1$ \cfload.
Then we denote by $\bSum_{k=\lbd,\Psi}^\ubd\Phi_k$ (we denote by $\Phi_\lbd\bSum_\Psi \Phi_{\lbd+1} \bSum_\Psi \dots \bSum_\Psi \Phi_\ubd$) the \ANN\ given by
\begin{equation}
\BSum{k=\lbd}{\Psi}{\ubd} \Phi_k 
= 
\bbbr{
\OSum{k=\lbd}{\ubd} 
\longerANN{\max_{j\in\{\lbd,\lbd+1,\dots,\ubd\}}\lengthANN(\Phi_j),\Psi}(\Phi_k) 
}
\in 
\ANNsm
\end{equation}
\cfout.
\end{definition}

In the following elementary result, \cref{diff_add_lemma} below, we collect a few basic properties for sums of \ANNs\ with different lengths (see \cref{def:ANN:sum_diff} above). 
\cref{diff_add_lemma} is a direct consequence of \cite[Lemma~2.20]{ackermann2023deep} and \cite[Lemma~2.2.11]{jentzen2023mathematical}, for example.

\cfclear
\begin{lemma}[Elementary properties of sums of \ANNs\ with different lengths]\label{diff_add_lemma}
Let $a\in C(\R,\R)$, 
$L\in\N$, 
$\lbd\in\Z$, 
$\ubd \in \Z \cap [\lbd,\infty)$, 
$\Phi_\lbd, \allowbreak \Phi_{\lbd+1}, \allowbreak \dots, \allowbreak \Phi_\ubd, \fJ, \G \in \ANNsm$ 
satisfy
for all $k \in \Z \cap [\lbd,\ubd]$ that
$L = \max_{k\in\Z \cap [\lbd,\ubd]} \lengthANN(\Phi_k)$, 
$\inDimANN(\Phi_k) = \inDimANN(\Phi_\lbd)$, 
$\outDimANN(\Phi_k) = \inDimANN(\fJ) = \outDimANN(\fJ)$, 
$\hiddenLength(\fJ) = 1$, 
$\Ra(\fJ) = \operatorname{id}_\R$,
and $\G=\bSum_{k=\lbd,\fJ}^\ubd \Phi_k$ \cfload. 
Then 
\begin{enumerate}[label=(\roman{*})]
\item\label{diff_add_lemma_item1}
it holds that 
$\lengthANN(\G ) = L$, 
\item\label{diff_add_lemma_item2}
it holds that
\begin{align}
& 
\dims(\G) 
\\
& 
= 
\Bigl( \inDimANN(\Phi_\lbd), \SmallSum{k=\lbd}{\ubd} \dimANNlevel_1\pr[\big]{ \longerANN{L,\fJ}(\Phi_k) }, \SmallSum{k=\lbd}{\ubd} \dimANNlevel_2\pr[\big]{ \longerANN{L,\fJ}(\Phi_k) } ,  \dots, \SmallSum{k=\lbd}{\ubd} \dimANNlevel_{L - 1} \pr[\big]{ \longerANN{L,\fJ}(\Phi_k) } , \outDimANN(\Phi_\lbd) \Bigr) 
\dc 
\nonumber
\end{align}
\item\label{diff_add_lemma_item3}
it holds that 
\begin{equation}
    \normmm{\dims(\G)} \le (\ubd - \lbd + 1) 
    \max\bbcu{ \dimANNlevel_1(\fJ), \max_{k\in\Z\cap[\lbd,\ubd]} \normmm{\dims(\Phi_k)} }, 
\end{equation}
\item\label{diff_add_lemma_item4}
it holds that $\Ra(\G) \in C(\R^{\inDimANN(\Phi_\lbd)},\R^{\outDimANN(\Phi_\lbd)})$, and
\item\label{diff_add_lemma_item5}
it holds for all $x \in \R^{\inDimANN(\Phi_\lbd)}$ that
\begin{equation}
(\Ra(\G))(x) 
= 
\SmallSum{k=\lbd}{\ubd} (\Ra(\Phi_k))(x)
\end{equation}
\end{enumerate}
\cfout.
\end{lemma}

\cfclear
\begin{aproof}
\Nobs that 
\eg \cite[Lemma~2.20]{ackermann2023deep} (applied with $\psi \with \G$, $(h_k)_{k\in \Z\cap [\lbd,\ubd]} \with (1)_{k\in \Z\cap [\lbd,\ubd]}$, $(B_k)_{k\in \Z\cap [\lbd,\ubd]} \with (0)_{k\in \Z\cap [\lbd,\ubd]}$ in the notation of \cite[Lemma~2.20]{ackermann2023deep}), 
\cref{ANNoperations:Composition}, and \cref{eq:ANNscalar} 
establish 
\cref{diff_add_lemma_item2,diff_add_lemma_item4,diff_add_lemma_item5}. 
\Nobs that \cref{diff_add_lemma_item2} implies \cref{diff_add_lemma_item1}. 
\startnewargseq
\argument{
\eg \cite[Lemma~2.2.11]{jentzen2023mathematical}
}{
that for all $k\in\Z\cap[\lbd,\ubd]$ it holds that 
\begin{equation}\llabel{eq:1236}
    \normmm{\dims(\longerANN{L,\fJ}(\Phi_k))} \le \max\cu{ \dimANNlevel_1(\fJ), \normmm{\dims(\Phi_k)} }
    \dott 
\end{equation}  
} 
\argument{\cref{diff_add_lemma_item2}}{ 
that 
\begin{equation}\label{eq:diff_add_dims_bound_1}
\normmm{\dims(\G)} \le \max\bbcu{ \inDimANN(\Phi_\lbd), \outDimANN(\Phi_\lbd), (\ubd-\lbd+1) \max_{k\in\Z\cap[\lbd,\ubd]}\normmm{\dims(\longerANN{L,\fJ}(\Phi_k))} }
\dott 
\end{equation} 
}
\argument{\cref{eq:diff_add_dims_bound_1};\lref{eq:1236}}{
\cref{diff_add_lemma_item3}\dott{}
}
\end{aproof}

\section{ANN approximations for linear interpolations of multilevel Picard (MLP) approximations}
\label{sec:ANN_approx_lin_interpolation_MLP}

One of the main goals of this section is to construct and study in \cref{ANN_for_MLP2}  below suitable \ANNs\ (with general/abstract activations) which approximate linear interpolations of appropriate \MLP\ approximations (see \cref{eq:ANN_for_MLP2_approx} in \cref{ANN_for_MLP2} in \cref{subsec:ANN_approx_MLP} below for details). 
We employ \cref{ANN_for_MLP2} in our proofs of the \ANN\ approximation results for \PDEs\ in \cref{sec:ANN_approx_PDE} and \cref{thm_intro} in the introduction, respectively. 
In our proof of \cref{ANN_for_MLP2} we employ the abstract \ANN\ approximation result for interpolation functions in \cref{ANN_for_MLP1} and the \ANN\ approximation result for \MLP\ approximations at fixed time points in \cite[Proposition~3.9]{ackermann2023deep}.  
Moreover, the statements and our proofs of \cref{ANN_for_MLP1} and \cref{ANN_for_MLP2} build up on the concepts and results of the \ANN\ calculus from \cref{subsec:structured_description}, \cite[Sections~2.2 and~2.3]{GrohsHornungJentzen2019}, \cite[Section~2.1]{ackermann2023deep}, and \cite[Section~2.2]{jentzen2023mathematical}, respectively.

In \cref{ANN_for_MLP1} and \cref{ANN_for_MLP2} the activation function of the considered \ANNs\ is a general continuous function $a\colon \R \to \R$ which fulfills, among other assumptions, the condition that it makes the class of \ANNs\ with this activation function flexible enough to exactly represent the one-dimensional identity function 
$\id_{\R} = ( \R \ni x \mapsto x \in \R)$ and to approximately represent the product function $\R^2 \ni (v,w) \mapsto v w \in \R$ (see \cref{eq:ANN_for_MLP1_product} in \cref{ANN_for_MLP1} and \cref{eq:ANN_for_MLP2_product} in \cref{ANN_for_MLP2} for details) in a suitable way. 
In the \ANN\ approximation results in \cref{ANN_for_product1}, \cref{ANN_for_product2}, \cref{ANN_for_product12}, \cref{ANN_for_product3} (\ReLU\ and leaky \ReLU\ activations), and \cref{ANN_for_product4} (softplus activation) 
in \cref{subsec:approx_squarefct,subsec:approx_product} 
below we verify that the assumption in \cref{ANN_for_MLP2} that the considered \ANNs\ can approximately represent the product function in a suitable way is satisfied in the situation of the \ReLU\ activation function 
$\fr = (\R \ni x \mapsto \max\{x,0\} \in \R)$, in the situation of the leaky \ReLU\ activation functions 
$\R \ni x \mapsto \max\{ x, \leaky x \} \in\R$ for $\leaky \in (0,1)$, and in the situation of the softplus activation function 
$\R \ni x \mapsto \ln(1+\exp(x)) \in \R$. 
\Cref{ANN_for_product1} (an appropriate \ANN\ approximation result for the square function $\R \ni x \mapsto x^2 \in \R$) is an extension of Grohs et al.~\cite[Proposition~3.4]{GrohsHornungJentzen2019} in which a result similar to \cref{ANN_for_product1} has been established in the special situation of the \ReLU\ activation function 
and \cref{ANN_for_product2} (a suitable \ANN\ approximation result for the product function $\R^2 \ni (v,w) \mapsto v w \in \R$) is an extension of Grohs et al.~\cite[Proposition~3.5]{GrohsHornungJentzen2019} in which a result similar to \cref{ANN_for_product2} has been established in the special situation of the \ReLU\ activation function. 
Our proofs of \cref{ANN_for_product1} and \cref{ANN_for_product2} are strongly based on the proofs in Grohs et al.~\cite[Proposition~3.4 and Proposition~3.5]{GrohsHornungJentzen2019}.

In \cref{subsec:lin_interpolation} we recall in \cref{def:lin_interp} (which coincides with Definition~4.5 in \cite{ackermann2023deep}), roughly speaking, the concept of a continuous piecewise linear (more accurately, piecewise affine) function which interpolates certain given values at certain given points/\allowbreak arguments/\allowbreak positions while being affine on the intervals between two neighboring position points and in 
\cref{lin_interp1}, \cref{lin_interpol_hat_fct}, and \cref{factorize_lin_interpolation} 
we collect a few elementary and well-known properties of such linear interpolation functions. 
In~\cref{eq:factorize_lin_interpolation} in \cref{factorize_lin_interpolation} we recall that such a linear interpolation function can be written as a linear combination of hat functions: the left hand side of~\cref{eq:factorize_lin_interpolation} is the considered linear interpolation function from \cref{def:lin_interp} and the right hand side of~\cref{eq:factorize_lin_interpolation} is the linear combination of the hat functions. 
The linear combination on the right hand side of \cref{eq:factorize_lin_interpolation} consists of a sum of $(K+1)$ summands (where $K \in \N$ is an arbitrary natural number) consisting of coefficients (real numbers) multiplied with appropriate hat functions. 
Roughly speaking, we apply~\cref{eq:factorize_lin_interpolation} in \cref{factorize_lin_interpolation} in the situation where the left hand side of~\cref{eq:factorize_lin_interpolation} are linear interpolations of \MLP\ approximations and where we then want to approximate the right hand side of~\cref{eq:factorize_lin_interpolation} through \ANN\ approximations to thereby obtain \ANN\ approximations for \MLP\ approximations. 

Taking this into account, in \cref{ANN_for_hatfct1} (\ReLU\ and leaky \ReLU\ activations) and \cref{ANN_for_hatfct2} (softplus activation) in \cref{subsec:hat_fct} we study \ANN\ approximations for hat functions (which appear on the right hand side of~\cref{eq:factorize_lin_interpolation}) 
and in \cref{ANN_for_product1}, \cref{ANN_for_product2}, \cref{ANN_for_product12}, \cref{ANN_for_product3} (\ReLU\ and leaky \ReLU\ activations), and \cref{ANN_for_product4} (softplus activation) in \cref{subsec:approx_squarefct,subsec:approx_product} we study \ANN\ approximations for the production function $\R^2 \ni (v,w) \mapsto v w \in \R$ (which appears on the right hand side of~\cref{eq:factorize_lin_interpolation} to present the products (the multiplications) of the coefficients with the hat functions on the right hand side of~\cref{eq:factorize_lin_interpolation}). 
In the situation of the \ReLU\ activation a result similar to \cref{ANN_for_hatfct1} has been established in Grohs et al.~\cite[Lemma~3.9]{GrohsHornungJentzen2019}. 
Our overall approach in this section to employ~\cref{eq:factorize_lin_interpolation} in \cref{factorize_lin_interpolation} to approximate suitable linear interpolations through \ANN\ approximations is strongly inspired by the approach in Grohs et al.~\cite[Section~3]{GrohsHornungJentzen2019}.

\subsection{Properties of linear interpolation functions}
\label{subsec:lin_interpolation}

\cfclear
\begin{definition}[Linear interpolation function]
\label{def:lin_interp}
Let 
$K\in\N$, 
$\fx_0,\fx_1,\dots,\fx_K, f_0, f_1, \dots, f_K \in\R$ 
satisfy 
$\fx_0 < \fx_1 < \ldots < \fx_K$. 
Then we denote by 
$\scrL_{\fx_0,\fx_1,\dots,\fx_K}^{f_0,f_1,\dots,f_K} \colon \R \to \R$ 
the function which satisfies for all 
$k\in\{1,2,\dots,K\}$, 
$x\in(-\infty,\fx_0)$, 
$y\in[\fx_{k-1},\fx_k)$, 
$z\in[\fx_K,\infty)$ 
that 
$\scrL_{\fx_0,\fx_1,\dots,\fx_K}^{f_0,f_1,\dots,f_K}(x) = f_0$, 
$\scrL_{\fx_0,\fx_1,\dots,\fx_K}^{f_0,f_1,\dots,f_K}(z) = f_K$, 
and
\begin{equation}\label{eq:lin_interp}
\scrL_{\fx_0,\fx_1,\dots,\fx_K}^{f_0,f_1,\dots,f_K}(y) 
= 
f_{k-1} + \bpr{ \tfrac{y - \fx_{k-1}}{\fx_k - \fx_{k-1}} } (f_k - f_{k-1})
. 
\end{equation}
\end{definition}

\cfclear
\begin{athm}{lemma}{lin_interp1}
Let $K\in\N$, $\fx_0,\fx_1,\dots,\fx_K, f_0, f_1, \dots, f_K \in\R$ satisfy $\fx_0 < \fx_1 < \ldots < \fx_K$. Then 
it holds for all $k\in\{1,2,\dots,K\}$, $t\in[\fx_{k-1},\fx_k]$ that
\begin{equation}\label{lin_interp1_item4}
\babs{\scrL_{\fx_0,\fx_1,\dots,\fx_K}^{f_0,f_1,\dots,f_K}(t) - f_k} 
\le \abs{f_k - f_{k-1}} 
\end{equation}
\cfout.
\end{athm}

\cfclear
\begin{aproof}
    \Nobs that \cref{eq:lin_interp} 
    implies that for all $k\in\{1,2,\dots,K\}$, $t\in [\fx_{k-1},\fx_k]$
    it holds that 
    \begin{equation}\llabel{eq:1705}
        \begin{split}
            \scrL_{\fx_0,\fx_1,\dots,\fx_K}^{f_0,f_1,\dots,f_K}(t) - f_k
            & = f_{k-1} + \bbpr{\frac{t-\fx_{k-1}}{\fx_k-\fx_{k-1}}} \bpr{f_k-f_{k-1}} - f_k 
            = \bbpr{\frac{t-\fx_{k}}{\fx_k-\fx_{k-1}}} \bpr{f_k-f_{k-1}} 
        \end{split}
    \end{equation}
    \cfload. 
    \argument{\lref{eq:1705}; the fact that for all $k\in\{1,2,\dots,K\}$, $t\in [\fx_{k-1},\fx_k]$ it holds that $\abs{\tfrac{t-\fx_{k}}{\fx_k-\fx_{k-1}}}\le 1$}{ \cref{lin_interp1_item4}\dott{} }
\end{aproof}

\cfclear
\begin{athm}{lemma}{lin_interpol_hat_fct}
	Let  
	$\fx_{\indexkmin}, \fx_{\indexk}, \fx_{\indexkplus} \in \R$ 
	satisfy 
	$\fx_{\indexkmin} < \fx_{\indexk} < \fx_{\indexkplus}$. 
	Then it holds for all $t \in \R$ that 
	\begin{equation}\label{eq:def_hatfct}
		\scrL_{\fx_{\indexkmin}, \fx_{\indexk}, \fx_{\indexkplus}}^{0,1,0}(t)
		= \frac{t-\fx_{\indexkmin}}{\fx_{\indexk}-\fx_{\indexkmin}} \1_{(\fx_{\indexkmin},\fx_{\indexk}]}(t) 
		+ \frac{\fx_{\indexkplus}-t}{\fx_{\indexkplus}-\fx_{\indexk}} \1_{(\fx_{\indexk},\fx_{\indexkplus})}(t) 
	\end{equation}
    \cfout.
\end{athm}

\cfclear
\begin{aproof}
    \Nobs that \cref{eq:lin_interp} establishes \cref{eq:def_hatfct}. 
\end{aproof}

\cfclear
\begin{athm}{lemma}{factorize_lin_interpolation}
Let 
$K\in\N$, $T\in(0,\infty)$, 
$\fx_{-1},\fx_0,\fx_1,\dots,\fx_K, \fx_{K+1}, f_0, f_1, \dots, f_K \in\R$ 
satisfy 
$\fx_{-1}<0=\fx_0 < \fx_1 < \ldots < \fx_K = T < \fx_{K+1}$. 
Then it holds for all $t\in[0,T]$ that 
\begin{equation}\label{eq:factorize_lin_interpolation}
\scrL_{\fx_0,\fx_1,\dots,\fx_K}^{f_0,f_1,\dots,f_K}(t)
= \sum_{k=0}^K \scrL_{\fx_{k-1},\fx_k,\fx_{k+1}}^{0,1,0}(t) f_k 
\end{equation}
\cfout.
\end{athm}

\cfclear
\begin{aproof}
    \Nobs that  
    \cref{lin_interpol_hat_fct} and
    \eg
    items~(i) and~(iii) in Lemma~4.6 in~\cite{ackermann2023deep} 
    show that for all $t\in[0,T]$ it holds that 
    \begin{equation}
        \begin{split}
        \sum_{k=0}^K \scrL_{\fx_{k-1},\fx_k,\fx_{k+1}}^{0,1,0}(t) f_k 
        & = \1_{\{\fx_0\}}(t) f_0 
        + \sum_{k=1}^K \frac{t-\fx_{k-1}}{\fx_k-\fx_{k-1}} \1_{(\fx_{k-1},\fx_k]}(t) f_k
        + \sum_{k=0}^{K-1} \frac{\fx_{k+1}-t}{\fx_{k+1}-\fx_{k}} \1_{(\fx_{k},\fx_{k+1})}(t) f_k \\
        & = \1_{\{\fx_0\}}(t) f_0 
        + \sum_{k=1}^K \frac{t-\fx_{k-1}}{\fx_k-\fx_{k-1}} \1_{(\fx_{k-1},\fx_k]}(t) f_k
        + \sum_{k=1}^{K} \frac{\fx_{k}-t}{\fx_{k}-\fx_{k-1}} \1_{(\fx_{k-1},\fx_{k})}(t) f_{k-1} \\
        & = \1_{\{\fx_0\}}(t) f_0 
        + \sum_{k=1}^K \bbbpr{ \frac{t-\fx_{k-1}}{\fx_k-\fx_{k-1}} f_k + \frac{\fx_{k}-t}{\fx_{k}-\fx_{k-1}} f_{k-1} } \1_{(\fx_{k-1},\fx_k]}(t)  \\
        & = \1_{\{\fx_0\}}(t) \scrL_{\fx_0,\fx_1,\dots,\fx_K}^{f_0,f_1,\dots,f_K}(t) 
        + \sum_{k=1}^K \scrL_{\fx_0,\fx_1,\dots,\fx_K}^{f_0,f_1,\dots,f_K}(t) \1_{(\fx_{k-1},\fx_k]}(t) \\
        & = \scrL_{\fx_0,\fx_1,\dots,\fx_K}^{f_0,f_1,\dots,f_K}(t) 
        \end{split}
    \end{equation}
    \cfload. 
\end{aproof}

\subsection{ANN representations and approximations for hat functions}
\label{subsec:hat_fct}

\cfclear
\begin{athm}{lemma}{ANN_for_hatfct1}
    Let $\fx_{\indexkmin}, \fx_{\indexk}, \fx_{\indexkplus} \in \R$ 
    satisfy 
    $\fx_{\indexkmin} < \fx_{\indexk} < \fx_{\indexkplus}$, 
    let $c_{\indexkmin},c_{\indexk},c_{\indexkplus} \in \R$ 
    satisfy 
    \begin{equation}
        c_{\indexkmin} = \frac{1}{\fx_{\indexk}-\fx_{\indexkmin}}, 
        \qquad 
        c_{\indexk} = -\frac{1}{\fx_{\indexkplus}-\fx_{\indexk}} - \frac{1}{\fx_{\indexk} - \fx_{\indexkmin}}, 
        \qquad \text{and} 
        \qquad 
        c_{\indexkplus} = \frac{1}{\fx_{\indexkplus}-\fx_{\indexk}}, 
    \end{equation}
    let $\leaky \in \R\backslash\{-1,1\}$, 
    let $a \colon \R\to\R$ satisfy for all $x \in \R$ that $a(x)=\max\{x,\leaky x\}$, 
    and let $\H \in \ANNsm$ 
    be given by 
    \begin{equation}
        \begin{split}
            \H & = 
            \bbbbr{
            \OSum{j=\indexkmin}{\indexkplus} 
            \bbbpr{
            \scalar{\bbpr{\frac{\abs{1-\leaky} \leaky c_j}{(1-\leaky)(1-\leaky^2)}}}{
            \bbpr{ \compANN{\ii_1}{\A_{\frac{-\abs{1-\leaky}}{1-\leaky}, \frac{\abs{1-\leaky}\fx_j}{1-\leaky} }} } }}
            } \\
            & \quad \oSum
            \bbbbr{
            \OSum{j=\indexkmin}{\indexkplus} 
            \bbbpr{
            \scalar{\bbpr{\frac{\abs{1-\leaky} c_j}{(1-\leaky)(1-\leaky^2)}}}{
            \bbpr{ \compANN{\ii_1}{\A_{\frac{\abs{1-\leaky}}{1-\leaky}, \frac{-\abs{1-\leaky}\fx_j}{1-\leaky} }} } }}
            }
        \end{split}
    \end{equation}    
    \cfload.
    Then 
    \begin{enumerate}[label=(\roman *)]
    \item
    \label{ANN_for_hatfct1_item1}
    it holds that $\Ra(\H) \in  C(\R,\R)$,
    \item
    \label{ANN_for_hatfct1_item2}
    it holds for all $t\in\R$ that
    $(\Ra(\H))(t) = \scrL_{\fx_{\indexkmin}, \fx_{\indexk}, \fx_{\indexkplus}}^{0,1,0}(t)$, 
    \item
    \label{ANN_for_hatfct1_item3}
    it holds that $\dims(\H)=(1,6,1)$,
    and 
    \item
    \label{ANN_for_hatfct1_item4}
    it holds that $\paramANN(\H) =19$
    \end{enumerate}
    \cfout. 
\end{athm}

\cfclear
\begin{aproof}
    Let $\fr \in C(\R,\R)$ satisfy for all $x \in \R$ that $\fr(x)=\max\{x,0\}$ and let $\F \in \ANNsm$ 
    be given by 
    \begin{equation}
        \F = 
        \compANN{\A_{1,0}}{
           \OSum{j=\indexkmin}{\indexkplus} 
            \bbpr{
            \scalar{c_j}{
            \bpr{ \compANN{\ii_1}{\A_{1,-\fx_j}} }
            }
            } 
        }
    \end{equation}
    \cfload. 
    \startnewargseq
    \argument{\cite[Lemma~4.10]{ackermann2023deep}}{ 
    that for all $t \in \R$ it holds that 
    \begin{equation}\label{eq:1005a}
        (\Rr(\F))(t) = \scrL_{\fx_{\indexkmin}, \fx_{\indexk}, \fx_{\indexkplus}}^{0,1,0}(t) 
        \dott
    \end{equation}
    }
    \argument{\eg item~(iii) in Lemma~4.9 in~\cite{ackermann2023deep}; \eg \cite[Lemma~3.2]{ackermann2023deep}}{
    that for all $t\in\R$ it holds that 
    \begin{equation}\llabel{eq:1005b}
        (\Rr(\F))(t) = \sum_{j=\indexkmin}^{\indexkplus} c_j \fr(t-\fx_j)
        =  \sum_{j=\indexkmin}^{\indexkplus} c_j (\Rr(\ii_1))(t-\fx_j)
        \dott 
    \end{equation}
    }
    \argument{\eg \cite[Lemma~2.19]{ackermann2023deep}
    }{
    that for all $t\in\R$ it holds that 
    $\Ra(\H) \in C(\R,\R)$ and 
    \begin{equation}\llabel{eq:1005c}
        \begin{split}
            & (\Ra(\H))(t) \\
            & = 
            \bbbbr{\sum_{j=\indexkmin}^{\indexkplus}
            \bbbpr{\frac{\abs{1-\leaky} \leaky c_j}{(1-\leaky)(1-\leaky^2)}}
            \bpr{\Ra(\ii_1)}
            \bbbpr{\frac{-\abs{1-\leaky}t}{1-\leaky}+\frac{\abs{1-\leaky} \fx_j}{1-\leaky}}
            } \\
            & \quad + 
            \bbbbr{\sum_{j=\indexkmin}^{\indexkplus} 
            \bbbpr{\frac{\abs{1-\leaky} c_j}{(1-\leaky)(1-\leaky^2)}}
            \bpr{\Ra(\ii_1)}
            \bbbpr{\frac{\abs{1-\leaky}t}{1-\leaky}+\frac{-\abs{1-\leaky} \fx_j}{1-\leaky}}
            } \\
            & = 
            \sum_{j=\indexkmin}^{\indexkplus}
            \frac{\abs{1-\leaky} c_j }{(1-\leaky)(1-\leaky^2)} 
            \bbbbr{
            \leaky 
            \bpr{\Ra(\ii_1)} \bbbpr{\frac{-\abs{1-\leaky}(t-\fx_j)}{1-\leaky}} 
            +
            \bpr{\Ra(\ii_1)} \bbbpr{\frac{\abs{1-\leaky}(t-\fx_j)}{1-\leaky}}
            }
            \dott
        \end{split}
    \end{equation}
    }
    \argument{\cref{lemma_relation_leakyReLU_ReLU};  \cref{eq:1005a}; \lref{eq:1005b}; \lref{eq:1005c}}{ 
    that for all $t\in\R$ it holds that 
    \begin{equation}\llabel{eq:0937}
        (\Ra(\H))(t)
        = \sum_{j=\indexkmin}^{\indexkplus} c_j (\Rr(\ii_1))(t-\fx_j)
        = (\Rr(\F))(t)
        = \scrL_{\fx_{\indexkmin}, \fx_{\indexk}, \fx_{\indexkplus}}^{0,1,0}(t) 
        \dott 
    \end{equation}
    }
    \argument{\lref{eq:0937}}{
    \cref{ANN_for_hatfct1_item1,ANN_for_hatfct1_item2}\dott{}
    }
    \startnewargseq
    \argument{\eg item~(i) in Lemma~2.19 in~\cite{ackermann2023deep}; \eg item~(i) in Lemma~3.2 in~\cite{ackermann2023deep}}{
    that 
    \begin{equation}\llabel{eq:1117}
        \dims(\H)
        = \bpr{ \inDimANN(\ii_1), 6\dimANNlevel_1(\ii_1) ,\outDimANN(\ii_1)  }
        = (1,6,1)
        \dott 
    \end{equation}
    }
    \argument{\lref{eq:1117}}{
    \cref{ANN_for_hatfct1_item3}\dott{}
    }
    \startnewargseq
    \argument{\cref{ANN_for_hatfct1_item3}}{
    that 
    \begin{equation}\llabel{eq:0940}
        \paramANN(\H) = 6(1+1) +(6+1) = 19
        \dott 
    \end{equation}
    }
    \argument{\lref{eq:0940}}{
    \cref{ANN_for_hatfct1_item4}\dott{}
    }
\end{aproof}

\cfclear
\begin{athm}{lemma}{ANN_for_hatfct2}
    Let $\fx_{\indexkmin}, \fx_{\indexk}, \fx_{\indexkplus} \in \R$ 
    satisfy 
    $\fx_{\indexkmin} < \fx_{\indexk} < \fx_{\indexkplus}$, 
    let $\varepsilon \in (0,1]$, $q \in (1,\infty)$, 
    $\fL = \max\{ \abs{\fx_{\indexk}-\fx_{\indexkmin}}^{-1}, \abs{\fx_{\indexkplus}-\fx_{\indexk}}^{-1} \}$ 
    and let $a\colon \R\to\R$ satisfy for all $x \in \R$ that $a(x)=\ln(1+\exp(x))$\cfload.
    Then there exists $\H \in \ANNsm$ such that for all $t \in \R$ it holds that 
    \begin{equation}\label{eq:ANN_for_hatfct2_properties}
        \begin{split}
            & \Ra(\H) \in C(\R,\R), \qquad 
            \paramANN(\H) \le
            12 \bpr{\max\cu{1,4\fL}}^{q/(q-1)} 2^{q/(q-1)} \varepsilon^{-q/(q-1)}, \\
            & \text{and} \qquad 
            \babs{ (\Ra(\H))(t) - \scrL_{\fx_{\indexkmin}, \fx_{\indexk}, \fx_{\indexkplus}}^{0,1,0}(t) }
            \le \varepsilon \max\{1,\abs{t}^q\}
        \end{split}
    \end{equation}
    \cfout. 
\end{athm}

\cfclear
\begin{aproof}
    Let $f\colon [\fx_{\indexkmin},\fx_{\indexkplus}] \to \R$ satisfy for all $t \in [\fx_{\indexkmin},\fx_{\indexkplus}]$ that 
    \begin{equation}\llabel{eq:1715}
        f(t)=\scrL_{\fx_{\indexkmin},\fx_{\indexk},\fx_{\indexkplus}}^{0,1,0}(t) 
    \end{equation}
    \cfload. 
    \startnewargseq
    \argument{\lref{eq:1715}; \eg \cite[item~(i) in Lemma~4.6]{ackermann2023deep}}{
    that 
    $f(\fx_{\indexkmin})=0$, $f(\fx_{\indexk})=1$, and $f(\fx_{\indexkplus})=0$\dott \,
    }
    \argument{the triangle inequality; \cref{eq:def_hatfct}}{
    that for all $t,s \in [\fx_{\indexkmin},\fx_{\indexkplus}]$ it holds that 
    \begin{equation}\llabel{eq:1657}
        \begin{split}
            \abs{ f(t) - f(s) } 
            & \le \abs{f(t)} + \abs{f(s)}
            \le 1+1=2 
            \dott
        \end{split}
    \end{equation}
    }
    \argument{\lref{eq:1657}; \eg item~(i) in Lemma~4.7 in~\cite{ackermann2023deep}}{
    that for all $t,s\in\R$ it holds that 
    \begin{equation}\llabel{eq:1703}
        \begin{split}
            \babs{
            \scrL_{\fx_{\indexkmin},\fx_{\indexk},\fx_{\indexkplus}}^{0,1,0}(t) 
            - \scrL_{\fx_{\indexkmin},\fx_{\indexk},\fx_{\indexkplus}}^{0,1,0}(s)
            } 
            & = \babs{
            \scrL_{\fx_{\indexkmin},\fx_{\indexk},\fx_{\indexkplus}}^{f(\fx_{\indexkmin}),f(\fx_{\indexk}),f(\fx_{\indexkplus})}(t) 
            - \scrL_{\fx_{\indexkmin},\fx_{\indexk},\fx_{\indexkplus}}^{f(\fx_{\indexkmin}),f(\fx_{\indexk}),f(\fx_{\indexkplus})}(s)
            } \\
            & \le \max\bbbcu{ \frac{2}{\abs{\fx_{\indexk}-\fx_{\indexkmin}}},  \frac{2}{\abs{\fx_{\indexkplus}-\fx_{\indexk}}} } 
            \abs{t-s} 
            = 2 \fL \abs{t-s}
            \dott 
        \end{split}
    \end{equation}
    }
    \argument{\lref{eq:1703}; \cite[Corollary~4.14]{ackermann2023deep} (applied with 
    $\varepsilon \with 2^{-1} \varepsilon$, 
    $L \with 2 \fL$,
    $q\with q$, 
    $f \with \scrL_{\fx_{\indexkmin},\fx_{\indexk},\fx_{\indexkplus}}^{0,1,0}$, 
    $a\with a$ 
    in the notation of \cite[Corollary~4.14]{ackermann2023deep})}{
    that there exists 
    $\H \in \ANNsm$ such that 
    \cref{eq:ANN_for_hatfct2_properties} is satisfied\dott 
    }
\end{aproof}

\subsection{ANN approximations for the square function}
\label{subsec:approx_squarefct}

\cfclear
\begin{athm}{lemma}{ANN_for_product1}
    Let $f\colon \R \to \R$ satisfy for all $x \in [0,1]$, $y \in \R\backslash[0,1]$ that $f(x)=x^2$ and $f(y)=\max\{y,0\}$, 
    let $\varepsilon \in (0,1]$, $q\in(2,\infty)$, 
    $\delta \in \R$ satisfy $\delta = 2^{-1} 4^{-2/(q-2)} \varepsilon^{q/(q-2)}$, 
    let $a\in C(\R,\R)$, $\G \in \ANNsm$, $r, c \in[1,\infty)$ satisfy for all $x \in \R$ that 
    $\Ra(\G) \in C(\R,\R)$, $\paramANN(\G) \le c \delta^{-r}$, and 
    \begin{equation}\label{eq:ANN_for_product1_1}
        \begin{split}
            & \abs{ \pr{\Ra(\G)}(x) - f(x) } \le \delta \max\{1,\abs{x}^q\},
        \end{split}
    \end{equation}
    let $W_1 \in \R^{2\times 1}$, $W_2 \in \R^{1\times 2}$ 
    be given by 
    \begin{equation}
        W_1 = 
        \begin{pmatrix}
            \bpr{\frac{\varepsilon}{4}}^{1/(q-2)} \\
            - \bpr{\frac{\varepsilon}{4}}^{1/(q-2)}
        \end{pmatrix}
        \qquad \text{and} \qquad 
        W_2 = 
        \begin{pmatrix}
            \bpr{\frac{\varepsilon}{4}}^{-2/(q-2)}
            & \bpr{\frac{\varepsilon}{4}}^{-2/(q-2)}
        \end{pmatrix}, 
    \end{equation}
    and let $\Phi \in \ANNsm$ 
    be given by 
    \begin{equation}
        \Phi = \compANN{\A_{W_2,0}}{\compANN{  
        \parallelizationSpecial_{2}(\G,\G)
        }{\A_{W_1,0}}}
    \end{equation}
    \cfload.
    Then 
    \begin{enumerate}[label=(\roman *)]
    \item
    \label{ANN_for_product1_item1}
    it holds that $\Ra(\Phi) \in C(\R,\R)$,
    \item
    \label{ANN_for_product1_item2}
    it holds for all $x \in \R$ that 
    $\abs{\pr{\Ra(\Phi)}(x) - x^2 } \le \varepsilon \max\{1,\abs{x}^q\}$,
    and 
    \item
    \label{ANN_for_product1_item3}
    it holds that $\paramANN(\Phi)\le 2^{r+2} 4^{2r/(q-2)} c \varepsilon^{-rq/(q-2)}$ 
    \cfout.
    \end{enumerate}
\end{athm}

\cfclear
\begin{aproof}
    Throughout this proof let $\tilde\varepsilon, \tilde q \in \R$ satisfy $\tilde \varepsilon = \varepsilon/4$ and $\tilde q = 1/(q-2)$. 
    \startnewargseq
    \argument{\eg \cite[Proposition~2.19]{GrohsHornungJentzen2019}; \eg item~(vi) in Proposition~2.6 in~\cite{GrohsHornungJentzen2019}}{
    that for all $x \in\R$ it holds that 
    $\Ra(\Phi) \in C(\R,\R)$ and 
    \begin{equation}\llabel{eq:1024}
        \begin{split}
            \bpr{\Ra(\Phi)}(x)
            & = \bpr{ \bpr{\Ra(\A_{W_2,0})} \circ \bpr{\Ra(\parallelizationSpecial_2(\G,\G))} \circ \bpr{\Ra(\A_{W_1,0})} }(x) \\
            & = \bpr{\Ra(\A_{W_2,0})} 
            \bpr{ \bpr{\Ra(\parallelizationSpecial_2(\G,\G))}\bpr{ \tilde\varepsilon^{\tilde q} x, - \tilde\varepsilon^{\tilde q} x } } \\
            & = \bpr{\Ra(\A_{W_2,0})} 
            \bpr{ \pr{\Ra(\G)}\bpr{\tilde\varepsilon^{\tilde q} x},  \pr{\Ra(\G)}\bpr{ - \tilde\varepsilon^{\tilde q} x} } \\
            & = \tilde\varepsilon^{-2\tilde q}  
            \bpr{
            \pr{\Ra(\G)}\bpr{\tilde\varepsilon^{\tilde q} x}
            + \pr{\Ra(\G)}\bpr{-\tilde\varepsilon^{\tilde q} x}
            }
            \dott 
        \end{split}
    \end{equation}
    } 
    \argument{\lref{eq:1024}}{
    \cref{ANN_for_product1_item1}\dott{}
    }
    \startnewargseq
    \argument{the fact that for all $x\in\R\backslash[0,1]$ it holds that $f(x)=\max\cu{x,0}$}{
    that for all $x\in\R\backslash[-1,1]$ it holds that 
    \begin{equation}\llabel{eq:1025}
        \begin{split}
            \abs{x}
            & = \max\{x,0\} - \min\{x,0\}
            = \max\{x,0\} + \max\{-x,0\} 
            = f(x) + f(-x) 
            \dott 
        \end{split}
    \end{equation}
    }
    \argument{\lref{eq:1025}; the triangle inequality; \cref{eq:ANN_for_product1_1}}{
    that for all $x \in \R\backslash[-1,1]$ it holds that 
    \begin{equation}\llabel{eq:1038}
        \begin{split}
            \babs{ \pr{\Ra(\G)}(x) + \pr{\Ra(\G)}(-x) - \abs{x} } 
            & \le 
            \babs{ \pr{\Ra(\G)}(x) - f(x) }
            +  \babs{ \pr{\Ra(\G)}(-x) - f(-x) } \\
            & \le \delta \max\{1,\abs{x}^q\} + \delta \max\{1,\abs{-x}^q\} 
            = 2 \delta \abs{x}^q 
            \dott 
        \end{split}
    \end{equation}
    }
    \argument{\lref{eq:1024}; \lref{eq:1038}}{
    that for all $x\in\R\backslash[-\tilde\varepsilon^{-\tilde q},\tilde\varepsilon^{-\tilde q}]$ it holds that 
    \begin{equation}\llabel{eq:1048}
        \begin{split}
            \babs{ \pr{\Ra(\Phi)}(x) - \tilde\varepsilon^{-\tilde q} \abs{x} } 
            & = \babs{ \tilde\varepsilon^{-2\tilde q} \bpr{ \bpr{
            \pr{\Ra(\G)}\bpr{\tilde\varepsilon^{\tilde q} x}
            + \pr{\Ra(\G)}\bpr{-\tilde\varepsilon^{\tilde q} x}
            }
            - \abs{\tilde\varepsilon^{\tilde q} x} 
            } } \\
            & \le \tilde\varepsilon^{-2\tilde q}
            2 \delta \abs{\tilde\varepsilon^{\tilde q}x}^q
            = 2\delta \tilde\varepsilon \abs{x}^q 
            \dott 
        \end{split}
    \end{equation}
    }
    \argument{the triangle inequality}{
    that for all $x\in\R\backslash[-\tilde\varepsilon^{-\tilde q},\tilde\varepsilon^{-\tilde q}]$ it holds that
    \begin{equation}\llabel{eq:1059}
        \begin{split}
            \babs{ \tilde\varepsilon^{-\tilde q} \abs{x} -x^2}
            & \le 
            \tilde\varepsilon^{-\tilde q} \abs{x} + \abs{x}^2
            = 
            \abs{x}^q \bpr{
            \tilde\varepsilon^{-\tilde q} \abs{x}^{-(q-1)} + \abs{x}^{-(q-2)}
            } \\
            & \le 
            \abs{x}^q \bpr{
            \tilde\varepsilon^{-\tilde q} \tilde\varepsilon^{(q-1)\tilde q} + 
            \tilde\varepsilon^{(q-2)\tilde q}
            } 
            = 2 \abs{x}^q \tilde\varepsilon
            \dott 
        \end{split}
    \end{equation}
    }
    \argument{\lref{eq:1048};\lref{eq:1059};the triangle inequality; the fact that $\delta \in (0,1]$}{
    that for all $x\in\R\backslash[-\tilde\varepsilon^{-\tilde q},\tilde\varepsilon^{-\tilde q}]$ it holds that
    \begin{equation}\llabel{eq:1107}
        \begin{split}
            \babs{ \pr{\Ra(\Phi)}(x) - x^2 }
            & \le 
            \babs{ \pr{\Ra(\Phi)}(x) - \tilde\varepsilon^{-\tilde q} \abs{x} } 
            + \babs{ \tilde\varepsilon^{-\tilde q} \abs{x} -x^2} \\
            & \le 2\delta \tilde\varepsilon \abs{x}^q  + 
            2 \abs{x}^q \tilde\varepsilon 
            \le 4 \tilde \varepsilon \abs{x}^q 
            \le \varepsilon \max\{1,\abs{x}^q\}
            \dott 
        \end{split}
    \end{equation}
    }
    \argument{the triangle inequality; \cref{eq:ANN_for_product1_1}}{
    that for all $x\in [0,\infty)$ it holds that 
    \begin{equation}\llabel{eq:1112}
        \begin{split}
            \abs{ \pr{\Ra(\G)}(-x) }
            & \le 
            \abs{ \pr{\Ra(\G)}(-x) - f(-x) } 
            + \abs{f(-x)} 
            \\
            & \le \delta \max\{1,\abs{-x}^q\} + 0 
            = \delta \max\{1,\abs{x}^q \} 
            \dott 
        \end{split}
    \end{equation}
    }
    \argument{the triangle inequality; \cref{eq:ANN_for_product1_1};\lref{eq:1112}}{
    that for all $x \in [0,1]$ it holds that 
    \begin{equation}\llabel{eq:1118}
        \begin{split}
            \babs{ \pr{\Ra(\G)}(x) + \pr{\Ra(\G)}(-x) - x^2 } 
            & = \babs{ \pr{\Ra(\G)}(x) + \pr{\Ra(\G)}(-x) - f(x)} \\
            & \le 
            \babs{ \pr{\Ra(\G)}(x) - f(x) } + \babs{ \pr{\Ra(\G)}(-x) } \\
            & \le \delta \max\{1,\abs{x}^q \} + \delta \max\{1,\abs{x}^q \} 
            = 2 \delta   
            \dott 
        \end{split}
    \end{equation}
    }
    \argument{\lref{eq:1024};\lref{eq:1118}}{
    that for all $x \in [-\tilde\varepsilon^{-\tilde q},\tilde\varepsilon^{-\tilde q}]$ it holds that 
    \begin{equation}\llabel{eq:0944}
        \begin{split}
            \abs{ \pr{\Ra(\Phi)}(x) - x^2 }
            & \le 
            \tilde\varepsilon^{-2\tilde q} 
            \babs{ 
            \pr{\Ra(\G)}\bpr{\tilde\varepsilon^{\tilde q} x}
            + \pr{\Ra(\G)}\bpr{-\tilde\varepsilon^{\tilde q} x}
            - \bpr{\tilde\varepsilon^{\tilde q} x}^2
            } \\
            & \le 
            \tilde\varepsilon^{-2\tilde q}  2 \delta 
            = \varepsilon^{-2\tilde q} \cdot 4^{2\tilde q} 
            \cdot 2 \cdot 2^{-1} \cdot 4^{-2\tilde q} \cdot \varepsilon^{q\tilde q}
            = \varepsilon \\
            & \le \varepsilon \max\{1,\abs{x}^q\}
            \dott 
        \end{split}
    \end{equation}
    }
    \argument{\lref{eq:0944}; \lref{eq:1107}}{
    \cref{ANN_for_product1_item2}\dott{}
    }
    \startnewargseq
    \argument{\eg \cite[Corollary~2.9]{GrohsHornungJentzen2019}}{
    that 
    \begin{equation}\llabel{eq:1149}
        \begin{split}
            \paramANN\bpr{\compANN{\A_{W_2,0}}{\compANN{  
        \parallelizationSpecial_{2}(\G,\G)
        }{\A_{W_1,0}}}}
            & \le 
            \paramANN\bpr{\compANN{\A_{W_2,0}}{  
        \parallelizationSpecial_{2}(\G,\G)
        } }
        \le \paramANN\bpr{ \parallelizationSpecial_{2}(\G,\G)
        } 
        \dott
        \end{split}
    \end{equation}
    }
    \argument{\lref{eq:1149}; the fact that $\paramANN(\G)\le c \delta^{-r}$; \eg \cite[Corollary~2.21]{GrohsHornungJentzen2019}}{
    that 
    \begin{equation}\llabel{eq:0946}
        \begin{split}
            \paramANN(\Phi)
            & \le 
            \paramANN\bpr{ \parallelizationSpecial_{2}(\G,\G) }
            \le 
            4 \paramANN(\G) 
            \le 4 c \delta^{-r} 
            = 4 c 2^r 4^{2r/(q-2)} \varepsilon^{-rq/(q-2)} 
            \dott 
        \end{split}
    \end{equation}
    }
    \argument{\lref{eq:0946}}{
    \cref{ANN_for_product1_item3}\dott{}
    }
\end{aproof}

\subsection{ANN approximations for the product function}
\label{subsec:approx_product}

\cfclear
\begin{athm}{lemma}{ANN_for_product2}
    Let $\varepsilon \in (0,1]$, $q\in(2,\infty)$, 
    $\delta \in\R$ satisfy $\delta = (2^{q-1}+1)^{-1} \varepsilon$, 
    let $a\in C(\R,\R)$, $\Phi \in \ANNsm$, $r, c \in [1,\infty)$ satisfy for all $x \in \R$ that 
    $\Ra(\Phi) \in C(\R,\R)$, $\paramANN(\Phi) \le c \delta^{-r}$, and 
    \begin{equation}\label{eq:ANN_for_product2_1}
        \begin{split}
            & \abs{ \pr{\Ra(\Phi)}(x) - x^2 } \le \delta \max\{1,\abs{x}^q\},
        \end{split}
    \end{equation}
    let $W_1 \in \R^{3\times 2}$, $W_2 \in \R^{1\times 3}$ 
    be given by 
    \begin{equation}
        W_1 = 
        \begin{pmatrix}
            1 & 1 \\
            1 & 0 \\
            0 & 1 
        \end{pmatrix}
        \qquad \text{and} \qquad 
        W_2 = 
        \begin{pmatrix}
            \frac12 & -\frac12 & -\frac12 
        \end{pmatrix}, 
    \end{equation}
    and let $\Gamma \in \ANNsm$ 
    be given by 
    \begin{equation}
        \Gamma = \compANN{\A_{W_2,0}}{\compANN{  
        \parallelizationSpecial_{3}(\Phi,\Phi,\Phi)
        }{\A_{W_1,0}}}
    \end{equation}
    \cfload.
    Then 
    \begin{enumerate}[label=(\roman *)]
    \item
    \label{ANN_for_product2_item1}
    it holds that $\Ra(\Gamma) \in C(\R^2,\R)$,
    \item
    \label{ANN_for_product2_item2}
    it holds for all $x,y \in \R$ that 
    $\abs{\pr{\Ra(\Gamma)}(x,y) - xy } \le \varepsilon \max\{1,\abs{x}^q, \abs{y}^q\}$,
    and 
    \item
    \label{ANN_for_product2_item3}
    it holds that $\paramANN(\Gamma)\le 9 c (2^{q-1}+1)^r \varepsilon^{-r}$ 
    \cfout.
    \end{enumerate}
\end{athm}

\cfclear
\begin{aproof}
    \Nobs that \eg \cite[Proposition~2.19]{GrohsHornungJentzen2019} and \eg item~(vi) in Proposition~2.6 in~\cite{GrohsHornungJentzen2019} 
    show that for all $x,y \in \R$ it holds that $\Ra(\Gamma) \in C(\R^2,\R)$ and 
    \begin{equation}\label{eq:repres_Ra_Gamma}
        \begin{split}
            \pr{ \Ra(\Gamma) }(x,y)
            & = 
            \bpr{ \bpr{\Ra(\A_{W_2,0})} \circ 
            \bpr{\Ra\bpr{\parallelizationSpecial_3(\Phi,\Phi,\Phi)}} \circ
            \bpr{\Ra(\A_{W_1,0})} }(x,y) \\
            & = 
            \bpr{\Ra(\A_{W_2,0})} 
            \bpr{
            \bpr{\Ra\bpr{\parallelizationSpecial_3(\Phi,\Phi,\Phi)}}
            \pr{x+y,x,y} 
            } \\
            & = 
            \bpr{\Ra(\A_{W_2,0})} 
            \bpr{
            \bpr{\Ra(\Phi)}(x+y), 
            \bpr{\Ra(\Phi)}(x), 
            \bpr{\Ra(\Phi)}(y)
            } \\
            & = \tfrac12 \bpr{\Ra(\Phi)}(x+y)
            - \tfrac12 \bpr{\Ra(\Phi)}(x)
            - \tfrac12 \bpr{\Ra(\Phi)}(y) 
        \end{split}
    \end{equation}
    \cfload. 
    This proves \cref{ANN_for_product2_item1}. 
    \startnewargseq
    \argument{\cref{eq:repres_Ra_Gamma}; the fact that for all $x,y\in\R$ it holds that $(x+y)^2-x^2-y^2=2xy$; the triangle inequality; \cref{eq:ANN_for_product2_1}}{
    that for all $x,y\in\R$ it holds that 
    \begin{equation}\llabel{eq:1235a}
        \begin{split}
            & 2\babs{ \bpr{\Ra(\Gamma)}(x,y) - xy } \\
            & 
            = 
            \babs{  
            \bpr{\Ra(\Phi)}(x+y) - (x+y)^2
            - \bpr{ \bpr{\Ra(\Phi)}(x) - x^2 }
            -  \bpr{ \bpr{\Ra(\Phi)}(y) - y^2 }
            } \\
            & \le 
            \babs{  
            \bpr{\Ra(\Phi)}(x+y) - (x+y)^2 }
            +
            \babs{  
            \bpr{\Ra(\Phi)}(x) - x^2 }
            +
            \babs{  
            \bpr{\Ra(\Phi)}(y) - y^2 } \\
            & \le 
            \delta \max\{1,\abs{x+y}^q\}
            + \delta \max\{1,\abs{x}^q\}
            + \delta \max\{1,\abs{y}^q\}
            \dott 
        \end{split}
    \end{equation}
    }
    \argument{Jensen's inequality}{
    that for all $x,y\in\R$ it holds that 
    \begin{equation}\llabel{eq:1235b}
        \begin{split}
            & \max\{1,\abs{x+y}^q\}
            + \max\{1,\abs{x}^q\}
            + \max\{1,\abs{y}^q\} \\
            & \le 2^{q-1} \max\{1,\abs{x}^q+\abs{y}^q\}
            + \max\{1,\abs{x}^q\}
            + \max\{1,\abs{y}^q\} \\
            & \le (2^{q-1}+1) 
            \bpr{
            \max\{1,\abs{x}^q\}
            + \max\{1,\abs{y}^q\}
            } \\
            & \le 2 (2^{q-1}+1) \max\{1,\abs{x}^q,\abs{y}^q\}
            \dott 
        \end{split}
    \end{equation}
    }
    \argument{\lref{eq:1235a};\lref{eq:1235b}}{
    that for all $x,y \in \R$ it holds that 
    \begin{equation}\llabel{eq:0949}
        \begin{split}
            \babs{ \bpr{\Ra(\Gamma)}(x,y) - xy }
            & \le 
            \delta (2^{q-1}+1) \max\{1,\abs{x}^q,\abs{y}^q\}
            = \varepsilon \max\{1,\abs{x}^q,\abs{y}^q\} 
            \dott 
        \end{split}
    \end{equation}
    }
    \argument{\lref{eq:0949}}{
    \cref{ANN_for_product2_item2}\dott{}
    }
    \startnewargseq
    \argument{\eg \cite[Corollary~2.9]{GrohsHornungJentzen2019}; \cite[Corollary~2.21]{GrohsHornungJentzen2019}; the fact that $\paramANN(\Phi)\le c \delta^{-r}$}{
    that 
    \begin{equation}\llabel{eq:0950}
        \begin{split}
            \paramANN(\Gamma)
            & = 
            \paramANN\bpr{ \compANN{\A_{W_2,0}}{\compANN{  
            \parallelizationSpecial_{3}(\Phi,\Phi,\Phi)
            }{\A_{W_1,0}}} } 
            \le \paramANN\bpr{ \compANN{\A_{W_2,0}}{\parallelizationSpecial_{3}(\Phi,\Phi,\Phi)} } \\
            & \le \paramANN\bpr{\parallelizationSpecial_{3}(\Phi,\Phi,\Phi)} 
            \le 9 \paramANN(\Phi) 
            \le 9 c \delta^{-r} 
            = 9c (2^{q-1}+1)^r \varepsilon^{-r}
            \dott 
        \end{split}
    \end{equation}
    }
    \argument{\lref{eq:0950}}{
    \cref{ANN_for_product2_item3}\dott{}
    }
\end{aproof}

\cfclear
\begin{athm}{corollary}{ANN_for_product12}
    Let $f\colon \R \to \R$ satisfy for all $x \in [0,1]$, $y \in \R\backslash[0,1]$ that $f(x)=x^2$ and $f(y)=\max\{y,0\}$, 
    let $\varepsilon \in (0,1]$, $q\in(2,\infty)$, 
    $r, c \in[1,\infty)$, 
    let $\delta \in \R$ satisfy $\delta = 2^{-1} 4^{-2/(q-2)} (2^{q-1}+1)^{-q/(q-2)} \varepsilon^{q/(q-2)}$, 
    let $a\in C(\R,\R)$, 
    and 
    let $\G \in \ANNsm$ satisfy for all $x \in \R$ that 
    $\Ra(\G) \in C(\R,\R)$, $\paramANN(\G) \le c \delta^{-r}$, and 
    \begin{equation}\label{eq:ANN_for_product12_1}
        \begin{split}
            & \abs{ \pr{\Ra(\G)}(x) - f(x) } \le \delta \max\{1,\abs{x}^q\}
        \end{split}
    \end{equation}
    \cfload.
    Then there exists $\Gamma \in \ANNsm$ such that for all $x,y \in \R$ it holds that 
    \begin{equation}
        \begin{split}
            & \Ra(\Gamma) \in C(\R^2,\R),
            \qquad 
            \paramANN(\Gamma)\le
            36 \cdot 2^{\tfrac{(q^2+q+2)r}{q-2}} c
            \varepsilon^{-\tfrac{rq}{q-2}}, \\
            & \text{and} \qquad 
            \abs{(\Ra(\Gamma))(x,y)-xy} \le \varepsilon \max\{1,\abs{x}^q,\abs{y}^q\} 
            \dott
        \end{split}
    \end{equation}
\end{athm}

\cfclear
\begin{aproof}
    \Nobs that \cref{ANN_for_product1} (applied with 
    $f\with f$, $\G \with \G$, $a\with a$, $q\with q$, 
    $\varepsilon \with (2^{q-1}+1)^{-1} \varepsilon$,
    $r \with r$,
    $c \with c$ 
    in the notation of \cref{ANN_for_product1}) 
    ensures that 
    there exists $\Phi \in \ANNsm$ such that for all $x \in \R$ it holds that 
    \begin{equation}
        \begin{split}
            & \Ra(\Phi) \in C(\R,\R), 
            \qquad 
            \paramANN(\Phi) \le
            2^{r+2} 4^{\frac{2r}{q-2}} c 
            \bpr{(2^{q-1}+1)^{-1} \varepsilon}^{-\frac{rq}{q-2}} \\
            & \text{and} \qquad 
            \abs{(\Ra(\Phi))(x) - x^2} \le 
            (2^{q-1} + 1)^{-1} \varepsilon \max\{1,\abs{x}^q\}
        \end{split}
    \end{equation}
    \cfload. 
    Therefore, \cref{ANN_for_product2} (applied with 
    $\Phi \with \Phi$, $a\with a$, $q\with q$, 
    $\varepsilon \with \varepsilon$,
    $r \with \tfrac{rq}{q-2}$,
    $c \with 2^{r+2} 4^{2r/(q-2)} c$   
    in the notation of \cref{ANN_for_product2})
    demonstrates 
    that there exists $\Gamma \in \ANNsm$ such that for all $x,y \in \R$ it holds that 
    \begin{equation}
        \begin{split}
            &\Ra(\Gamma) \in C(\R^2,\R), 
            \qquad
            \paramANN(\Gamma) \le
            9 \cdot 2^{r+2} 4^{\tfrac{2r}{q-2}} c (2^{q-1}+1)^{\tfrac{rq}{q-2}} \varepsilon^{-\tfrac{rq}{q-2}}, \\
            & \text{and} \qquad 
            \abs{(\Ra(\Gamma))(x,y) - xy } \le 
            \varepsilon \max\{1,\abs{x}^q, \abs{y}^q\}
            \dott
        \end{split}
    \end{equation}
    \argument{the fact that $2^{q-1}+1\le 2^q$}{
    that 
    \begin{equation}
    \begin{split}
        9 \cdot 2^{r+2} 4^{\tfrac{2r}{q-2}} c (2^{q-1}+1)^{\tfrac{rq}{q-2}} 
        & = 
        36 \cdot 2^{r+\tfrac{4r}{q-2}} c (2^{q-1}+1)^{\tfrac{rq}{q-2}} 
        \\
        & \le
        36 \cdot 2^{r+\tfrac{4r}{q-2}} c 2^{\tfrac{rq^2}{q-2}}
        = 36 \cdot 2^{\tfrac{(q^2+q+2)r}{q-2}} c 
        \dott 
    \end{split}
    \end{equation}
    }
\end{aproof}

\cfclear
\begin{athm}{corollary}{ANN_for_product3}
    Let $\varepsilon \in (0,1]$, $q\in(2,\infty)$, $\leaky \in \R\backslash\{-1,1\}$ and let $a \colon \R\to\R$ satisfy for all $x \in \R$ that $a(x)=\max\{x,\leaky x\}$\cfload.
    Then there exists $\Gamma \in \ANNsm$ such that for all $x,y \in \R$ it holds that 
    \begin{equation}
        \begin{split}
            &\Ra(\Gamma) \in C(\R^2,\R), \qquad
            \paramANN(\Gamma)\le
            864 \cdot 2^{\tfrac{q^3+3q^2-2q}{(q-2)(q-1)}}
            \varepsilon^{-\tfrac{q^2}{(q-2)(q-1)}}, \\
            & \text{and} \qquad 
            \abs{(\Ra(\Gamma))(x,y)-xy} \le \varepsilon \max\{1,\abs{x}^q,\abs{y}^q\}
        \end{split}
    \end{equation}
    \cfout.
\end{athm}

\cfclear
\begin{aproof}
    Let $\delta = 2^{-1} 4^{-2/(q-2)} (2^{q-1} + 1)^{-q/(q-2)} \varepsilon^{q/(q-2)}$ 
    and let $f\colon \R \to \R$ satisfy for all $x \in [0,1]$, $y \in \R\backslash[0,1]$ that $f(x)=x^2$ and $f(y)=\max\{y,0\}$. 
    \Nobs that for all $x,y\in \R$ it holds that 
    $\abs{f(x)-f(y)}\le 2 \abs{x-y}$. 
    Therefore, \cite[Corollary~4.13]{ackermann2023deep} (applied with $\varepsilon \with \delta$, 
    $L \with 2$, 
    $q \with q$, $\alpha \with \leaky$, $f \with f$, $a\with a$ 
    in the notation of \cite[Corollary~4.13]{ackermann2023deep})
    establishes 
    that 
    there exists $\G \in \ANNsm$ such that for all $x \in \R$ it holds that 
    \begin{equation}\llabel{eq:1032}
        \begin{split}
            &\Ra(\G) \in C(\R,\R), \qquad
            \paramANN(\G) \le 24 \cdot 4^{\frac{q}{q-1}} \cdot \delta^{-\frac{q}{q-1}}, \\
            &\text{and} \qquad 
            \abs{(\Ra(\G))(x) - f(x)} \le 
            \delta \max\{1,\abs{x}^q\}
            \dott
        \end{split}
    \end{equation}
    \cfload\dott 
    \argument{\lref{eq:1032}; \cref{ANN_for_product12} (applied with 
    $f\with f$, $\G \with \G$, $a\with a$, $q\with q$, 
    $\varepsilon \with \varepsilon$,
    $r \with \tfrac{q}{q-1}$,
    $c \with 24\cdot 4^{q/(q-1)}$ 
    in the notation of \cref{ANN_for_product12})}{
    that there exists $\Gamma \in \ANNsm$ such that for all $x,y \in \R$ it holds that 
    \begin{equation}
        \begin{split}
            &\Ra(\Gamma) \in C(\R^2,\R), \qquad 
            \paramANN(\Gamma) \le
            36 \cdot 2^{\tfrac{(q^2+q+2)q}{(q-2)(q-1)}} \cdot 
            24 \cdot 4^{\frac{q}{q-1}} 
            \varepsilon^{-\tfrac{q^2}{(q-2)(q-1)}}, \\
            & \text{and} \qquad 
            \abs{(\Ra(\Gamma))(x,y) - xy } \le 
            \varepsilon \max\{1,\abs{x}^q, \abs{y}^q\}
            \dott
        \end{split}
    \end{equation}
    }
    \argument{the fact that $\tfrac{(q^2+q+2)q}{(q-2)(q-1)} + \frac{2q}{q-1}=\tfrac{(q^2+3q-2)q}{(q-2)(q-1)}$}{
    that 
    \begin{equation}
    \begin{split}
        36 \cdot 2^{\tfrac{(q^2+q+2)q}{(q-2)(q-1)}} \cdot 
        24 \cdot 4^{\frac{q}{q-1}} 
        & 
        = 864 \cdot 2^{\tfrac{(q^2+3q-2)q}{(q-2)(q-1)}}
        \dott 
    \end{split}
    \end{equation}
    }
\end{aproof}

\cfclear
\begin{athm}{corollary}{ANN_for_product4}
    Let $\varepsilon \in (0,1]$, $q\in(2,\infty)$ 
    and 
    let $a \colon \R\to\R$ satisfy for all $x \in \R$ that $a(x)=\ln(1+\exp(x))$\cfload.
    Then there exists $\Gamma \in \ANNsm$ such that for all $x,y \in \R$ it holds that 
    \begin{equation}
        \begin{split}
            &\Ra(\Gamma) \in C(\R^2,\R), \qquad 
            \paramANN(\Gamma)\le
            1728 \cdot 2^{\tfrac{q^3+3q^2-2q}{(q-2)(q-1)}} 
            \varepsilon^{-\tfrac{q^2}{(q-2)(q-1)}}, \\
            &\text{and} \qquad 
            \abs{(\Ra(\Gamma))(x,y)-xy} \le \varepsilon \max\{1,\abs{x}^q,\abs{y}^q\}
        \end{split}
    \end{equation}
    \cfout.
\end{athm}

\cfclear
\begin{aproof}
    Let $\delta = 2^{-1} 4^{-2/(q-2)} (2^{q-1} + 1)^{-q/(q-2)} \varepsilon^{q/(q-2)}$ 
    and let $f\colon \R \to \R$ satisfy for all $x \in [0,1]$, $y \in \R\backslash[0,1]$ that $f(x)=x^2$ and $f(y)=\max\{y,0\}$. 
    \Nobs that for all $x,y\in \R$ it holds that 
    $\abs{f(x)-f(y)}\le 2 \abs{x-y}$.  
    Hence, \cite[Corollary~4.14]{ackermann2023deep} (applied with $\varepsilon \with \delta/2$, 
    $L \with 2$, 
    $q \with q$, $f \with f$, $a\with a$ 
    in the notation of \cite[Corollary~4.14]{ackermann2023deep})
    demonstrates 
    that 
    there exists $\G \in \ANNsm$ such that for all $x \in \R$ it holds that 
    \begin{equation}\llabel{eq:1035}
        \begin{split}
            &\Ra(\G) \in C(\R,\R), \qquad
            \paramANN(\G) \le 12 \cdot 4^{\frac{q}{q-1}} \cdot 2^{\frac{q}{q-1}} \delta^{-\frac{q}{q-1}}, \\
            &\text{and} \qquad
            \abs{(\Ra(\G))(x) - f(x)} \le 
            \delta \max\{1,\abs{x}^q\}
            \dott
        \end{split}
    \end{equation}
    \cfload\dott
    \argument{\lref{eq:1035}; \cref{ANN_for_product12} (applied with 
    $f\with f$, $\G \with \G$, $a\with a$, $q\with q$, 
    $\varepsilon \with \varepsilon$,
    $r \with \tfrac{q}{q-1}$,
    $c \with 12 \cdot 4^{q/(q-1)} \cdot 2^{q/(q-1)}$ 
    in the notation of \cref{ANN_for_product12})}{
    that there exists $\Gamma \in \ANNsm$ such that for all $x,y \in \R$ it holds that 
    \begin{equation}
        \begin{split}
            &\Ra(\Gamma) \in C(\R^2,\R), \qquad
            \paramANN(\Gamma) \le
            36 \cdot 2^{\tfrac{(q^2+q+2)q}{(q-2)(q-1)}} \cdot 
            12 \cdot 4^{\frac{q}{q-1}} \cdot 2^{\frac{q}{q-1}} 
            \varepsilon^{-\tfrac{q^2}{(q-2)(q-1)}} , \\
            &\text{and} \qquad 
            \abs{(\Ra(\Gamma))(x,y) - xy } \le 
            \varepsilon \max\{1,\abs{x}^q, \abs{y}^q\}
            \dott
        \end{split}
    \end{equation}
    }
    \argument{the fact that $\tfrac{q}{q-1}\le 2$}{
    that 
    \begin{equation}
    \begin{split}
        36 \cdot 2^{\tfrac{(q^2+q+2)q}{(q-2)(q-1)}} \cdot 
        12 \cdot 4^{\frac{q}{q-1}} \cdot 2^{\frac{q}{q-1}} 
        & 
        = 432 \cdot 2^{\tfrac{(q^2+3q-2)q}{(q-2)(q-1)}}
        \cdot 2^{\frac{q}{q-1}} 
        \le 1728 \cdot 2^{\tfrac{(q^2+3q-2)q}{(q-2)(q-1)}} 
        \dott 
    \end{split}
    \end{equation}
    } 
\end{aproof}

\subsection{ANN approximations for interpolation functions}
\label{subsec:ANN_approx_lin_interpol}

\cfclear
\begin{athm}{proposition}{ANN_for_MLP1}
    Let 
    $K, d, \fd\in\N$, 
    $\varepsilon \in (0,1]$, 
    $q \in (2,\infty)$, 
    $a,\allowbreak h_{0},\allowbreak h_{1},\allowbreak \dots,\allowbreak h_{K} \in C(\R,\R)$, 
    $f_0,\allowbreak f_1,\allowbreak \dots,\allowbreak f_K \in C(\R^d,\R)$, 
    $\ANNhatfct_{0},\allowbreak \ANNhatfct_{1},\allowbreak \dots,\allowbreak \ANNhatfct_{K},\allowbreak \F_0,\allowbreak \F_1,\allowbreak \dots,\allowbreak \F_K \in \ANNsm$ 
    satisfy for all $k\in\{0,1,\dots,K\}$, $t\in\R$ that
    $\Ra(\F_k)= f_k$, $\Ra(\ANNhatfct_{k}) \in C(\R,\R)$, and 
    \begin{equation}
        \abs{h_k(t)-(\Ra(\ANNhatfct_{k}))(t)} 
        \le \varepsilon \max\cu{1,\abs{t}^q},
    \end{equation}
    let $\fJ, \Gamma \in \ANNsm$ satisfy 
    for all $v,w \in \R$ that 
    $\dims(\fJ)=(1,\fd,1)$, $\functionANN{a}(\fJ)=\id_{\R}$,
    $\Ra(\Gamma) \in C(\R^2,\R)$, 
    and 
    \begin{equation}\label{eq:ANN_for_MLP1_product}
        \abs{vw-(\Ra(\Gamma))(v,w)} 
        \le \varepsilon \max\cu{1,\abs{v}^q,\abs{w}^q}, 
    \end{equation}
    and 
    let $\Phi \in \ANNsm$ 
    be given by 
    \begin{equation}
        \Phi
        = \BSum{k=0}{\fJ}{K} \bpr{ \compANN{\Gamma}{\parallelization_{2,\fJ}\pr{\ANNhatfct_{k},\F_k}} }
    \end{equation}
    \cfload.
    Then 
    \begin{enumerate}[label=(\roman *)]
    \item
    \label{ANN_for_MLP1_item0}
    it holds that $\Ra(\Phi) \in C(\R^{d+1},\R)$, 
    \item
    \label{ANN_for_MLP1_item1}
    it holds for all $t\in\R$, $x \in \R^d$ that 
    \begin{equation}
        \bpr{ \functionANN{a}(\Phi) }(t,x) = \sum_{k=0}^K \bpr{\functionANN{a}(\Gamma)}\bpr{ \bpr{\Ra(\ANNhatfct_{k})}(t), f_k(x)},
    \end{equation}
    \item
    \label{ANN_for_MLP1_item1b}
    it holds for all $t\in\R$, $x \in \R^d$ that 
    \begin{equation}
        \bbbabs{\bpr{ \functionANN{a}(\Phi) }(t,x) - \sum_{k=0}^K h_k(t) f_k(x) }
        \le 
        2\varepsilon \sum_{k=0}^{K} 
        \bpr{1+\abs{f_k(x)}^q } 
        \bpr{ \max\cu{1,\abs{t}^q } + \abs{h_k(t)} }^q
        ,
    \end{equation}
    \item
    \label{ANN_for_MLP1_item2}
    it holds that 
    \begin{equation}
    \lengthANN(\Phi) 
    \le 2 \lengthANN(\Gamma) 
    \bbbbr{\max_{k\in\{0,1,\dots,K\}} \lengthANN(\ANNhatfct_{k}) }
    \bbbbr{\max_{k\in\{0,1,\dots,K\}} \lengthANN(\F_k) } , 
    \end{equation}
    \item
    \label{ANN_for_MLP1_item3}
    it holds that
    \begin{equation}
    \normmm{ \dims(\Phi) } \le 
    2\fd
    (K+1)\, 
    \normmm{\dims(\Gamma)}\, \bbbbr{ \max_{k \in \{0,1,\dots,K\}}\normmm{ \dims(\ANNhatfct_{k}) } } \bbbbr{ \max_{k \in \{0,1,\dots,K\}}\normmm{ \dims(\F_k) } },
    \end{equation}
    and 
    \item
    \label{ANN_for_MLP1_item4}
    it holds that 
    \begin{equation}
        \begin{split}
            \paramANN(\Phi)
            & \le 16 \fd^2 (K+1)^2 
            \bbr{\paramANN(\Gamma)}^3
            \bbbbr{ \max_{k \in \{0,1,\dots,K\}} \paramANN(\H_{k}) }^3 \\
            & \quad \cdot
            \bbbbr{\max_{k\in\{0,1,\dots,K\}} \lengthANN(\F_k) }
            \bbbbr{ \max_{k \in \{0,1,\dots,K\}}\normmm{ \dims(\F_k) } }^2 
        \end{split}
    \end{equation}
    \end{enumerate}
    \cfout.
\end{athm}

\cfclear
\begin{aproof}
    \Nobs that \eg 
    item~(i) in Corollary~2.23 in~\cite{GrohsHornungJentzen2019}, 
    item~(vi) in Proposition~2.6 in~\cite{GrohsHornungJentzen2019}, 
    and \cref{diff_add_lemma_item4} in \cref{diff_add_lemma}
    establish \cref{ANN_for_MLP1_item0}.
    \startnewargseq
    \argument{\cref{diff_add_lemma_item5} in \cref{diff_add_lemma}}{
    that for all $t \in \R$, $x \in \R^d$ it holds that 
    \begin{equation}\llabel{eq:1208a}
        \bpr{ \functionANN{a}(\Phi) }(t,x) = \sum_{k=0}^K \bpr{\functionANN{a}\bpr{\compANN{\Gamma}{\parallelization_{2,\fJ}\pr{\ANNhatfct_{k},\F_k}}}}(t,x)
        \dott
    \end{equation}
    }
    \argument{\eg item~(vi) in Proposition~2.6 in~\cite{GrohsHornungJentzen2019}; item~(ii) in Corollary~2.23 in~\cite{GrohsHornungJentzen2019};the fact that for all $k\in\{0,1,\dots,K\}$ it holds that $\Ra(\F_k)=f_k$}{ 
    that for all $t \in \R$, $x \in \R^d$, $k\in\{0,1,\dots,K\}$ 
    it holds that 
    \begin{equation}\llabel{eq:1208b}
    \begin{split}
        \bpr{\functionANN{a}\bpr{\compANN{\Gamma}{\parallelization_{2,\fJ}\pr{\ANNhatfct_{k},\F_k}}}}(t,x)
        & = \bpr{\functionANN{a}(\Gamma)}\bpr{\functionANN{a}\bpr{\parallelization_{2,\fJ}\pr{\ANNhatfct_{k},\F_k}}(t,x)} \\ 
        & = \bpr{\functionANN{a}(\Gamma)} \bpr{ \bpr{\functionANN{a}(\ANNhatfct_{k})}(t), f_k(x) }
        \dott 
    \end{split}
    \end{equation}
    } 
    \argument{\lref{eq:1208b}; \lref{eq:1208a}}{
    \cref{ANN_for_MLP1_item1}\dott{}
    }
    \startnewargseq
    \argument{\cref{ANN_for_MLP1_item1}; the triangle inequality}{
    that for all $t\in\R$, $x\in\R^d$ it holds that 
    \begin{equation}\llabel{eq:1931a}
        \begin{split}
            & \bbbabs{\bpr{ \functionANN{a}(\Phi) }(t,x) - \sum_{k=0}^K h_k(t) f_k(x) } 
            \le 
            \sum_{k=0}^K \babs{ \bpr{\functionANN{a}(\Gamma)}\bpr{ \bpr{\Ra(\ANNhatfct_{k})}(t), f_k(x)}
            - h_k(t) f_k(x) } 
            \dott 
        \end{split}
    \end{equation}
    }
    \argument{the triangle inequality; the fact that for all $v,w \in \R$ it holds that $\abs{vw-(\Ra(\Gamma))(v,w)} 
    \le \varepsilon \max\cu{1,\abs{v}^q,\abs{w}^q}$; the fact that for all $t\in\R$, $k\in\{0,1,\dots,K\}$ it holds that $\abs{h_k(t)-(\Ra(\ANNhatfct_{k}))(t)} \le \varepsilon \max\cu{1,\abs{t}^q}$}{
    that for all $t\in\R$, $x\in\R^d$, $k\in\{0,1,\dots,K\}$ it holds that 
    \begin{equation}\llabel{eq:1931b}
        \begin{split}
            & \babs{ \bpr{\functionANN{a}(\Gamma)}\bpr{ \bpr{\Ra(\ANNhatfct_{k})}(t), f_k(x)}
            - h_k(t) f_k(x) 
            } \\
            & \le 
            \babs{ \bpr{\functionANN{a}(\Gamma)}\bpr{ \bpr{\Ra(\ANNhatfct_{k})}(t), f_k(x)}
            - \bpr{\Ra(\ANNhatfct_{k})}(t) f_k(x) 
            } 
            + \babs{ \bpr{\Ra(\ANNhatfct_{k})}(t)
            - h_k(t)} \abs{ f_k(x) } \\
            & \le 
            \varepsilon \max\bcu{1,\babs{\bpr{\Ra(\ANNhatfct_{k})}(t)}^q,\abs{f_k(x)}^q}
            + \varepsilon \max\cu{1,\abs{t}^q}
             \abs{ f_k(x) } \\
            & \le 
            \varepsilon \max\bcu{1,\babs{\bpr{\Ra(\ANNhatfct_{k})}(t)}^q}
            \bpr{ 1+\abs{ f_k(x) }^q} 
            + \varepsilon \max\cu{1,\abs{t}^q}
             \bpr{ 1+\abs{ f_k(x) }^q}
             \dott 
        \end{split}
    \end{equation}
    }
    \argument{the triangle inequality; the fact that for all $t\in\R$, $k\in\{0,1,\dots,K\}$ it holds that $\abs{h_k(t)-(\Ra(\ANNhatfct_{k}))(t)} \le \varepsilon \max\cu{1,\abs{t}^q}$}{
    that for all $t \in \R$, $k\in\{0,1,\dots,K\}$ it holds that 
    \begin{equation}\llabel{eq:1931c}
        \begin{split}
            \babs{\bpr{\Ra(\ANNhatfct_{k})}(t)}
            & \le 
            \babs{\bpr{\Ra(\ANNhatfct_{k})}(t) - h_k(t)}
            + \abs{h_k(t)} 
            \le \max\cu{1,\abs{t}^q} + \abs{h_k(t)} 
            \dott 
        \end{split}
    \end{equation}
    }
    \argument{\lref{eq:1931b}; \lref{eq:1931c}}{
    that for all $t\in\R$, $x\in\R^d$, $k\in\{0,1,\dots,K\}$ it holds that 
    \begin{equation}\llabel{eq:1931d}
        \begin{split}
            & \babs{ \bpr{\functionANN{a}(\Gamma)}\bpr{ \bpr{\Ra(\ANNhatfct_{k})}(t), f_k(x)}
            - h_k(t) f_k(x) 
            }
            \le 
            2 \varepsilon 
            \bpr{ \max\cu{1,\abs{t}^q} + \abs{h_k(t)} }^q 
            \bpr{ 1+\abs{ f_k(x) }^q} 
            \dott 
        \end{split}
    \end{equation}
    }
    \argument{\lref{eq:1931d}; \lref{eq:1931a}}{
    \cref{ANN_for_MLP1_item1b}\dott{}
    }
    \startnewargseq
    \argument{\cref{diff_add_lemma_item1} in \cref{diff_add_lemma}; \eg item~(ii) in Proposition~2.6 in~\cite{GrohsHornungJentzen2019}}{
    that  
    \begin{equation}\llabel{eq:1300a}
        \lengthANN(\Phi) = \max_{k\in\{0,1,\dots,K\}} \lengthANN\bpr{\compANN{\Gamma}{\bpr{\parallelization_{2,\fJ}\pr{\ANNhatfct_k,\F_k}}}}
        = \max_{k\in\{0,1,\dots,K\}} 
        \bpr{ \lengthANN(\Gamma) + \lengthANN\bpr{\parallelization_{2,\fJ}\pr{\ANNhatfct_k,\F_k}} - 1 }
        \dott
    \end{equation}
    }
    \argument{
    \eg item~(ii) in Lemma~2.13 in~\cite{GrohsHornungJentzen2019} 
    }{
    that for all $k\in\{0,1,\dots,K\}$ it holds that 
    \begin{equation}\llabel{eq:1300b}
    \begin{split}
        \lengthANN\bpr{\parallelization_{2,\fJ}\pr{\ANNhatfct_k,\F_k}}
        & = \lengthANN\bpr{\parallelizationSpecial_{2}\bpr{ \longerANN{\max\cu{ \lengthANN(\ANNhatfct_k), \lengthANN(\F_k)},\fJ}(\ANNhatfct_k),\longerANN{\max\cu{ \lengthANN(\ANNhatfct_k), \lengthANN(\F_k)},\fJ}(\F_k) } } \\
        & = \max\bcu{ \lengthANN(\ANNhatfct_k), \lengthANN(\F_k) } \\
        & \le \bbbbr{\max_{j\in\{0,1,\dots,K\}} \lengthANN(\ANNhatfct_j) }
        \bbbbr{\max_{j\in\{0,1,\dots,K\}} \lengthANN(\F_j) }
        \dott
    \end{split}
    \end{equation}
    } 
    \argument{\lref{eq:1300b}; \lref{eq:1300a}}{
    \cref{ANN_for_MLP1_item2}. \, 
    }
    \startnewargseq
    \argument{\cref{diff_add_lemma_item3} in \cref{diff_add_lemma}}{
    that 
    \begin{equation}\llabel{eq:1356a}
        \normmm{\dims(\Phi)}
        \le (K+1) \max\bbcu{ \fd, \max_{k\in\{0,1,\dots,K\}} \normmm{\dims\bpr{\compANN{\Gamma}{\parallelization_{2,\fJ}\pr{\ANNhatfct_k,\F_k}}}} }
        \dott
    \end{equation}
    }
    \argument{\eg item~(i) in Proposition~2.6 in~\cite{GrohsHornungJentzen2019}}{
    that for all $k\in\{0,1,\dots,K\}$ it holds that 
    \begin{equation}\llabel{eq:1356b}
    \begin{split}
        \normmm{\dims\bpr{\compANN{\Gamma}{\parallelization_{2,\fJ}\pr{\ANNhatfct_k,\F_k}}}}
        & \le \max\bcu{ \normmm{\dims\bpr{\parallelization_{2,\fJ}\pr{\ANNhatfct_k,\F_k}}} , \normmm{\dims(\Gamma)} }
        \dott
    \end{split}
    \end{equation}
    }
    \argument{\eg item~(i) in Proposition~2.20 in~\cite{GrohsHornungJentzen2019}; 
    \eg \cite[Lemma~2.2.11]{jentzen2023mathematical}}{
    that for all $k\in\{0,1,\dots,K\}$ it holds that 
    \begin{equation}\llabel{eq:1356c}
        \begin{split}
            \normmm{\dims\bpr{\parallelization_{2,\fJ}\pr{\ANNhatfct_k,\F_k}}}
            & =
            \normmm{\dims\bpr{ \parallelizationSpecial_{2}\bpr{ \longerANN{\max\cu{ \lengthANN(\ANNhatfct_k), \lengthANN(\F_k)},\fJ}(\ANNhatfct_k),\longerANN{\max\cu{ \lengthANN(\ANNhatfct_k), \lengthANN(\F_k)},\fJ}(\F_k) } }} \\
            & \le \normmm{\dims\bpr{\longerANN{\max\cu{ \lengthANN(\ANNhatfct_k), \lengthANN(\F_k)},\fJ}(\ANNhatfct_k)}} + \normmm{\dims\bpr{\longerANN{\max\cu{ \lengthANN(\ANNhatfct_k), \lengthANN(\Xi_k)},\fJ}(\F_k)}}  \\
            & \le \max\bcu{ \fd, \normmm{\dims(\ANNhatfct_k)} }  + \max\bcu{ \fd, \normmm{\dims(\F_k)} } \\
            & \le \fd \normmm{\dims(\ANNhatfct_k)} + \fd \normmm{\dims(\F_k)} 
            \le 2 \fd \normmm{\dims(\ANNhatfct_k)} \normmm{\dims(\F_k)} 
            \dott
        \end{split}
    \end{equation}
    }
    \argument{\lref{eq:1356b};\lref{eq:1356c} 
    }{
    that for all $k\in\{0,1,\dots,K\}$ it holds that 
    \begin{equation}\llabel{eq:1716}
        \begin{split}
            \normmm{\dims\bpr{\compANN{\Gamma}{\parallelization_{2,\fJ}\pr{\ANNhatfct_k,\F_k}}}}
            & \le 
            2 \fd \normmm{\dims(\ANNhatfct_k)} \normmm{\dims(\F_k)} \normmm{ \dims(\Gamma) } 
            \dott
        \end{split}
    \end{equation}
    }
    \argument{\lref{eq:1716}; \lref{eq:1356a}}{
    \cref{ANN_for_MLP1_item3}\dott{}
    }
    \startnewargseq 
    \argument{\cref{eq:param_length_indim_outdim}; \cref{eq:dimlevel}; \cref{eq:dimsvector}}{
    that 
    \begin{equation}\llabel{eq:1624a}
    \begin{split}
    \paramANN\pr{\Phi}
    & = \sum_{m=1}^{\lengthANN(\Phi)} \dimANNlevel_{m}\pr{\Phi} \bpr{ \dimANNlevel_{m -1}\pr{\Phi} +1 } 
    \le \sum_{m=1}^{\lengthANN(\Phi)} \normmm{\dims\pr{\Phi}} \bpr{ \ \normmm{\dims\pr{\Phi}} +  \normmm{\dims\pr{\Phi}} } 
    = 2 \lengthANN\pr{\Phi} \normmm{\dims\pr{\Phi}}^2 
    \dott
    \end{split}
    \end{equation}
    }
    \argument{\cref{ANN_for_MLP1_item2}; \eg \cite[Lemma 2.4]{ackermann2023deep}}{
    that 
    \begin{equation}\llabel{eq:1624b}
    \lengthANN(\Phi) 
    \le 2 \paramANN(\Gamma) 
    \bbbbr{\max_{k\in\{0,1,\dots,K\}} \paramANN(\ANNhatfct_k) }
    \bbbbr{\max_{k\in\{0,1,\dots,K\}} \lengthANN(\F_k) } 
    \dott
    \end{equation}
    }
    \argument{\cref{ANN_for_MLP1_item3}; \eg \cite[Lemma 2.4]{ackermann2023deep}}{
    that 
    \begin{equation}\llabel{eq:1624c}
    \normmm{ \dims(\Phi) } \le 
    2\fd
    (K+1)\, 
    \paramANN(\Gamma) \bbbbr{ \max_{k \in \{0,1,\dots,K\}}\paramANN(\ANNhatfct_k) } \bbbbr{ \max_{k \in \{0,1,\dots,K\}}\normmm{ \dims(\F_k) } }
    \dott
    \end{equation}
    }
    \argument{\lref{eq:1624c}; \lref{eq:1624a}; \lref{eq:1624b}}{
    \cref{ANN_for_MLP1_item4}\dott{}
    }
\end{aproof}

\subsection{ANN approximations for linear interpolations of MLP  approximations}
\label{subsec:ANN_approx_MLP}

\cfclear
\begin{athm}{corollary}{ANN_for_MLP2}
    Let 
    $K, d, \fd, M\in\N$, $T\in(0,\infty)$, 
    $q \in (2,\infty)$, 
    $\fx_{-1},\fx_0,\fx_1,\dots,\fx_K, \fx_{K+1} \in \R$ 
    satisfy 
    $\fx_{-1}<0=\fx_0 < \fx_1 < \ldots < \fx_K = T < \fx_{K+1}$, 
    let $a \in C(\R,\R)$,   
    $\fJ, \F, \G \in \ANNsm$ satisfy $\dims(\fJ)=(1,\fd,1)$, $\functionANN{a}(\fJ)=\id_{\R}$,
    $\functionANN{a}(\F) \in C(\R,\R)$,
    and $\functionANN{a}(\G) \in C(\R^d,\R)$, 
    for every $k \in \{0,1,\dots,K\}$, $\gamma \in (0,1]$ let $\ANNhatfct_{k,\gamma} \in \ANNsm$ 
    satisfy for all $t \in \R$ that 
    $\functionANN{a}(\ANNhatfct_{k,\gamma})\in C(\R,\R)$ and 
    \begin{equation}
        \babs{ \scrL_{\fx_{k-1},\fx_k,\fx_{k+1}}^{0,1,0}(t) - \bpr{\functionANN{a}(\ANNhatfct_{k,\gamma})}(t) }
        \le \gamma \max\cu{1,\abs{t}^q},
    \end{equation}
    for every $\gamma \in (0,1]$ let $\Gamma_\gamma \in\ANNsm$ 
    satisfy for all $v,w \in \R$ that 
    $\functionANN{a}(\Gamma_{\gamma})\in C(\R^2,\R)$ and 
    \begin{equation}\label{eq:ANN_for_MLP2_product}
        \babs{vw-\bpr{\functionANN{a}(\Gamma_{\gamma})}(v,w)} 
        \le \gamma \max\bcu{ 1, \abs{v}^q, \abs{w}^q }, 
    \end{equation}  
    let $\Theta = \bigcup_{n\in\N} \Z^n$, 
    for every 
    $\theta \in \Theta$
    let
    $\cU^\theta \colon [0,T]\to [0,T]$
    and
    $W^\theta\colon[0,T]\to \R^d$
    be functions, 
    let
    $\mlp_{n}^\theta \colon [0,T] \times \R^d\to \R$, $n \in \N_0$, $\theta \in \Theta$,
    satisfy for all 
    $\theta \in \Theta$, $n \in \N$, 
    $t\in[0,T]$, 
    $x\in\R^d$ 
    that $\mlp_0^\theta(t,x)=0$ and
    \begin{align}\label{def:mlp}
    & 
    \mlp_{n}^\theta(t,x) 
    = 
    \frac{1}{M^n} \bbbbbr{ \sum_{k=1}^{M^n} \pr[\big]{ \Ra(\G) } \lrSpace \pr[\big]{ x + W_{T-t}^{(\theta,0,-k)} } }
    \\
    & 
    + 
    \sum_{i=0}^{n-1} \frac{(T-t)}{M^{n-i}} \! \bbbbbr{ \sum_{k=1}^{M^{n-i}} \pr[\big]{ \pr[]{ \Ra(\F) \circ \mlp_{i}^{(\theta,i,k)} } - \1_{\N}(i) \pr[]{ \Ra(\F) \circ \mlp_{\max\{i-1,0\}}^{(\theta,-i,k)} } } \lrSpace \pr[\big]{ \cU_t^{(\theta,i,k)}, x + W_{\cU_t^{(\theta,i,k)}-t}^{(\theta,i,k)} } }
    \dc 
    \nonumber
    \end{align}
    let 
    $\U_{n,t}^\theta \in \cu{ \Phi \in \ANNsm \colon \functionANN{a}(\Phi) \in C(\R^d,\R) }$, $t\in[0,T]$, $n\in\N_0$, $\theta \in \Theta$,  
    satisfy 
    for all  
    $\theta \in \Theta$, 
    $n\in\N$, 
    $t\in[0,T]$ 
    that 
    $\U_{0,t}^\theta = ((0\ 0\ \dots\ 0),0) \in \R^{1\times d}\times \R^1$ 
    and 
    \begin{align}
    \U_{n,t}^\theta 
    & 
    = 
    \left[ \OSum{k=1}{M^n} \pr[\Big]{  \scalar{\tfrac{1}{M^n}}{\pr[\big]{ \compANN{ \G }{ \A_{\idMatrix_d,W_{T-t}^{(\theta,0,-k)}} } } } } \right]
    \nonumber
    \\
    & 
    \quad 
    \bSum_{\,\fJ} 
    \left[ \BSum{i=0}{\fJ}{n-1} \br[\Bigg]{ \scalar{ \pr[\Big]{ \tfrac{(T-t)}{M^{n-i}} } }{ \pr[\bigg]{ \BSum{k=1}{\fJ}{M^{n-i}} \pr[\Big]{ \compANN{ \pr[\big]{ \compANN{\F}{\U^{(\theta,i,k)}_{i,\mathcal{U}_t^{(\theta,i,k)}} } } }{ \A_{\idMatrix_d, W_{\mathcal{U}_t^{(\theta,i,k)}-t}^{(\theta,i,k)}} } } } } } \right]
    \\
    & 
    \quad 
    \bSum_{\,\fJ} 
    \left[ \BSum{i=0}{\fJ}{n-1} \br[\Bigg]{ \scalar{ \pr[\Big]{ \tfrac{(t-T) \, \1_{\N}(i)}{M^{n-i}} } }{ \pr[\bigg]{ \BSum{k=1}{\fJ}{M^{n-i}} \pr[\Big]{ \compANN{ \pr[\big]{ \compANN{\F}{\U^{(\theta,-i,k)}_{\max\{i-1,0\},\mathcal{U}_t^{(\theta,i,k)}} } } }{ \A_{\idMatrix_d, W_{\mathcal{U}_t^{(\theta,i,k)}-t}^{(\theta,i,k)}} } } } } } \right] ,
    \nonumber
    \end{align}
    and 
    for every $\theta \in \Theta$, $\gamma \in (0,1]$, $n \in \N$ 
    let $\Phi_{\gamma,n}^{\theta} \in \ANNsm$
    satisfy 
    \begin{equation}
        \Phi_{\gamma,n}^{\theta} = \BSum{k=0}{\fJ}{K} \bpr{ \compANN{\Gamma_{\gamma}}{\parallelization_{2,\fJ}\bpr{\ANNhatfct_{k,\gamma},\U_{n,\fx_k}^\theta }} }
    \end{equation}
    \cfload. 
    Then 
    \begin{enumerate}[label=(\roman *)] 
    \item
    \label{ANN_for_MLP2_item1}
    it holds for all $\theta\in\Theta$, $\gamma \in (0,1]$, $n\in\N$,  $t\in \R$, $x \in \R^d$ that 
    $\Ra(\Phi_{\gamma,n}^{\theta}) \in C(\R^{d+1},\R)$
    and 
    \begin{equation}
        \bpr{ \functionANN{a}(\Phi_{\gamma,n}^{\theta}) }(t,x) = \sum_{k=0}^K \bpr{\functionANN{a}(\Gamma_\gamma)}\bpr{ \bpr{\Ra(\ANNhatfct_{k,\gamma})}(t), \mlp_{n}^\theta(\fx_k,x) },
    \end{equation}
    \item
    \label{ANN_for_MLP2_item2}
    it holds for all  $\theta\in\Theta$, $\gamma \in (0,1]$, $n\in\N$,  $t\in[0,T]$, $x \in \R^d$ that
    \begin{equation}\label{eq:ANN_for_MLP2_approx}
    \begin{split}
    & \babs{\scrL_{\fx_0,\fx_1,\dots,\fx_K}^{\mlp_{n}^\theta(\fx_0,x),\mlp_{n}^\theta(\fx_1,x),\dots,\mlp_{n}^\theta(\fx_K,x)}(t) - \bpr{ \functionANN{a}(\Phi_{\gamma,n}^{\theta}) }(t,x) } \\
    & \le 2 \gamma \bpr{1+(T+1)^q}^q \bbbbpr{ \sum_{k=0}^K \bpr{1 + \abs{\mlp_{n}^\theta(\fx_k,x)}^{q}} }, 
    \end{split}
    \end{equation}
    and
    \item
    \label{ANN_for_MLP2_item3}
    it holds for all $\theta\in\Theta$, $\gamma \in (0,1]$, $n\in\N$ that 
    \begin{equation}
    \begin{split}
    \paramANN\bpr{\Phi_{\gamma,n}^{\theta}}
    & \le 16 \bpr{\max\{\fd,\lengthANN(\G)\} + \lengthANN(\F)} 
    \bbr{ \max\{\fd,\normmm{\dims(\F)},\normmm{\dims(\G)}\} }^2
    \bbr{n^{\frac12}(3M)^n}^2 \\
    & \quad \cdot \bbr{\paramANN(\Gamma_{\gamma})}^3 
    \bbbbr{\max_{k\in\{0,1,\dots,K\}} \paramANN(\ANNhatfct_{k,\gamma}) }^3
    (K+1)^2 \fd^2 
    \end{split}
    \end{equation}
    \end{enumerate}
    \cfout.
\end{athm}

\begin{aproof}
    \startnewargseq
    \argument{items~(ii) and~(v) in Proposition~3.9 in~\cite{ackermann2023deep}; 
    \cref{ANN_for_MLP1_item0,ANN_for_MLP1_item1} in \cref{ANN_for_MLP1} 
    (applied for every
    $\gamma \in (0,1]$, $n\in\N$, $\theta \in \Theta$ 
    with 
    $K\with K$, $d\with d$, $\fd \with \fd$, 
    $\varepsilon \with \gamma$, $q\with q$, $a\with a$, 
    $(h_k)_{k\in\{0,1,\dots,K\}} \with (\scrL_{\fx_{k-1},\fx_k,\fx_{k+1}}^{0,1,0})_{k\in\{0,1,\dots,K\}}$, 
    $(f_k)_{k\in\{0,1,\dots,K\}} \with (\R^d \ni x \mapsto \mlp_{n}^\theta(\fx_k,x) \in\R )_{k\in\{0,1,\dots,K\}}$, 
    $(\ANNhatfct_k)_{k\in\{0,1,\dots,K\}} \with (\ANNhatfct_{k,\gamma})_{k\in\{0,1,\dots,K\}}$, 
    $(\F_k)_{k\in\{0,1,\dots,K\}} \with (\U_{n,\fx_k}^\theta )_{k\in\{0,1,\dots,K\}}$, $\fJ \with \fJ$, $\Gamma \with \Gamma_{\gamma}$, $\Phi \with \Phi_{\gamma,n}^{\theta}$ 
    in the notation of \cref{ANN_for_MLP1})}{
    \cref{ANN_for_MLP2_item1}\dott{}
    }
    \startnewargseq 
    \argument{\cref{ANN_for_MLP1_item1b} in \cref{ANN_for_MLP1}; \cref{factorize_lin_interpolation}}{
    that for all $t\in[0,T]$, $x \in \R^d$, $\gamma \in (0,1]$, $n\in\N$, $\theta \in \Theta$ it holds that 
    \begin{equation}\llabel{eq:936}
        \begin{split}
            & \babs{\scrL_{\fx_0,\fx_1,\dots,\fx_K}^{\mlp_{n}^\theta(\fx_0,x),\mlp_{n}^\theta(\fx_1,x),\dots,\mlp_{n}^\theta(\fx_K,x)}(t) - \bpr{ \functionANN{a}(\Phi_{\gamma,n}^{\theta}) }(t,x) } \\
            & \le 
            2 \gamma \sum_{k=0}^K \bpr{1+ \abs{\mlp_{n}^\theta(\fx_k,x)}^q}
            \bpr{ \max\cu{1,\abs{t}^q} + \babs{\scrL_{\fx_{k-1},\fx_k,\fx_{k+1}}^{0,1,0}(t)} }^q
            \dott 
        \end{split}
    \end{equation}
    }
    \argument{\lref{eq:936};
    the fact that for all $t\in[0,T]$ it holds that $\max\cu{1,\abs{t}^q} \le (1+T)^q$; 
    the fact that for all $t\in \R$, $k\in\{0,1,\dots,K\}$ it holds that $\scrL_{\fx_{k-1},\fx_k,\fx_{k+1}}^{0,1,0}(t) \in [0,1]$}{
    \cref{ANN_for_MLP2_item2}\dott{}
    }
    \startnewargseq 
    \argument{item~(iii) in Proposition~3.9 in~\cite{ackermann2023deep}}{
    that for all $n\in\N$, $\theta\in\Theta$, $t\in[0,T]$ it holds that 
    \begin{equation}\llabel{eq:943}
        \begin{split}
            \lengthANN\pr{\U_{n,t}^\theta }
            & \le 
            \max\cu{ \fd , \lengthANN(\G) } + n \hiddenLength(\F) 
            \le 
            n \bpr{ \max\cu{ \fd , \lengthANN(\G) } + \lengthANN(\F) } 
            \dott
        \end{split}
    \end{equation}
    }
    \argument{item~(iv) in Proposition~3.9 in~\cite{ackermann2023deep}}{
    that for all $n\in\N$, $\theta\in\Theta$, $t\in[0,T]$ it holds that 
    \begin{equation}\llabel{eq:951}
        \begin{split}
            \normmm{\dims\pr{\U_{n,t}^\theta } }^2
            & \le 
            \bpr{ \max\cu{ \fd , \normmm{\dims(\F)}, \normmm{\dims(\G)} } }^2
            \bpr{(3M)^n}^2 
            \dott
        \end{split}
    \end{equation}
    }
    \argument{\lref{eq:951}; \lref{eq:943}; \cref{ANN_for_MLP1_item4} in \cref{ANN_for_MLP1}}{
    \cref{ANN_for_MLP2_item3}\dott{}
    }
\end{aproof}

\section{ANN approximations for solutions of semilinear heat PDEs}
\label{sec:ANN_approx_PDE}

In this section we establish the main \ANN\ approximation results of this work. 
In particular, in \cref{theorem:main} in \cref{subsec:ANN_approx_PDE_general_activation} we show that for every arbitrarily large absolute moment $\fq \in [2,\infty)$ and every arbitrarily large time horizon $T \in (0,\infty)$ we have that the solutions $u_d\colon [0,T] \times \R^d \to \R$, $d \in \N$, of the semilinear heat \PDEs\ in~\cref{eq:mainThm:heateq} below can be approximated on $[0,T] \times \R^d$, $d \in \N$, in the $L^{\fq}$-sense with respect to the measures $\nu_d \colon \cB(\R^{d+1}) \to [0,\infty)$, $d \in \N$, in~\cref{eq:mainthm_assump_measure} (see~\cref{final_item2_error}) 
without the \COD\ (see~\cref{final_item2_noCOD}) through realizations of \ANNs\ with a general activation function provided that there exist 
\begin{enumerate}[label=(\roman*)]
    \item 
    appropriate \ANN\ approximations $\Gamma_{\varepsilon} \in\ANNsm$, $\varepsilon \in (0,1]$, for the product function (see \cref{eq:assump_on_Gamma_params} and~\cref{eq:assump_on_Gamma_prodapprox} below), 
    \item 
    a shallow \ANN\ representation $\fJ \in \ANNsm$ for the one-dimensional identity function $\id_{\R} = ( \R \ni x \mapsto x \in \R)$, 
    \item
    appropriate \ANN\ approximations $\ANNhatfct_{K,k,\varepsilon} \in \ANNsm$, $\varepsilon \in (0,1]$, $k\in\{0,1,\dots,K\}$, $K\in\N$, for the hat functions (see~\cref{eq:assump_on_ANNforHatfct} and~\cref{eq:assump_on_ANNforHatfct_approx} below), 
    \item 
    appropriate \ANN\ approximations $\F_{0,\varepsilon} \in \ANNsm$, $\varepsilon \in (0,1]$, for the nonlinearity in the \PDE\ (see~\cref{eq:mainthm_assump_params,eq:mainthm_assump_pgrowth,eq:mainthm_assump_Lip} below), and 
    \item 
    appropriate \ANN\ approximations $\F_{d,\varepsilon}\in\ANNsm$, $\varepsilon \in (0,1]$, $d \in \N$, for the terminal value functions (see~\cref{eq:mainthm_assump_params} and~\cref{eq:mainthm_assump_pgrowth} below). 
\end{enumerate}
In our proof of \cref{theorem:main} we employ, among other things, 
\cref{ANN_for_MLP2} from \cref{sec:ANN_approx_lin_interpolation_MLP}, 
\cref{sol_stab2}, \cref{lem:BM_power_bound}, and \cref{temp_u_reg1} from \cref{sec:PropPDEs}, 
the strong $L^p$-error estimates for the employed \MLP\ approximations from Hutzenthaler et al.~\cite{PadgettJentzen2021}, 
the elementary complexity estimate in \cref{lemma_littleMM}, 
and the elementary and well-known measurability property in \cref{lin_interpolation_measurable}. 

In \cref{cor_of_mainthm1} and \cref{cor_of_mainthm2} in \cref{subsec:ANN_approx_PDE_specific_activation} we then specialize \cref{theorem:main} to the situation of \ANNs\ with the \ReLU, the leaky \ReLU, and the softplus activation function. 
Our proofs of \cref{cor_of_mainthm1} and \cref{cor_of_mainthm2}, respectively, employ 
the general \ANN\ approximation result for \PDEs\ in \cref{theorem:main}, 
the \ANN\ representation and approximation results for hat functions in \cref{subsec:hat_fct}, 
the \ANN\ approximation results for the product function from \cref{subsec:approx_product}, 
the \ANN\ representation results for the one-dimensional identity function in~\cite[Section~3.2]{ackermann2023deep}, 
and the \ANN\ approximation results for Lipschitz continuous nonlinearities in~\cite[Section~4.2]{ackermann2023deep}.

\subsection{ANN approximations for PDEs with general activation functions}
\label{subsec:ANN_approx_PDE_general_activation}

\cfclear
\begin{athm}{lemma}{lemma_littleMM}
    Let $\LipConstF, T \in(0,\infty)$, $p\in[2,\infty)$, 
    $\fB \in[1,\infty)$, 
    $(\littleMM_k)_{k\in\N} \subseteq \N$ 
    satisfy for all 
    $k \in \N$ 
    that 
    $\liminf_{j\to\infty} \littleMM_j = \infty$, 
    $\limsup_{j\to\infty} \nicefrac{(\littleMM_j)^{p/2}}{j} < \infty$,
    and
    $\littleMM_{k+1} \le \fB \littleMM_k$,
    and let $(N_{\varepsilon})_{\varepsilon \in (0,1]}$ satisfy for all $\varepsilon \in (0,1]$
    that 
    \begin{equation}\label{eq:lemma_littleM_def_N_eps}
        N_{\varepsilon}
        = \inf\bbbbpr{\bbbcu{  
        n\in\N \colon
        \bbbbr{ \pr{1+\LipConstF T} \pr{\littleMM_n}^{-\frac12} \exp\bbbpr{\frac{\pr{\littleMM_n}^{\frac{p}{2}}}{n}} }^n
        \le \varepsilon
        }
        \cup \{\infty\}
        }
        \dott
    \end{equation}
    Then
    \begin{enumerate}[label=(\roman *)]
    \item
    \label{lemma_littleMM_item1}
        it holds for all $\varepsilon \in (0,1]$ that $N_{\varepsilon}< \infty$
        and 
    \item
    \label{lemma_littleMM_item2a}
        it holds for all $\delta \in (0,\infty)$ that
        \begin{equation}
        \sup_{ \varepsilon \in (0,1] }
        \bpr{
          \varepsilon^{2+\delta}
          \pr{ 
            N_{ \varepsilon } 
          }^{ \nicefrac{ 1 }{ 2 } } 
          \pr{ 
            3  \littleMM_{ N_{ \varepsilon } }
          }^{ N_{ \varepsilon } } 
        }
        < \infty
        \dott 
        \end{equation}  
    \end{enumerate}
\end{athm}

\cfclear
\begin{aproof}
    Throughout this proof 
    let $(\fm_n)_{n \in \N}$ satisfy for all 
    $n \in \N$ that 
    \begin{equation}\label{eq:lemma_littleM_constant_m}
        \fm_n = 
        \bbbbr{ (1+2\LipConstF T)  \pr{\littleMM_n}^{-\frac12} \exp\bbbpr{\frac{\pr{\littleMM_n}^{\frac{p}{2}}}{n}}  }^n
        \dott 
    \end{equation}
    \Nobs that the fact that $\limsup_{j\to\infty} \nicefrac{(\littleMM_j)^{p/2}}{j} < \infty$ and the fact that 
    ${\liminf_{j\to\infty} \littleMM_j = \infty}$
    show that 
    $\limsup_{n\to\infty} \fm_n=0$. 
    Combining this and \cref{eq:lemma_littleM_def_N_eps} proves  
    \cref{lemma_littleMM_item1}.
    \startnewargseq 
    \argument{\cref{eq:lemma_littleM_def_N_eps}; \cref{eq:lemma_littleM_constant_m}}{
    that for all $\varepsilon \in (0,1]$ with $N_{\varepsilon} \in \N \cap [2,\infty)$ it holds that 
    $\fm_{N_\varepsilon-1} >\varepsilon$\dott\,
    }
    Hence, \cref{eq:lemma_littleM_constant_m}
    shows 
    that for all $\delta \in (0,\infty)$, $\varepsilon \in (0,1]$ with $N_{\varepsilon} \in \N \cap [2,\infty)$ it holds that
    \begin{equation}\label{eq:2235}
        \begin{split}
            & \pr{ N_{ \varepsilon } }^{ \nicefrac{ 1 }{ 2 } } 
            (3\littleMM_{N_\varepsilon})^{N_{\varepsilon}} \\
            & \le 
            \pr{ N_{ \varepsilon } }^{ \nicefrac{ 1 }{ 2 } }
            (3\littleMM_{N_\varepsilon})^{N_{\varepsilon}} 
            \bpr{\varepsilon^{-1} \fm_{N_{\varepsilon}-1} }^{2+\delta} \\
            & \le 
            \varepsilon^{-(2+\delta)} 
            \sup_{n\in\N}
            \bbpr{ (n+1)^{\frac12} (3\littleMM_{n+1})^{n+1} \pr{\fm_n}^{2+\delta}  }
            \\
            & \le   
            \varepsilon^{-(2+\delta)}
            \sup_{n\in\N}
            \bbbbpr{ (n+1)^{\frac12}  \pr{\littleMM_{n+1}}^{n+1}
            \bbbbr{ 3 (1+2\LipConstF T) \exp\bbbpr{\frac{\pr{\littleMM_{n}}^{\frac{p}{2}}}{n}} \pr{\littleMM_{n}}^{-\frac12} }^{n(2+\delta)}
            }
            \dott
        \end{split}
    \end{equation}
    \argument{the fact that for all $n\in\N$ it holds that $\littleMM_{n+1}\le \fB \littleMM_n$}{
    that for all $n\in\N$ it holds that $(\littleMM_{n+1})^{n+1}\le \littleMM_{n+1} \fB^n (\littleMM_n)^n \le \littleMM_1 \fB^{2n} (\littleMM_n)^n$\dott\,
    }
    Combining this and \cref{eq:2235} establishes 
    that for all $\delta \in (0,\infty)$, $\varepsilon \in (0,1]$ with $N_{\varepsilon} \in \N \cap [2,\infty)$ it holds that
    \begin{equation}\llabel{eq:2334a}
        \begin{split}
            \pr{ N_{ \varepsilon } }^{ \nicefrac{ 1 }{ 2 } } 
            (3\littleMM_{N_\varepsilon})^{N_{\varepsilon}}
            & \le   
            \varepsilon^{-(2+\delta)}
            \sup_{n\in\N}
            \bbbbpr{ (n+1)^{\frac12}  \littleMM_{1} \pr{\littleMM_n}^{-\frac{n\delta}{2}} 
            \bbbbr{ 3 (1+2\LipConstF T) \fB \exp\bbbpr{\frac{\pr{\littleMM_{n}}^{\frac{p}{2}}}{n}}  }^{n(2+\delta)}
            } 
            \dott
        \end{split}
    \end{equation}
    \argument{\lref{eq:2334a}; the fact that $\fB\in[1,\infty)$; the fact that for all $n\in\N$, $\delta \in (0,\infty)$ it holds that 
    \begin{equation}
        \begin{split}
            \pr{\littleMM_1}^{-\frac{\delta}{2}} \bpr{\exp\bpr{\pr{\littleMM_1}^{\frac{p}{2}}}}^{2+\delta} & \ge 
            \pr{\littleMM_1}^{-\frac{\delta}{2}} \bpr{1+\pr{\littleMM_1}^{\frac{p}{2}}}^{2+\delta}
            \ge 
            \pr{\littleMM_1}^{ \frac{(p -1 )\delta}{2} + p }
            \ge 1
        \end{split}
    \end{equation}
    }{
    that for all $\delta \in (0,\infty)$, $\varepsilon \in (0,1]$ it holds that
    \begin{equation}\llabel{eq:2334b}
        \begin{split}
            \pr{ N_{ \varepsilon } }^{ \nicefrac{ 1 }{ 2 } } 
            (3\littleMM_{N_\varepsilon})^{N_{\varepsilon}}
            & \le   
            \varepsilon^{-(2+\delta)}
            \littleMM_1 
            \sup_{n\in\N}
            \bbbbbr{ (n+1)^{\frac{1}{2n}} \pr{\littleMM_n}^{-\frac{\delta}{2}}
            \bbbbr{ 3(1+2\LipConstF T) \fB \exp\bbbpr{\frac{(\littleMM_n)^{\frac{p}{2}}}{n}} }^{2+\delta}
            }^n 
            \dott
        \end{split}
    \end{equation}
    }
    \argument{the fact that
    $\limsup_{n\to\infty} \nicefrac{(\littleMM_n)^{p/2}}{n} < \infty$; the fact that $\liminf_{n\to\infty} \littleMM_n = \infty$; the fact that $\lim_{n\to\infty} (n+1)^{1/(2n)} = 1$}{
    that for all $\delta\in(0,\infty)$ it holds that 
    \begin{equation}\llabel{eq:2352}
        \begin{split}
            & \sup_{n\in\N}
            \bbbbbr{ (n+1)^{\frac{1}{2n}} \pr{\littleMM_n}^{-\frac{\delta}{2}}
            \bbbbr{ 3(1+2\LipConstF T) \fB \exp\bbbpr{\frac{(\littleMM_n)^{\frac{p}{2}}}{n}} }^{2+\delta}
            }^n
            < \infty \dott
        \end{split}
    \end{equation}
    }
    \argument{\lref{eq:2352}; \lref{eq:2334b}}{
    \cref{lemma_littleMM_item2a}\dott{}
    } 
\end{aproof}


\cfclear
\begin{athm}{lemma}{lin_interpolation_measurable}
    Let $(\Omega, \cF)$ be a measurable space, let $T\in(0,\infty)$, $d, K \in\N$,  
    for every $k\in\{0,1,\dots,K\}$ let $h_k\colon [0,T] \to \R$ and 
    $f_k \colon \R^d \times \Omega \to \R$ be measurable, and let $\psi\colon \R^2\to\R$ be measurable
    \cfload.
    Then 
    \begin{equation}\label{eq:measurable}
    [0,T] \times \R^d \times \Omega \ni (t,x,\omega) \mapsto 
    \sum_{k=0}^K \psi\bpr{ h_k(t), f_k(x,\omega)} 
    \in\R
    \end{equation}
    is measurable
    \cfout.
\end{athm}

\cfclear
\begin{aproof}
    \startnewargseq
    \argument{the fact that for all $k\in\{0,1,\dots,K\}$ it holds that $h_k$, $f_k$, and $\psi$ are measurable}{
    that for all $k\in\{0,1,\dots,K\}$ it holds that 
    $[0,T]\times \R^d\times \Omega \ni (t,x,\omega) = \psi(h_k(t),f_k(x,\omega))$ is measurable\dott{}
    } 
    This establishes \cref{eq:measurable}.
\end{aproof}

\cfclear
\begin{athm}{theorem}{theorem:main}
Let 
$\LipConstF, \constantAssumpMainThm, \alpha_0, \alpha_1, \beta_0, \beta_1, T \in (0,\infty)$, 
$r,p \in[1,\infty)$, 
$\fq \in [2,\infty)$, 
$q \in (2,\infty)$, 
$\activation \in C(\R,\R)$, 
for every $d \in \N_0$ let
$\smallF_d \in C^{\min\{d,1\}}(\R^{\max\{d,1\}},\R)$, 
for every 
$d \in \N$ 
let 
$\nu_d \colon \mathcal{B}(\R^{d+1}) \to [0,\infty)$ 
be a measure with 
\begin{equation}\label{eq:mainthm_assump_measure}
\textstyle{\int_{\R^{d+1}} \pr{ 1+ \norm{y}^{p^2q\fq} } \,\nu_{d}(\dxx y) \le \constantAssumpMainThm d^{rp^2q\fq}}, 
\end{equation}
let $\fJ \in \ANNsm$ satisfy 
$\hiddenLength(\fJ) = 1$
and 
$\Ra(\fJ) = \operatorname{id}_\R$, 
for every $\varepsilon\in(0,1]$ let $\Gamma_{\varepsilon} \in \ANNsm$ 
satisfy for all 
$v, w \in \R$ that 
\begin{equation}\label{eq:assump_on_Gamma_params}
    \textstyle{\Ra(\Gamma_{\varepsilon}) \in C(\R^2,\R)}, \qquad 
    \textstyle{\paramANN(\Gamma_{\varepsilon}) \le \constantAssumpMainThm \varepsilon^{-r} }, 
    \qquad 
    \text{and}
\end{equation}
\begin{equation}\label{eq:assump_on_Gamma_prodapprox}
    \textstyle{\abs{ vw - (\Ra(\Gamma_{\varepsilon}))(v,w) } \le \varepsilon \max\cu{ 1, \abs{v}^q,  \abs{w}^q }}, 
\end{equation}
for every $K\in\N$, $k\in\{0,1,\dots,K\}$, $\varepsilon \in (0,1]$ 
let $\ANNhatfct_{K,k,\varepsilon} \in \ANNsm$  
satisfy for all  
$t\in\R$ that 
\begin{equation}\label{eq:assump_on_ANNforHatfct}
    \textstyle{\Ra(\ANNhatfct_{K,k,\varepsilon}) \in C(\R,\R)}, \qquad 
    \textstyle{\paramANN(\ANNhatfct_{K,k,\varepsilon}) \le \constantAssumpMainThm K^r \varepsilon^{-r} }, 
    \qquad 
    \text{and}
\end{equation}
\begin{equation}\label{eq:assump_on_ANNforHatfct_approx}
    \textstyle{\babs{ \scrL_{\frac{(k-1)T}{K},\frac{kT}{K},\frac{(k+1)T}{K}}^{0,1,0} (t) - (\Ra(\ANNhatfct_{K,k,\varepsilon}))(t) } \le \varepsilon \max\cu{ 1, \abs{t}^q} }, 
\end{equation}
for every $d\in\N_0$, $\varepsilon \in (0,1]$  
let $\interpolatingDNN_{d,\varepsilon} \in\ANNsm$ 
satisfy for all 
$x \in \R^{\max\{d,1\}}$, 
that 
\begin{equation}\label{eq:mainthm_assump_params}
\textstyle{\Ra(\interpolatingDNN_{d,\varepsilon}) \in C(\R^{\max\{d,1\}},\R)}, \quad 
\textstyle{\varepsilon^{\alpha_{\min\{d,1\}}} \lengthANN(\interpolatingDNN_{d,\varepsilon}) 
+
\varepsilon^{\beta_{\min\{d,1\}}} \normmm{\dims(\interpolatingDNN_{d,\varepsilon})} \le \constantAssumpMainThm (\max\{d,1\})^p},
\end{equation}
and
\begin{equation}\label{eq:mainthm_assump_pgrowth}
\begin{split}
&\textstyle{
\varepsilon \norm{\nabla f_{\max\{d,1\}}(x)} 
+ \varepsilon \abs{ (\Ra(\interpolatingDNN_{d,\varepsilon}))(x) } 
+ \abs{ \smallF_d(x) - (\Ra(\interpolatingDNN_{d,\varepsilon}))(x) }} 
\textstyle{\le \varepsilon \constantAssumpMainThm (\max\{d,1\})^p(1 + \norm{x} )^{p}},
\end{split}
\end{equation}
assume for all 
$v, w \in \R$, $\varepsilon \in (0,1]$ that 
\begin{equation}\label{eq:mainthm_assump_Lip}
\textstyle{\max\{ \abs{ \smallF_0(v) - \smallF_0(w) } , \abs{ (\Ra(\interpolatingDNN_{0,\varepsilon}))(v) - (\Ra(\interpolatingDNN_{0,\varepsilon}))(w) } \} \le \LipConstF \abs{v-w}},
\end{equation} 
and let $\delta, \constantMainThmDelta \in (0,\infty)$ satisfy $\constantMainThmDelta = \max\{\alpha_0,\alpha_1\}+2\max\{\beta_0,\beta_1\} + 2\delta + 24r + 8$
\cfload.
Then
\begin{enumerate}[label=(\roman *)]
\item 
\label{mainThm_item1}
for every
$d \in \N$ 
there exists a unique at most polynomially growing viscosity solution
$\smallU_d \in C([0,T]\times\R^d,\R)$
of
\begin{equation}\label{eq:mainThm:heateq}
\tfrac{\partial}{\partial t} \smallU_d(t,x)
+
\tfrac{1}{2} \Delta_x \smallU_d(t,x)
+
\smallF_0\pr{ \smallU_d(t,x) }
=
0
\end{equation}
with $\smallU_d(T,x) = \smallF_d(x)$ for $(t,x) \in (0,T) \times \R^d$
and 
\item 
\label{mainThm_item2}
there exists $\constantMainThmParam \in \R$ such that for all $d \in \N$, $\varepsilon \in (0,1]$ there exists $\U \in \{ \Phi \in \ANNsm \colon \Ra(\Phi) \in C(\R^{d+1},\R) \}$ such that 
\begin{equation}\label{final_item2_error}
	\textstyle{
	\bigl[ \int_{[0,T]\times \R^d}\, \abs{ 
		\smallU_d(y) 
		- 
		\pr{ \Ra\pr{ \U } }(y) 
	}^\fq \, \nu_d(\dxx y)  \bigr]^{\nicefrac{1}{\fq}} 
	\le 
	\varepsilon }
\end{equation}
\begin{equation}\label{final_item2_noCOD}
\begin{split}
       & \text{and} \qquad
        \textstyle{ \paramANN(\U) 
        \le \constantMainThmParam
        d^{
        (2(r+1)p^2 + (p+1)^2 )(6 r+\constantMainThmDelta)q + 3p
        } 
        \varepsilon^{-\constantMainThmDelta} } 
        \dpp
\end{split}
\end{equation}
\end{enumerate}
\end{athm}

\cfclear
\begin{aproof}
\newcommand{\fnl}{f}
\newcommand{\fterm}{f}
\newcommand{\fnlnn}{\interpolatingDNN}
\newcommand{\ftermnn}{\mathbf{F}}
Throughout this proof 
let 
$\fB \in[1,\infty)$, 
$(\littleMM_k)_{k\in\N} \subseteq \N$ 
satisfy for all 
$k \in \N$ 
that 
$\liminf_{j\to\infty} \littleMM_j = \infty$, 
$\limsup_{j\to\infty} \nicefrac{(\littleMM_j)^{q\fq/2}}{j} < \infty$,
and
$\littleMM_{k+1} \le \fB \littleMM_k$,
let $\fd\in \N$
satisfy
$\dims(\fJ) = (1,\fd,1)$,
let $\Theta = \bigcup_{n\in\N} \! \Z^n$,
let $(\Omega, \cF, \P)$ be a probability space,  
let 
$\fu^\theta\colon \Omega \to [0,1]$, $\theta \in \Theta$, 
be i.i.d.\ random variables, 
assume for all $t\in (0,1)$ that $\P(\fu^0\le t)=t$,
let 
$\cU^\theta \colon [0,T] \times \Omega \to [0,T]$, $\theta \in \Theta$, 
satisfy for all 
$t \in [0,T]$, 
$\theta \in \Theta$ 
that 
$\cU_t^\theta = t + (T-t)\fu^\theta$, 
let 
$W^{d,\theta} \colon [0,T] \times  \Omega \to \R^d$, $d \in \N$, $\theta \in \Theta$, 
be independent standard Brownian motions,
assume for every 
$d \in \N$ 
that 
$(\cU^\theta)_{\theta\in\Theta}$ 
and 
$(W^{d,\theta})_{\theta\in\Theta}$ 
are independent,
for every $d, j \in \N$, $\varepsilon \in (0,1]$   
let 
$\mlp_{n,j,\varepsilon}^{d,\theta} \colon [0,T] \times \R^d \times \Omega \to \R$, 
$n \in \N_0$, 
$\theta \in \Theta$,  
satisfy for all  
$\theta \in \Theta$, 
$n \in \N$, 
$t \in [0,T]$, 
$x \in \R^d$, 
$\omega\in\Omega$
that $\mlp_{0,j,\varepsilon}^{d,\theta}(t,x,\omega)=0$ and 
\begin{align}
\mlp_{n,j,\varepsilon}^{d,\theta}(t,x) 
& 
= 
\frac{1}{(\littleMM_{j})^n} 
\br[\Bigg]{ \SmallSum{k=1}{(\littleMM_{j})^n} 
\pr[\big]{ \Ra(\ftermnn_{d,\varepsilon})} \lrSpace \pr[\big]{ x + W_{T-t}^{d,(\theta,0,-k)} } } 
\nonumber
\\
& 
\quad 
+ 
\sum_{i=0}^{n-1} 
\frac{(T-t)}{(\littleMM_{j})^{n-i}} 
\Biggl[ \SmallSum{k=1}{(\littleMM_{j})^{n-i}} 
\Bigl[ \pr[\big]{(\Ra(\fnlnn_{0,\varepsilon})} \lrSpace \pr[\big]{ \mlp_{i,j,\varepsilon}^{d,(\theta,i,k)} \pr[\big]{ \cU_t^{(\theta,i,k)}, x + W_{\cU_t^{(\theta,i,k)}-t}^{d,(\theta,i,k)} } } 
\\
& 
\quad 
- 
\1_{\N}(i) 
\pr[\big]{(\Ra(\fnlnn_{0,\varepsilon})} \lrSpace \pr[\big]{ \mlp_{\max\{i-1,0\},j,\varepsilon}^{d,(\theta,-i,k)} \pr[\big]{ \cU_t^{(\theta,i,k)}, x + W_{\cU_t^{(\theta,i,k)}-t}^{d,(\theta,i,k)} } } \Bigr] \Biggr]
\dc
\nonumber
\end{align}
for every $d, j \in \N$, $\varepsilon \in (0,1]$  
let 
$\U_{n,j,t}^{d,\theta,\varepsilon}\colon \Omega \to \cu{ \Phi \in \ANNsm \colon \functionANN{a}(\Phi) \in C(\R^d,\R) }$, 
$t \in [0,T]$, 
$n \in \N_0$, 
$\theta \in \Theta$, 
satisfy for all  
$\theta \in \Theta$, 
$n \in \N$, 
$t \in [0,T]$, 
$\omega\in \Omega$ 
that 
$\U_{0,j,t}^{d,\theta,\varepsilon}(\omega) = ((0\ 0\ \dots\ 0),0) \in \R^{1\times d}\times \R^1$ 
and
\begin{align}\llabel{mlp_final_th_rep}
\U_{n,j,t}^{d,\theta,\varepsilon} 
& 
= 
\left[ 
\OSum{k=1}{(\littleMM_j)^n} 
\pr[\Big]{  \scalar{\tfrac{1}{(\littleMM_{j})^n}}{\pr[\big]{ \compANN{ \ftermnn_{d,\varepsilon} }{ \A_{\idMatrix_d,W_{T-t}^{d,(\theta,0,-k)}} } } } } 
\right]
\nonumber
\\
& 
\quad 
\bSum_{\,\fJ} 
\left[ 
\BSum{i=0}{\fJ}{n-1} 
\br[\Bigg]{ \scalar{ \pr[\Big]{ \tfrac{(T-t)}{(\littleMM_{j})^{n-i}} } }{ \pr[\bigg]{ \BSum{k=1}{\fJ}{(\littleMM_{j})^{n-i}} \pr[\Big]{ \compANN{ \pr[\big]{ \compANN{\fnlnn_{0,\varepsilon}}{\U^{d,(\theta,i,k),\varepsilon}_{i,j,\mathcal{U}_t^{(\theta,i,k)}} } } }{ \A_{\idMatrix_d, W_{\mathcal{U}_t^{(\theta,i,k)}-t}^{d,(\theta,i,k)}} } } } } } 
\right]
\\
& 
\quad 
\bSum_{\,\fJ} 
\left[ 
\BSum{i=0}{\fJ}{n-1} 
\br[\Bigg]{ \scalar{ \pr[\Big]{ \tfrac{(t-T) \, \1_{\N}(i)}{(\littleMM_{j})^{n-i}} } }{ \pr[\bigg]{ \BSum{k=1}{\fJ}{(\littleMM_{j})^{n-i}} \pr[\Big]{ \compANN{ \pr[\big]{ \compANN{\fnlnn_{0,\varepsilon}}{\U^{d,(\theta,-i,k),\varepsilon}_{\max\{i-1,0\},j,\mathcal{U}_t^{(\theta,i,k)}} } } }{ \A_{\idMatrix_d, W_{\mathcal{U}_t^{(\theta,i,k)}-t}^{d,(\theta,i,k)}} } } } } } 
\right] ,
\nonumber
\end{align} 
for every $K\in\N$, $k\in\Z$ let $\fx_{K,k} \in \R$ satisfy 
$\fx_{K,k} = \tfrac{kT}{K}$, 
for every $d,n,j,K \in\N$, $\theta\in\Theta$, $\varepsilon, \gamma \in (0,1]$ 
let $\Phi_{\gamma,n,j,K}^{d,\theta,\varepsilon}\colon \Omega \to \ANNsm$  satisfy for all 
$\omega \in \Omega$
that 
\begin{equation}
    \Phi_{\gamma,n,j,K}^{d,\theta,\varepsilon}(\omega) = \BSum{k=0}{\fJ}{K} \bbbr{ \compANN{\Gamma_{\gamma}}{\parallelization_{2,\fJ}\bbpr{\bpr{ \ANNhatfct_{K,k,\gamma}, \U_{n,j,\fx_{K,k}}^{d,\theta,\varepsilon}(\omega)  } }} } , 
\end{equation}
assume without loss of generality that
$\max\{ \abs{\fnl_0(0)} , \fd, 1 \}
\le \constantAssumpMainThm$,  
let $(\fa_d)_{d \in \N}$ satisfy for all $d\in\N$ that 
\begin{equation}\llabel{eq:proofmainthm_constant_a}
    \fa_d = e^{2 \LipConstF T} (T+1)^{2} (\LipConstF +1) (\constantAssumpMainThm d^p +1) 8^{p +2} d^{\frac12},
\end{equation}
let $(\fb_d)_{d\in\N}$ satisfy for all $d\in\N$ that 
\begin{equation}\llabel{eq:proofmainthm_constant_b}
    \fb_d = 2^{p-1} \constantAssumpMainThm d^p \bpr{ e^{\LipConstF T} (T+1) }^{p+1} \bpr{(\constantAssumpMainThm d^p)^p +1} 3^{p^2 -1} ,
\end{equation}
let $(\fc_d)_{d \in \N}$ satisfy for all 
$d \in \N$ that 
\begin{equation}\llabel{eq:proofmainthm_constant_c}
    \fc_d = 
    2^p \constantAssumpMainThm d^p (T+1) e^{\LipConstF T} \sqrt{q\fq-1},
\end{equation}
let $(\fm_{j,n})_{(j,n) \in \N^2}$ satisfy for all 
$j, n \in \N$ that 
\begin{equation}\llabel{eq:proofmainthm_constant_m}
    \fm_{j,n} = 
    \bbbbr{ (1+2\LipConstF T)  \pr{\littleMM_j}^{-\frac12} \exp\bbbpr{\frac{\pr{\littleMM_j}^{\frac{q\fq}{2}}}{n}} }^n,
\end{equation}
let $(c_d)_{d \in \N}$ satisfy for all 
$d \in \N$ that 
\begin{equation}\llabel{eq:proofmainthm_constant_BM}
    c_d =  \bbbbbr{ \int_{[0,T] \times \R^d}  
        \bbpr{ 1 + \norm{x}^{p^2 q \fq} + \sup_{s \in [0,T]} \E\bbr{ \norm{ W_s^{d,0} }^{2 p^2 q \fq} } } 
        \nu_d(\dx t, \dx x)  }^{\nicefrac{1}{\fq}} ,
\end{equation} 
let 
$(N_{\varepsilon})_{\varepsilon \in (0,1]}$ satisfy for all $\varepsilon\in(0,1]$ that
\begin{equation}\llabel{eq:proofmainthm_N}
\begin{split}
N_{\varepsilon}
& = 
\inf\bcu{ n \in \N \colon \fm_{n,n} \le \varepsilon } ,
\end{split}
\end{equation}
let $(K_{\varepsilon})_{\varepsilon \in(0,1]}$ satisfy for all $\varepsilon\in(0,1]$ that
\begin{equation}\llabel{eq:proofmainthm_K}
\begin{split}
K_{\varepsilon}
& = 
\inf\bcu{ n \in \N \colon n \ge \varepsilon^{-2} }  ,
\end{split}
\end{equation}
let $(\delta_{d,\varepsilon})_{(d,\varepsilon) \in \N\times (0,1]}$ satisfy for all $d\in\N$, $\varepsilon\in(0,1]$ that
\begin{equation}\llabel{eq:proofmainthm_delta}
\begin{split}
\delta_{d,\varepsilon}
& = 
\frac{\varepsilon}{1+ 180 (T+1) 2^p (\fa_d+\fb_d+\fc_d) c_d} ,
\end{split}
\end{equation}
and
let $(\gamma_{d,\varepsilon})_{(d,\varepsilon) \in \N\times (0,1]}$ satisfy for all $d\in\N$, $\varepsilon\in(0,1]$ that
\begin{equation}\llabel{eq:proofmainthm_gamma}
\begin{split}
\gamma_{d,\varepsilon}
& = \frac{\varepsilon}{2 \constantAssumpMainThm d^{rp^2 q}(K_{\varepsilon}+1) \bpr{1+(T+1)^q}^q \bpr{ 1 + 3^{3p^2 q} c_d \bpr{ \pr{\fc_d}^q + (T+1)^q \pr{\pr{\fa_d}^q + \pr{\fb_d}^q} + \constantAssumpMainThm^q d^{pq} } }} 
\end{split}
\end{equation}
\cfload.
\startnewargseq
\argument{
the triangle inequality;
the assumption that for all 
$d \in \N$, 
$x \in \R^d$, 
$\varepsilon \in (0,1]$ 
it holds that 
$\varepsilon \abs{(\Ra(\ftermnn_{d,\varepsilon}))(x)} + \abs{\fterm_d(x) - (\Ra(\ftermnn_{d,\varepsilon}))(x)} \le \varepsilon \constantAssumpMainThm d^p (1 + \norm{x})^{p}$
}{
that for all 
$d \in \N$, 
$x \in \R^d$ 
it holds that
\begin{equation}\llabel{eq:fd_bound}
\abs{\fterm_d(x)} 
\le 
\abs{\fterm_d(x) - (\Ra(\ftermnn_{d,1}))(x)} + \abs{(\Ra(\ftermnn_{d,1}))(x)} 
\le 
\constantAssumpMainThm d^p(1 + \norm{x})^{p}
\dott
\end{equation}
}
\argument{
\lref{eq:fd_bound}; 
the assumption that for all 
$w, z \in \R$ 
it holds that 
$\abs{\fnl_0(w) - \fnl_0(z)} \le \LipConstF \abs{w-z}$;
Beck et al.\ \cite[Theorem~1.1]{beck2021nonlinear}
(applied for every 
$d \in \N$
with
$d \with d$,
$m \with d$,
$L \with \LipConstF + 1$,
$T \with T$,
$\mu \with (\R^d \ni x \mapsto (0,0,\dots,0) \in \R^d)$,
$\sigma \with (\R^d \ni x \mapsto \idMatrix_d \in\R^{d\times d})$,
$f \with (\R^d\times\R \ni (x,w) \mapsto \fnl_0(w) \in \R)$,
$g \with \fterm_d$,
$W \with \fwpr^{d,0}$
in the notation of 
Beck et al.\ \cite[Theorem~1.1]{beck2021nonlinear}); 
Fubini's theorem}{
that for every
$d \in \N$ 
there exists a unique at most polynomially growing viscosity solution
$\smallU_d \in C([0,T]\times\R^d,\R)$
of \cref{eq:mainThm:heateq} 
with $\smallU_d(T,x) = \smallF_d(x)$, $x \in \R^d$,
and that it moreover holds for all $d \in \N$, $t \in [0,T]$, $x \in \R^d$ 
that $\E\br{\abs{f_d(x+W_{T-t}^{d,0})} + \int_t^T \abs{f_0(u_d(s,x+W_{s-t}^{d,0}))} \dx s  } < \infty$
and 
\begin{equation}\llabel{eq:1249}
    u_d(t,x) = \E\bbr{ f_d(x+W^{d,0}_{T-t}) } 
    + 
    \int_t^T \E\bbr{ f_0\bpr{u_{d}(s,x+W^{d,0}_{s-t})} } \dx s 
    \dott
\end{equation}
}
This proves \cref{mainThm_item1}. 
\startnewargseq
\argument{
the assumption that for all 
$w, z \in \R$, $\varepsilon \in (0,1]$
it holds that 
$\abs{ (\Ra(\fnlnn_{0,\varepsilon}))(z) - (\Ra(\interpolatingDNN_{0,\varepsilon}))(w) } \le \LipConstF \abs{z-w}$;
the assumption that for all 
$d\in \N_0$, $x \in \R^{\max\{d,1\}}$, $\varepsilon \in (0,1]$  
it holds that 
$ \abs{ (\Ra(\ftermnn_{d,\varepsilon}))(x) }\le \constantAssumpMainThm (\max\{d,1\})^p(1 + \norm{x} )^{p}
$;
Beck et al.\ \cite[Theorem~1.1]{beck2021nonlinear}
(applied for every 
$d \in \N$, $\varepsilon\in (0,1]$
with
$d \with d$,
$m \with d$,
$L \with \LipConstF +1$,
$T \with T$,
$\mu \with (\R^d \ni x \mapsto (0,0,\dots,0) \in \R^d)$,
$\sigma \with (\R^d \ni x \mapsto \idMatrix_d \in\R^{d\times d})$,
$f \with (\R^d\times\R \ni (x,w) \mapsto (\Ra(\fnlnn_{0,\varepsilon}))(w) \in \R)$,
$g \with \Ra(\ftermnn_{d,\varepsilon})$,
$W \with \fwpr^{d,0}$
in the notation of 
Beck et al.\ \cite[Theorem~1.1]{beck2021nonlinear})
}{
that for every $d \in \N$, $\varepsilon \in (0,1]$ there exists a unique at most polynomially growing
$v_{d,\varepsilon} \in C([0,T]\times \R^d, \R)$ such that
for all
$t \in [0,T]$, 
$x \in \R^d$ 
it holds that 
$\E\br{\abs{\pr{\Ra(\ftermnn_{d,\varepsilon})}(x+W^{d,0}_{T-t})} + \int_t^T \abs{ \pr{\Ra(\interpolatingDNN_{0,\varepsilon})}\pr{v_{d,\varepsilon}(s,x+W^{d,0}_{s-t})} } \dx s  } < \infty$
and 
\begin{equation}\llabel{eq:soln_PDE_ANNfcts}
v_{d,\varepsilon}(t,x) 
= 
\E\bbr{ \bpr{\Ra(\ftermnn_{d,\varepsilon})}(x+W^{d,0}_{T-t}) } 
+ 
\int_t^T \E\bbr{ \bpr{\Ra(\interpolatingDNN_{0,\varepsilon})}\bpr{v_{d,\varepsilon}(s,x+W^{d,0}_{s-t})} } \dx s 
\dott
\end{equation}
}
\startnewargseq 
\argument{item~(i) in Lemma~3.3 in Hutzenthaler et al.~\cite{PadgettJentzen2021}}{
that for all $\varepsilon \in (0,1]$, $n\in\N_0$, $d,j\in\N$, $\theta \in \Theta$ it holds that $\mlp_{n,j,\varepsilon}^{d,\theta}\colon [0,T]\times \R^d\times \Omega \to \R$ 
is 
measurable\dott\,
}
\Hence 
that for all $\varepsilon \in (0,1]$, $n\in\N_0$, $d,j\in\N$, $\theta \in \Theta$, $t\in[0,T]$ 
$\R^d\times \Omega 
\ni (x,\omega) \mapsto \mlp_{n,j,\varepsilon}^{d,\theta}(t,x,\omega) \in \R$ 
is 
measurable. 
Combining this, 
the fact that for all $\gamma \in (0,1]$ it holds that $\Ra(\Gamma_{\gamma}) \in C(\R^2,\R)$,  
the fact that for all $K\in\N$, $k\in\{0,1,\dots,K\}$, $\gamma \in (0,1]$ it holds that $\Ra(\ANNhatfct_{K,k,\gamma})\in C(\R,\R)$,  
\cref{ANN_for_MLP2_item1} in \cref{ANN_for_MLP2}, 
and \cref{lin_interpolation_measurable}
proves 
that for all $\gamma, \varepsilon \in (0,1]$, $n,j,K,d \in\N$, $\theta \in \Theta$ 
it holds that 
$[0,T]\times\R^d\times \Omega \ni (t,x,\omega) \mapsto (\functionANN{a}\pr{\Phi_{\gamma,n,j,K}^{d,\theta,\varepsilon}(\omega)})(t,x) \in \R$
is 
measurable. 
\argument{the fact that for all $\varepsilon \in (0,1]$, $n\in\N_0$, $d,j\in\N$, $\theta \in \Theta$, $t\in[0,T]$ 
it holds that 
$\R^d\times \Omega 
\ni (x,\omega) \mapsto \mlp_{n,j,\varepsilon}^{d,\theta}(t,x,\omega) \in \R$ 
is measurable; 
the fact that for all $K\in\N$, $k\in\{0,1,\dots,K\}$ it holds that $[0,T] \ni t \mapsto \scrL_{\fx_{K,k-1},\fx_{K,k},\fx_{K,k+1}}^{0,1,0}(t) \in \R$ is continuous; 
\cref{factorize_lin_interpolation};
\cref{lin_interpolation_measurable}}{
that for all $\varepsilon \in (0,1]$, $n,j,K,d \in\N$, $\theta \in \Theta$ 
it holds that 
\begin{equation}
    [0,T]\times \R^d \times \Omega \ni (t,x,\omega)
    \mapsto  \scrL_{\fx_{K,0},\fx_{K,1},\dots,\fx_{K,K}}^{\mlp_{n,j,\varepsilon}^{d,\theta}(\fx_{K,0},x,\omega),\mlp_{n,j,\varepsilon}^{d,\theta}(\fx_{K,1},x,\omega),\dots,\mlp_{n,j,\varepsilon}^{d,\theta}(\fx_{K,K},x,\omega)}(t) 
    \in \R
\end{equation}
is measurable\dott 
}
\startnewargseq 
\argument{the triangle inequality}{
that for all 
$d,K,j,n \in \N$, $\varepsilon,\gamma \in (0,1]$ 
it holds that 
\begin{equation}\llabel{eq:0001a}
    \begin{split}
        & \bbbbr{ \int_{[0,T] \times \R^d} \E\bbbr{ \babs{ u_d(t,x) - \bpr{\functionANN{a}\bpr{\Phi_{\gamma,n,j,K}^{d,0,\varepsilon}}}(t,x) }^{\fq} } \nu_d(\dx t, \dx x)  }^{\nicefrac{1}{\fq}} \\
        & \le 
        \bbbbr{ \int_{[0,T] \times \R^d} \E\bbbr{ \babs{ u_d(t,x) - \scrL_{\fx_{K,0},\dots,\fx_{K,K}}^{\mlp_{n,j,\varepsilon}^{d,0}(\fx_{K,0},x),\dots,\mlp_{n,j,\varepsilon}^{d,0}(\fx_{K,K},x)}(t) }^{\fq} } \nu_d(\dx t, \dx x)  }^{\nicefrac{1}{\fq}} \\
        & \quad + \bbbbr{ \int_{[0,T] \times \R^d} \E\bbbr{ \babs{ \scrL_{\fx_{K,0},\dots,\fx_{K,K}}^{\mlp_{n,j,\varepsilon}^{d,0}(\fx_{K,0},x),\dots,\mlp_{n,j,\varepsilon}^{d,0}(\fx_{K,K},x)}(t) - \bpr{\functionANN{a}\bpr{\Phi_{\gamma,n,j,K}^{d,0,\varepsilon}}}(t,x) }^{\fq} } \nu_d(\dx t, \dx x)  }^{\nicefrac{1}{\fq}}
        \dott
    \end{split}
\end{equation}
}
\argument{the triangle inequality;
Jensen's inequality}{
that for all 
$d,K,j,n \in \N$, $\varepsilon \in (0,1]$, $k\in\{1,2,\dots,K\}$, $t\in [0,T]$, $x\in\R^d$ 
it holds that 
\begin{equation}\llabel{eq:1112a}
    \begin{split}
        &\E\bbbr{ \babs{ u_d(t,x) - \scrL_{\fx_{K,0},\dots,\fx_{K,K}}^{\mlp_{n,j,\varepsilon}^{d,0}(\fx_{K,0},x),\dots,\mlp_{n,j,\varepsilon}^{d,0}(\fx_{K,K},x)}(t) }^{\fq} } \\
        & \le 4^{\fq-1}  
        \bigg(
        \abs{ u_d(t,x) - u_d(\fx_{K,k},x) }^{\fq}
        + \abs{ u_d(\fx_{K,k},x) - v_{d,\varepsilon}(\fx_{K,k},x) }^{\fq} \\
        & \qquad \qquad + 
        \E\bbbr{ \abs{ v_{d,\varepsilon}(\fx_{K,k},x) - \mlp_{n,j,\varepsilon}^{d,0}(\fx_{K,k},x) }^{\fq} } \\
        & \qquad \qquad + \E\bbbr{ \babs{ \mlp_{n,j,\varepsilon}^{d,0}(\fx_{K,k},x) - \scrL_{\fx_{K,0},\dots,\fx_{K,K}}^{\mlp_{n,j,\varepsilon}^{d,0}(\fx_{K,0},x),\dots,\mlp_{n,j,\varepsilon}^{d,0}(\fx_{K,K},x)}(t) }^{\fq} } \bigg) 
        \dott 
    \end{split}
\end{equation}
}
\argument{\cref{lin_interp1};
the triangle inequality;
Jensen's inequality}{
that for all 
$d,K,j,n \in \N$, $\varepsilon \in (0,1]$, $k\in\{1,2,\dots,K\}$, $x \in \R^d$, $t\in[\fx_{K,k-1},\fx_{K,k}]$  
it holds that 
\begin{equation}\llabel{eq:1112b}
    \begin{split}
        & \E\bbbr{ \babs{ \mlp_{n,j,\varepsilon}^{d,0}(\fx_{K,k},x) - \scrL_{\fx_{K,0},\dots,\fx_{K,K}}^{\mlp_{n,j,\varepsilon}^{d,0}(\fx_{K,0},x),\dots,\mlp_{n,j,\varepsilon}^{d,0}(\fx_{K,K},x)}(t) }^{\fq} } \\
        & \le 
        \E\bbr{
        \abs{ \mlp_{n,j,\varepsilon}^{d,0}(\fx_{K,k},x) - \mlp_{n,j,\varepsilon}^{d,0}(\fx_{K,k-1},x) }^{\fq} 
        } \\
        & \le 5^{\fq-1}
        \bigg(
        \E\bbr{
        \abs{ \mlp_{n,j,\varepsilon}^{d,0}(\fx_{K,k},x) - v_{d,\varepsilon}(\fx_{K,k},x) }^{\fq} } 
        + \abs{ v_{d,\varepsilon}(\fx_{K,k},x) - u_d(\fx_{K,k},x) }^{\fq}  \\
        & \qquad \qquad + \abs{ u_d(\fx_{K,k},x) - u_d(\fx_{K,k-1},x) }^{\fq} 
        + \abs{ u_d(\fx_{K,k-1},x) - v_{d,\varepsilon}(\fx_{K,k-1},x) }^{\fq}  \\
        & \qquad \qquad + \E\bbr{
        \abs{ v_{d,\varepsilon}(\fx_{K,k-1},x) - \mlp_{n,j,\varepsilon}^{d,0}(\fx_{K,k-1},x) }^{\fq} } 
        \bigg)
        \dott 
    \end{split}
\end{equation}
}
\argument{\lref{eq:1112a};\lref{eq:1112b}}{
that for all 
$d,K,j,n \in \N$, $\varepsilon \in (0,1]$, $k\in\{1,2,\dots,K\}$, $x \in \R^d$, $t\in[\fx_{K,k-1},\fx_{K,k}]$  
it holds that 
\begin{equation}\llabel{eq:diff_u_interpol}
    \begin{split}
        &\E\bbbr{ \babs{ u_d(t,x) - \scrL_{\fx_{K,0},\dots,\fx_{K,K}}^{\mlp_{n,j,\varepsilon}^{d,0}(\fx_{K,0},x),\dots,\mlp_{n,j,\varepsilon}^{d,0}(\fx_{K,K},x)}(t) }^{\fq} } \\
        & \le 20^{\fq-1}  
        \bigg(
        \abs{ u_d(t,x) - u_d(\fx_{K,k},x) }^{\fq}
        + 2 \abs{ u_d(\fx_{K,k},x) - v_{d,\varepsilon}(\fx_{K,k},x) }^{\fq} \\
        & \qquad \qquad + \abs{ u_d(\fx_{K,k},x) - u_d(\fx_{K,k-1},x) }^{\fq} 
        + \abs{ u_d(\fx_{K,k-1},x) - v_{d,\varepsilon}(\fx_{K,k-1},x) }^{\fq}  \\
        & \qquad \qquad 
        + 2 \E\bbr{ \abs{ v_{d,\varepsilon}(\fx_{K,k},x) - \mlp_{n,j,\varepsilon}^{d,0}(\fx_{K,k},x) }^{\fq} }
        + \E\bbr{
        \abs{ v_{d,\varepsilon}(\fx_{K,k-1},x) - \mlp_{n,j,\varepsilon}^{d,0}(\fx_{K,k-1},x) }^{\fq} } 
        \bigg) \\
        & \le 
        20^{\fq-1} \bigg(
        2 \bbpr{ \sup\nolimits_{s \in [\fx_{K,k-1},\fx_{K,k}]} \abs{ u_d(s,x) - u_d(\fx_{K,k},x) }^{\fq} }
        + 3 \bbpr{ \sup\nolimits_{s \in [0,T]} \abs{ u_d(s,x) - v_{d,\varepsilon}(s,x) }^{\fq} }\\
        & \qquad \qquad + 
        3 \bbpr{ \sup\nolimits_{s \in [0,T]} \E\bbr{
        \abs{ v_{d,\varepsilon}(s,x) - \mlp_{n,j,\varepsilon}^{d,0}(s,x) }^{\fq} } }
        \bigg)
        \dott 
    \end{split}
\end{equation}
}
\argument{\lref{eq:fd_bound}; \lref{eq:1249}; the fact that for all $w,z \in \R$ it holds that $\abs{f_0(z) - f_0(w)}\le \LipConstF \abs{z-w}$; 
the fact that for all $d\in\N$, $x \in \R^d$ it holds that $\norm{\nabla f_d(x)}\le\constantAssumpMainThm d^p (1+\norm{x})^{p}$; 
\cref{temp_u_reg1} (applied for every $d \in \N$ 
with  $u\with u_d$, $T\with T$, $d\with d$, $W \with W^{d,0}$, $g \with f_d$, $f\with ([0,T]\times \R^d\times \R \ni (t,x,z) \mapsto f_0(z) \in\R)$, $\LipConstF\with \LipConstF$, $\boundFG \with \constantAssumpMainThm d^p$, $p \with p$ 
in the notation of \cref{temp_u_reg1}); 
\lref{eq:proofmainthm_constant_a}; 
Jensen's inequality}{
that for all 
$d,K \in \N$, $k\in\{1,2,\dots,K\}$, $x \in \R^d$ 
it holds that 
\begin{equation}\llabel{eq:diff_u_interpol_term1}
    \begin{split}
        \sup_{s \in [\fx_{K,k-1},\fx_{K,k}]} \abs{ u_d(s,x) - u_d(\fx_{K,k},x) }^{\fq}
        & \le         
        \bpr{\fx_{K,k}-\fx_{K,k-1}}^{\frac{\fq}{2}} d^{\frac{\fq}{2}} 
        \bpr{e^{2 \LipConstF T} (T+1)^{2} (\LipConstF +1) (\constantAssumpMainThm d^p +1)}^{\fq}
        \\
        & \quad \cdot 8^{(p + 2)\fq} \bpr{ 1 + \norm{x}^{p} + \sup\nolimits_{s \in [0,T]} \E\bbr{\norm{W_s^{d,0}}^{2 p}} }^{\fq} \\
        & \le
        \bbbbr{\frac{T}{K}}^{\frac{\fq}{2}}  
        (\fa_d)^{\fq}  3^{\fq-1}
        \bpr{ 1 + \norm{x}^{p \fq} + \sup\nolimits_{s \in [0,T]} \E\bbr{\norm{W_s^{d,0}}^{2 p \fq}} } \\
        & \le 
        \bbbbr{\frac{T}{K}}^{\frac{\fq}{2}}  
        (\fa_d)^{\fq}  3^{\fq}
        \bpr{ 1 + \norm{x}^{p^2 q \fq} + \sup\nolimits_{s \in [0,T]} \E\bbr{\norm{W_s^{d,0}}^{2 p^2 q \fq}} }
        \dott 
    \end{split}
\end{equation}
}
\argument{\cref{eq:mainthm_assump_pgrowth}; Jensen's inequality}{
that for all 
$d \in \N$, $\varepsilon \in (0,1]$, $z \in \R$, $x\in\R^d$   
it holds that 
\begin{equation}\llabel{eq:1059a}
    \abs{ f_0(z) - (\Ra(\interpolatingDNN_{0,\varepsilon}))(z) }
    \le \varepsilon \constantAssumpMainThm d^p (1+\abs{z})^p
    \le \varepsilon 2^{p-1} \constantAssumpMainThm d^p \bpr{ (1+\norm{x})^{p^2} +\abs{z}^p } 
    \dott
\end{equation}
}
\argument{\cref{eq:mainthm_assump_pgrowth}}{
that for all 
$d \in \N$, $\varepsilon \in (0,1]$, $z \in \R$, $x\in\R^d$   
it holds that 
\begin{equation}\llabel{eq:1059b}
    \abs{ f_d(x) - (\Ra(\ftermnn_{d,\varepsilon}))(x) }
    \le \varepsilon \constantAssumpMainThm d^p (1+\norm{x})^p
    \le \varepsilon 2^{p-1} \constantAssumpMainThm d^p \bpr{ (1+\norm{x})^{p^2} +\abs{z}^p } \dott
\end{equation}
}
\argument{\lref{eq:1059a}; \lref{eq:1059b}; \cref{eq:mainthm_assump_Lip}; \lref{eq:fd_bound}; \lref{eq:1249};  \lref{eq:soln_PDE_ANNfcts}; 
the assumption that for all $d \in \N_0$, $x \in \R^{\max\{d,1\}}$, $\varepsilon\in (0,1]$ it holds that $\abs{(\Ra(\interpolatingDNN_{d,\varepsilon}))(x)} \le \constantAssumpMainThm (\max\{d,1\})^p (1+\norm{x})^p$; 
\cref{sol_stab2} (applied for every $d\in\N$, $\varepsilon \in (0,1]$ with 
$u_1 \with u_d$, $u_2 \with v_{d,\varepsilon}$,
$T\with T$, $d\with d$, $W\with W^{d,0}$, $\LipConstF \with \LipConstF$, $\boundFG \with \constantAssumpMainThm d^p$, 
$\diffBd \with \varepsilon 2^{p-1} \constantAssumpMainThm d^p$, $p\with p$, $q\with p$, 
$f_1 \with ([0,T] \times \R^d \times \R \ni (t,x,z) \mapsto f_0(z) \in\R)$, 
$f_2 \with ([0,T] \times \R^d \times \R \ni (t,x,z) \mapsto 
(\Ra(\interpolatingDNN_{0,\varepsilon}))(z) \in\R)$, 
$g_1 \with f_d$, $g_2 \with \Ra(\ftermnn_{d,\varepsilon})$ 
in the notation of \cref{sol_stab2}); \lref{eq:proofmainthm_constant_b}}{
that for all 
$d \in \N$, $\varepsilon \in (0,1]$, $x \in \R^d$ 
it holds that 
\begin{equation}\llabel{eq:sup_diff_ud_vdeps}
    \begin{split}
        \sup_{s \in [0,T]} \abs{ u_d(s,x) - v_{d,\varepsilon}(s,x) }
        & \le \varepsilon \fb_d 
        \bpr{ 1 + \norm{x}^{p^2} + \sup\nolimits_{s\in[0,T]} \E\bbr{ \norm{ W_s^{d,0} }^{p^2} } }
        \dott
    \end{split}
\end{equation}
}
\argument{\lref{eq:sup_diff_ud_vdeps}; Jensen's inequality}{
that for all 
$d \in \N$, $\varepsilon \in (0,1]$, $x \in \R^d$ 
it holds that 
\begin{equation}\llabel{eq:diff_u_interpol_term2}
    \begin{split}
        \sup_{s \in [0,T]} \abs{ u_d(s,x) - v_{d,\varepsilon}(s,x) }^{\fq} 
        & \le 
        \varepsilon^{\fq} 
        (\fb_d)^{\fq} 3^{\fq}
        \bpr{ 1 + \norm{x}^{p^2 q \fq} + \sup\nolimits_{s \in [0,T]} \E\bbr{ \norm{ W_s^{d,0} }^{2 p^2 q \fq} } } 
        \dott
    \end{split}
\end{equation}
}
\argument{\lref{eq:soln_PDE_ANNfcts}; the fact that for all $\varepsilon \in (0,1]$, $w, z \in \R$ it holds that $\abs{(\Ra(\interpolatingDNN_{0,\varepsilon}))(w) - (\Ra(\interpolatingDNN_{0,\varepsilon}))(z)} \le \LipConstF \abs{w-z}$;
the fact that for all $\varepsilon \in (0,1]$, $d\in\N_0$, $x \in \R^{\max\{d,1\}}$ it holds that $\abs{ (\Ra(\interpolatingDNN_{d,\varepsilon}))(x) } \le \constantAssumpMainThm (\max\{d,1\})^p (1+\norm{x})^p $;
Hutzenthaler et al.~\cite[Corollary~3.15]{PadgettJentzen2021} (applied for every $d\in\N$, $\varepsilon \in (0,1]$, $j\in\N$ with 
$\fp \with \fq$, $d \with d$, $m\with \littleMM_j$, $T\with T$, $\LipConstF \with \LipConstF$, $\fL \with 2^{p-1} \constantAssumpMainThm d^p$, $p\with p$, 
$u \with v_{d,\varepsilon}$, $f\with ([0,T] \times \R^d \times \R \ni (t,x,z)\mapsto (\Ra(\interpolatingDNN_{0,\varepsilon}))(z) \in\R)$,
$g \with \Ra(\ftermnn_{d,\varepsilon})$, 
$\Theta \with \Theta$, 
$(W^{\theta})_{\theta \in \Theta} \with (W^{d,\theta})_{\theta \in \Theta}$, 
$(U_{n}^{\theta})_{(n,\theta)\in\Z \times \Theta}\with (\mlp_{n,j,\varepsilon}^{d,\theta})_{(n,\theta) \in \Z \times \Theta}$, 
$(\fu^\theta)_{\theta \in \Theta} \with (\fu^\theta)_{\theta \in \Theta}$, 
$(\cU^\theta)_{\theta \in \Theta} \with (\cU^\theta)_{\theta \in \Theta}$ 
in the notation of Hutzenthaler et al.~\cite[Corollary~3.15]{PadgettJentzen2021});
\lref{eq:proofmainthm_constant_c}}{
that for all 
$d,j,n \in \N$, $\varepsilon \in (0,1]$, $x \in \R^d$ 
it holds that 
\begin{equation}\llabel{eq:diff_u_interpol_term3}
    \begin{split}
        & \sup_{s \in [0,T]} \E\bbr{
        \abs{ v_{d,\varepsilon}(s,x) - \mlp_{n,j,\varepsilon}^{d,0}(s,x) }^{\fq} } \\
        & \le 
        (\fc_d)^{\fq} 
        \bbbbr{(1+2\LipConstF T)^n  
        (\littleMM_j)^{-\frac{n}{2}} \exp\bbbpr{ \frac{(\littleMM_j)^{\frac{\fq}{2}}}{\fq} }
        }^{\fq} 
        \bbbpr{ \sup_{s \in [0,T]} \E\bbr{( 1 + \norm{ x + W_s^{d,0} }^p )^{\fq}} }
        \dott 
    \end{split}
\end{equation}
}
\argument{Jensen's inequality; the triangle inequality}{
that for all $d \in \N$, $x \in \R^d$, $s \in [0,T]$ it holds that 
\begin{equation}\llabel{eq:2307}
    \begin{split}
        \E\bbr{( 1 + \norm{ x + W_s^{d,0} }^p )^{\fq}}
        & \le 2^{\fq-1} \bbpr{ 1 + 2^{p\fq -1} \bpr{ \norm{x}^{p\fq} + \E\bbr{ \norm{ W_s^{d,0} }^{p\fq} } }  } \\
        & \le 2^{\fq-1} \bbpr{ 1 + 2^{p\fq-1} \bpr{ \norm{x}^{p^2 q\fq} + \E\bbr{ \norm{ W_s^{d,0} }^{2p^2 q\fq} } } + 2^{p\fq} }  \\
        & \le 2^{(p+1)\fq} \bbpr{ 1 + \norm{x}^{p^2 q\fq} + \E\bbr{ \norm{ W_s^{d,0} }^{2p^2 q\fq} }  }
        \dott 
    \end{split}
\end{equation}
}
\argument{\lref{eq:proofmainthm_constant_m}}{
that for all $j,n\in\N$ it holds that 
\begin{equation}\llabel{eq:1335}
    \begin{split}
        (1+2\LipConstF T)^n  
        (\littleMM_j)^{-\frac{n}{2}} \exp\bbbpr{ \frac{(\littleMM_j)^{\frac{\fq}{2}}}{\fq} } 
        & \le \bbbbr{ (1+2\LipConstF T) (\littleMM_j)^{-\frac12} \exp\bbbpr{ \frac{(\littleMM_j)^{\frac{q\fq}{2}}}{n} } }^n
        = \fm_{j,n} 
        \dott
    \end{split}
\end{equation}
}
\argument{\lref{eq:diff_u_interpol_term3}; \lref{eq:2307}; \lref{eq:1335}}{
that for all 
$d,j,n \in \N$, $\varepsilon \in (0,1]$, $x \in \R^d$ 
it holds that 
\begin{equation}\llabel{eq:diff_u_interpol_term3b}
    \begin{split}
        \sup_{s \in [0,T]} \E\bbr{
        \abs{ v_{d,\varepsilon}(s,x) - \mlp_{n,j,\varepsilon}^{d,0}(s,x) }^{\fq} } 
        & \le 
        (\fc_d)^{\fq} 
        (\fm_{j,n})^{\fq} 
        2^{(p+1)\fq} \bbbpr{ 1 + \norm{x}^{p^2 q\fq} + \sup_{s\in[0,T]} \E\bbr{ \norm{ W_s^{d,0} }^{2p^2 q\fq} }  }
        \dott 
    \end{split}
\end{equation}
}
\argument{\lref{eq:diff_u_interpol};\lref{eq:diff_u_interpol_term1}; \lref{eq:diff_u_interpol_term2}; \lref{eq:diff_u_interpol_term3b}}{
that for all 
$d,K,j,n \in \N$, $\varepsilon \in (0,1]$, $x \in \R^d$, $t\in[0,T]$  
it holds that 
\begin{equation}\llabel{eq:20290}
    \begin{split}
        & \E\bbbr{ \babs{ u_d(t,x) - \scrL_{\fx_{K,0},\dots,\fx_{K,K}}^{\mlp_{n,j,\varepsilon}^{d,0}(\fx_{K,0},x),\dots,\mlp_{n,j,\varepsilon}^{d,0}(\fx_{K,K},x)}(t) }^{\fq} } \\
        & \le 
        20^{\fq-1} \bbbbpr{
        2\bbbbr{\frac{T}{K}}^{\frac{\fq}{2}}  
        (\fa_d)^{\fq}  3^{\fq}
        + 3 \varepsilon^{\fq} (\fb_d)^{\fq} 3^{\fq}
        + 3 (\fc_d)^{\fq}
        (\fm_{j,n})^{\fq}
        2^{(p+1)\fq} } \\
        & \quad \cdot 
        \bbbpr{ 1 + \norm{x}^{p^2 q \fq} + \sup_{s \in [0,T]} \E\bbr{ \norm{ W_s^{d,0} }^{2 p^2 q \fq} } } 
        \dott 
    \end{split}
\end{equation} 
}
\argument{\lref{eq:20290}; the triangle inequality; \lref{eq:proofmainthm_constant_BM}}{
that for all $d,K,j,n \in \N$, $\varepsilon \in (0,1]$
it holds that 
\begin{equation}\llabel{eq:1334}
    \begin{split}
        & \bbbbr{ \int_{[0,T] \times \R^d} \E\bbbr{ \babs{ u_d(t,x) - \scrL_{\fx_{K,0},\dots,\fx_{K,K}}^{\mlp_{n,j,\varepsilon}^{d,0}(\fx_{K,0},x),\dots,\mlp_{n,j,\varepsilon}^{d,0}(\fx_{K,K},x)}(t) }^{\fq} } \nu_d(\dx t, \dx x)  }^{\nicefrac{1}{\fq}} \\
        & \le 20 
        \bpr{ 6 (T+1) \fa_d K^{-\frac12} 
        + 9 \varepsilon \fb_d + 2^p 6 \fc_d  
        \fm_{j,n} 
        }
        c_d 
        \dott
    \end{split}
\end{equation}
}
\argument{\lref{eq:1334}; \lref{eq:proofmainthm_constant_m}; \lref{eq:proofmainthm_N}; \lref{eq:proofmainthm_K};  \cref{lemma_littleMM_item1} in \cref{lemma_littleMM}}{
that 
for all $d \in \N$, $\varepsilon \in (0,1]$
it holds that 
\begin{equation}\llabel{eq:0001b}
    \begin{split}
        & \bbbbr{ \int_{[0,T] \times \R^d} \E\bbbr{ \babs{ u_d(t,x) - \scrL_{\fx_{K_{\varepsilon},0},\dots,\fx_{K_{\varepsilon},K_{\varepsilon}}}^{\mlp_{N_{\varepsilon},N_{\varepsilon},\varepsilon}^{d,0}(\fx_{K_{\varepsilon},0},x),\dots,\mlp_{N_{\varepsilon},N_{\varepsilon},\varepsilon}^{d,0}(\fx_{K_{\varepsilon},K_{\varepsilon}},x)}(t) }^{\fq} } \nu_d(\dx t, \dx x)  }^{\nicefrac{1}{\fq}} \\
        & \le 20 
        \bpr{ 6 (T+1) \fa_d \pr{K_{\varepsilon}}^{-\frac12} 
        + 9 \varepsilon \fb_d + 2^p 6 \fc_d  
        \fm_{N_{\varepsilon},N_{\varepsilon}}
        }
        c_d \\
        & \le 180 (T+1) 2^p 
        \bpr{ \fa_d + \fb_d + \fc_d }
        c_d \varepsilon 
        \dott
    \end{split}
\end{equation}
}
\startnewargseq 
\argument{
\cref{eq:assump_on_Gamma_prodapprox}; 
\cref{eq:assump_on_ANNforHatfct_approx};
\cref{ANN_for_MLP2_item2} in \cref{ANN_for_MLP2}; the triangle inequality; 
\cref{eq:mainthm_assump_measure}}{
that for all $d,K,j,n \in \N$, $\varepsilon,\gamma \in (0,1]$ 
it holds that 
\begin{equation}\llabel{eq:diff_interpol_NNfct}
    \begin{split}
        & \bbbbr{ \int_{[0,T] \times \R^d} \E\bbbr{ \babs{ \scrL_{\fx_{K,0},\dots,\fx_{K,K}}^{\mlp_{n,j,\varepsilon}^{d,0}(\fx_{K,0},x),\dots,\mlp_{n,j,\varepsilon}^{d,0}(\fx_{K,K},x)}(t) - \bpr{\functionANN{a}\bpr{\Phi_{\gamma,n,j,K}^{d,0,\varepsilon}}}(t,x) }^{\fq} } \nu_d(\dx t, \dx x)  }^{\nicefrac{1}{\fq}} \\
        & \le 
        2 \gamma \bpr{1+(T+1)^q}^q 
        \bbbbbr{ (K+1) \constantAssumpMainThm d^{rp^2 q}
        + \sum_{k=0}^K 
        \bbbbr{ \int_{[0,T] \times \R^d} \E\bbbr{ \babs{ \mlp_{n,j,\varepsilon}^{d,0}(\fx_{K,k},x) }^{q\fq} } \nu_d(\dx t, \dx x)  }^{\nicefrac{1}{\fq}} } 
        \dott 
    \end{split}
\end{equation}
}
\argument{
Jensen's inequality; the triangle inequality}{
that for all $d,K,j,n \in \N$, $\varepsilon \in (0,1]$, $k \in \{0,1,\dots,K\}$ 
it holds that 
\begin{equation}\llabel{eq:int_mlp_split}
    \begin{split}
        & \bbbbr{ \int_{[0,T] \times \R^d} \E\bbbr{ \babs{ \mlp_{n,j,\varepsilon}^{d,0}(\fx_{K,k},x) }^{q\fq} } \nu_d(\dx t, \dx x)  }^{\nicefrac{1}{\fq}} \\
        & \le 3^{q-1} \Bigg(
        \bbbbr{ \int_{[0,T] \times \R^d} \E\bbbr{ \babs{ \mlp_{n,j,\varepsilon}^{d,0}(\fx_{K,k},x) - v_{d,\varepsilon}(\fx_{K,k},x) }^{q\fq} } \nu_d(\dx t, \dx x)  }^{\nicefrac{1}{\fq}} \\
        & \qquad \qquad + \bbbbr{ \int_{[0,T] \times \R^d}  \abs{ v_{d,\varepsilon}(\fx_{K,k},x) - v_{d,\varepsilon}(T,x) }^{q\fq}\, \nu_d(\dx t, \dx x)  }^{\nicefrac{1}{\fq}}  \\
        & \qquad \qquad + \bbbbr{ \int_{[0,T] \times \R^d} \abs{ v_{d,\varepsilon}(T,x) }^{q\fq}\, \nu_d(\dx t, \dx x)  }^{\nicefrac{1}{\fq}} 
        \Bigg) \dott
    \end{split}
\end{equation}
}
\argument{\lref{eq:soln_PDE_ANNfcts}; 
the fact that for all $\varepsilon \in (0,1]$, $w, z \in \R$ it holds that $\abs{(\Ra(\interpolatingDNN_{0,\varepsilon}))(w) - (\Ra(\interpolatingDNN_{0,\varepsilon}))(z)} \le \LipConstF \abs{w-z}$;
the fact that for all $\varepsilon \in (0,1]$, $d\in\N_0$, $x \in \R^{\max\{d,1\}}$ it holds that $\abs{ (\Ra(\interpolatingDNN_{d,\varepsilon}))(x) } \le \constantAssumpMainThm (\max\{d,1\})^p (1+\norm{x})^p $;
Hutzenthaler et al.~\cite[Corollary 3.15]{PadgettJentzen2021} (applied for every $d\in\N$, $\varepsilon \in (0,1]$, $j\in\N$ with 
$\fp \with q\fq$, $d \with d$, $m\with \littleMM_j$, $T\with T$, $\LipConstF \with \LipConstF$, $\fL \with 2^{p-1} \constantAssumpMainThm d^p$, $p\with p$, 
$u \with v_{d,\varepsilon}$, $f\with ([0,T] \times \R^d \times \R \ni (t,x,z)\mapsto (\Ra(\interpolatingDNN_{0,\varepsilon}))(z) \in\R)$,
$g \with \Ra(\ftermnn_{d,\varepsilon})$, 
$\Theta \with \Theta$, 
$(W^{\theta})_{\theta \in \Theta} \with (W^{d,\theta})_{\theta \in \Theta}$, 
$(U_{n}^{\theta})_{(n,\theta)\in\Z \times \Theta}\with (\mlp_{n,j,\varepsilon}^{d,\theta})_{(n,\theta) \in \Z \times \Theta}$, 
$(\fu^\theta)_{\theta \in \Theta} \with (\fu^\theta)_{\theta \in \Theta}$, 
$(\cU^\theta)_{\theta \in \Theta} \with (\cU^\theta)_{\theta \in \Theta}$ 
in the notation of Hutzenthaler et al.~\cite[Corollary 3.15]{PadgettJentzen2021});
\lref{eq:proofmainthm_constant_c}; 
\lref{eq:proofmainthm_constant_m}; 
Jensen's inequality; the triangle inequality}{
that for all $d,K,j,n \in \N$, $\varepsilon \in (0,1]$, $k \in \{0,1,\dots,K\}$, $x \in \R^d$  
it holds that 
\begin{equation}\llabel{eq:int_mlp_split_diff_mlp_v}
    \begin{split}
        & \E\bbbr{ \babs{ \mlp_{n,j,\varepsilon}^{d,0}(\fx_{K,k},x) - v_{d,\varepsilon}(\fx_{K,k},x) }^{q\fq} } \\
        & \le 
        (\fc_d)^{q\fq} 
        \bbbbr{
        (1+2\LipConstF T)^n  
        (\littleMM_j)^{-\frac{n}{2}} \exp\bbbpr{ \frac{(\littleMM_j)^{\frac{q\fq}{2}}}{q\fq} }
        }^{q\fq}
        \bpr{ \sup\nolimits_{s \in [0,T]}  \E\bbr{( 1 + \norm{ x + W_s^{d,0} }^p )^{q\fq}}  } \\
        & \le 
        (\fc_d)^{q\fq} 
        (\fm_{j,n})^{q\fq}
        2^{(p+1)q\fq } 
        \bpr{ 1 + \norm{x}^{p^2q\fq} + \sup\nolimits_{s \in [0,T]}  \E\bbr{ \norm{ W_s^{d,0} }^{2p^2q\fq} }  } 
        \dott 
    \end{split}
\end{equation}
}
\argument{Jensen's inequality;
the triangle inequality}{
that for all $d,K \in \N$, $\varepsilon \in (0,1]$, $k \in \{0,1,\dots,K\}$
it holds that 
\begin{equation}\llabel{eq:int_mlp_subsplit}
    \begin{split}
         & \bbbbr{ \int_{[0,T] \times \R^d}  \abs{ v_{d,\varepsilon}(\fx_{K,k},x) - v_{d,\varepsilon}(T,x) }^{q\fq}\, \nu_d(\dx t, \dx x)  }^{\nicefrac{1}{\fq}} \\
         & \le 3^{q-1} \Bigg( \bbbbr{ \int_{[0,T] \times \R^d}  \abs{ v_{d,\varepsilon}(\fx_{K,k},x) - u_{d}(\fx_{K,k},x) }^{q\fq} \, \nu_d(\dx t, \dx x)  }^{\nicefrac{1}{\fq}} \\
         & \qquad \qquad + \bbbbr{ \int_{[0,T] \times \R^d}  \abs{ u_{d}(\fx_{K,k},x) - u_d(T,x) }^{q\fq} \, \nu_d(\dx t, \dx x)  }^{\nicefrac{1}{\fq}} \\
         & \qquad \qquad + \bbbbr{ \int_{[0,T] \times \R^d}  \abs{ u_d(T,x) - v_{d,\varepsilon}(T,x) }^{q\fq} \, \nu_d(\dx t, \dx x)  }^{\nicefrac{1}{\fq}} 
         \Bigg)
         \dott 
    \end{split}
\end{equation}
}
\argument{\lref{eq:proofmainthm_constant_BM}; \lref{eq:sup_diff_ud_vdeps}; Jensen's inequality}{
that for all $d, K\in \N$, $\varepsilon \in (0,1]$, $k \in \{0,1,\dots,K\}$, $\ft\in \{\fx_{K,k},T\}$
it holds that 
\begin{equation}\llabel{eq:int_mlp_subsplit_diff_u_v}
    \begin{split}
        & \bbbbr{ \int_{[0,T] \times \R^d}  \abs{ u_d(\ft,x) - v_{d,\varepsilon}(\ft,x) }^{q\fq}\, \nu_d(\dx t, \dx x)  }^{\nicefrac{1}{\fq}} \\
        & \le \varepsilon^q (\fb_d)^{q}
        \bbbbr{ \int_{[0,T] \times \R^d}  \bpr{ 1 + \norm{x}^{p^2} + \sup\nolimits_{s\in[0,T]} \E\bbr{ \norm{ W_s^{d,0} }^{p^2} } }^{q\fq}\, \nu_d(\dx t, \dx x)  }^{\nicefrac{1}{\fq}} \\
        & \le 3^q \varepsilon^q (\fb_d)^{q} 
         \bbbbr{ \int_{[0,T] \times \R^d}  \bpr{ 1 + \norm{x}^{p^2q\fq} + \sup\nolimits_{s\in[0,T]} \E\bbr{ \norm{ W_s^{d,0} }^{2 p^2 q\fq} } }\, \nu_d(\dx t, \dx x)  }^{\nicefrac{1}{\fq}} \\
        & =  3^q \varepsilon^q (\fb_d)^{q}  c_d 
         \dott 
    \end{split}
\end{equation}
}
\argument{\lref{eq:fd_bound}; \lref{eq:1249}; 
the fact that for all $w,z \in \R$ it holds that $\abs{f_0(z) - f_0(w)}\le \LipConstF \abs{z-w}$; 
the fact that for all $d\in\N$, $x \in \R^d$ it holds that $\norm{\nabla f_d(x)}\le\constantAssumpMainThm d^p (1+\norm{x})^{p}$; 
\cref{temp_u_reg1} (applied for every $d \in \N$ 
with  $u\with u_d$, $T\with T$, $d\with d$, $W \with W^{d,0}$, $g \with f_d$, $f\with ([0,T]\times \R^d\times \R \ni (t,x,z) \mapsto f_0(z) \in\R)$, $\LipConstF\with \LipConstF$, $\boundFG \with \constantAssumpMainThm d^p$, $p \with p$ 
in the notation of \cref{temp_u_reg1}); 
\lref{eq:proofmainthm_constant_a}; 
Jensen's inequality; 
\lref{eq:proofmainthm_constant_BM}}{
that for all $d,K \in \N$, $k \in \{0,1,\dots,K\}$ 
it holds that 
\begin{equation}\llabel{eq:int_mlp_subsplit_diff_u_time}
    \begin{split}
        & \bbbbr{ \int_{[0,T] \times \R^d}  \abs{ u_{d}(\fx_{K,k},x) - u_d(T,x) }^{q\fq}\, \nu_d(\dx t, \dx x)  }^{\nicefrac{1}{\fq}} \\
        & \le       
        \bpr{T-\fx_{K,k}}^{\frac{q}{2}} d^{\frac{q}{2}} 
        \bpr{e^{2 \LipConstF T} (T+1)^{2} (\LipConstF +1) (\constantAssumpMainThm d^p +1) 8^{p+2} }^q
        \\
        & \quad \cdot 
        \bbbbr{ \int_{[0,T] \times \R^d}          
        \bpr{ 1 + \norm{x}^{p} + \sup\nolimits_{s \in [0,T]} \E\bbr{\norm{W_s^{d,0}}^{2 p }} }^{q\fq} \,
        \nu_d(\dx t, \dx x)  }^{\nicefrac{1}{\fq}} \\
        & \le
        T^{\frac{q}{2}}  
        (\fa_d)^{q}  3^{q}
        \bbbbr{ \int_{[0,T] \times \R^d}          
        \bpr{ 1 + \norm{x}^{p^2 q \fq} + \sup\nolimits_{s \in [0,T]} \E\bbr{\norm{W_s^{d,0}}^{2 p^2 q \fq}} } \, 
        \nu_d(\dx t, \dx x)  }^{\nicefrac{1}{\fq}} \\
        & = T^{\frac{q}{2}}  
        (\fa_d)^{q}  3^{q} c_d 
        \dott 
    \end{split}
\end{equation}
}
\argument{\lref{eq:int_mlp_subsplit}; \lref{eq:int_mlp_subsplit_diff_u_v}; \lref{eq:int_mlp_subsplit_diff_u_time}}{
that for all $d,K \in \N$, $\varepsilon \in (0,1]$, $k \in \{0,1,\dots,K\}$
it holds that 
\begin{equation}\llabel{eq:int_mlp_subsplit_estimate}
    \begin{split}
        & \bbbbr{ \int_{[0,T] \times \R^d}  \abs{ v_{d,\varepsilon}(\fx_{K,k},x) - v_{d,\varepsilon}(T,x) }^{q\fq} \, \nu_d(\dx t, \dx x)  }^{\nicefrac{1}{\fq}} \\
         & \le 3^{q-1} \bpr{ 3^{q} 2 \varepsilon^q (\fb_d)^q c_d + T^{\frac{q}{2}} (\fa_d)^q 3^q c_d } 
         \le 3^{2q} (T+1)^{\frac{q}{2}} \bpr{(\fa_d)^q + (\fb_d)^q} c_d 
         \dott 
    \end{split}
\end{equation}
}
\argument{\lref{eq:soln_PDE_ANNfcts};
the fact that for all $d\in\N$, $x \in \R^d$, $\varepsilon \in (0,1]$ it holds that $\abs{\bpr{\Ra(\ftermnn_{d,\varepsilon})}(x)} \le \constantAssumpMainThm d^p (1+\norm{x})^{p}$;
Jensen's inequality; 
\lref{eq:proofmainthm_constant_BM}
}{
that for all $d \in \N$, $\varepsilon \in (0,1]$ 
it holds that 
\begin{equation}\llabel{eq:int_mlp_split_v}
    \begin{split}
        \bbbbr{ \int_{[0,T] \times \R^d}  \abs{ v_{d,\varepsilon}(T,x) }^{q\fq} \, \nu_d(\dx t, \dx x)  }^{\nicefrac{1}{\fq}} 
        & = \bbbbr{ \int_{[0,T] \times \R^d} \babs{ \bpr{\Ra(\ftermnn_{d,\varepsilon})}(x) }^{q\fq} \, \nu_d(\dx t, \dx x)  }^{\nicefrac{1}{\fq}} \\
        & \le \constantAssumpMainThm^q d^{pq} \bbbbr{ \int_{[0,T] \times \R^d} (1+\norm{x})^{pq\fq} \, \nu_d(\dx t, \dx x)  }^{\nicefrac{1}{\fq}} \\
        & \le \constantAssumpMainThm^q d^{pq} 
        2^{\frac{p^2q\fq -1}{\fq}}
        \bbbbr{ \int_{[0,T] \times \R^d} \bpr{1+\norm{x}^{p^2q\fq}} \, \nu_d(\dx t, \dx x)  }^{\nicefrac{1}{\fq}} \\
        & \le \constantAssumpMainThm^q d^{pq} 
        2^{p^2 q} c_d 
        \dott
    \end{split}
\end{equation}
}
\argument{\lref{eq:proofmainthm_constant_BM}; \lref{eq:int_mlp_split}; \lref{eq:int_mlp_split_diff_mlp_v}; \lref{eq:int_mlp_subsplit_estimate}; \lref{eq:int_mlp_split_v}}{
that for all $d,K,j,n \in \N$, $\varepsilon \in (0,1]$, $k \in \{0,1,\dots,K\}$ 
it holds that 
\begin{equation}\llabel{eq:2039}
    \begin{split}
        & \bbbbr{ \int_{[0,T] \times \R^d} \E\bbbr{ \babs{ \mlp_{n,j,\varepsilon}^{d,0}(\fx_{K,k},x) }^{q\fq} } \nu_d(\dx t, \dx x)  }^{\nicefrac{1}{\fq}} \\
        & \le 3^{q-1} c_d \bpr{
        (\fc_d)^{q} 
        (\fm_{j,n})^{q} 
        2^{(p+1)q} 
        + 3^{2q} (T+1)^{\frac{q}{2}} \pr{(\fa_d)^q + (\fb_d)^q }  
        + \constantAssumpMainThm^q d^{pq} 
        2^{p^2 q}  
        } \\
        & \le 3^{3p^2 q}
        c_d 
        \bpr{
        (\fc_d)^{q} 
        (\fm_{j,n})^q 
        + (T+1)^{\frac{q}{2}} ((\fa_d)^q + (\fb_d)^q)  
        + \constantAssumpMainThm^q d^{pq} 
        } 
        \dott
    \end{split}
\end{equation}
}
\argument{\lref{eq:2039}; \lref{eq:diff_interpol_NNfct}}{
that for all $d,K,j,n \in \N$, $\varepsilon,\gamma \in (0,1]$ 
it holds that 
\begin{equation}\llabel{eq:diff_interpol_NNfct_estimate}
    \begin{split}
        & \bbbbr{ \int_{[0,T] \times \R^d} \E\bbbr{ \babs{ \scrL_{\fx_{K,0},\dots,\fx_{K,K}}^{\mlp_{n,j,\varepsilon}^{d,0}(\fx_{K,0},x),\dots,\mlp_{n,j,\varepsilon}^{d,0}(\fx_{K,K},x)}(t) - \bpr{\functionANN{a}\bpr{\Phi_{\gamma,n,j,K}^{d,0,\varepsilon}}}(t,x) }^{\fq} } \nu_d(\dx t, \dx x)  }^{\nicefrac{1}{\fq}} \\
        & \le 
        2 \gamma \bpr{1+(T+1)^q}^q 
        (K+1) 
        \constantAssumpMainThm d^{rp^2 q} \\
        & \quad \cdot 
        \bbr{
        1 + 3^{3p^2 q}
        c_d 
        \bpr{
        (\fc_d)^{q} 
        (\fm_{j,n})^q
        + (T+1)^{\frac{q}{2}} ((\fa_d)^q + (\fb_d)^q)  
        + \constantAssumpMainThm^q d^{pq} 
        } 
        }
        \dott 
    \end{split}
\end{equation}
}
\argument{\lref{eq:diff_interpol_NNfct_estimate}; \lref{eq:proofmainthm_constant_m}; \lref{eq:proofmainthm_N}; 
\cref{lemma_littleMM_item1} in \cref{lemma_littleMM}; 
\lref{eq:proofmainthm_K}; \lref{eq:proofmainthm_gamma}}{
that for all $d\in \N$, $\varepsilon  \in (0,1]$ it holds that 
\begin{equation}\llabel{eq:0001c}
    \begin{split}
        & \bbbbr{ \int_{[0,T] \times \R^d} \E\bbbr{ \babs{ \scrL_{\fx_{K_{\varepsilon},0},\dots,\fx_{K_{\varepsilon},K_{\varepsilon}}}^{\mlp_{N_{\varepsilon},N_{\varepsilon},\varepsilon}^{d,0}(\fx_{K_{\varepsilon},0},x),
        \dots,\mlp_{N_{\varepsilon},N_{\varepsilon},\varepsilon}^{d,0}(\fx_{K_{\varepsilon},K_{\varepsilon}},x)}(t) - \bpr{\functionANN{a}\bpr{\Phi_{\gamma_{d,\varepsilon},N_{\varepsilon},N_{\varepsilon},K_{\varepsilon}}^{d,0,\varepsilon}}}(t,x) }^{\fq} } \nu_d(\dx t, \dx x)  }^{\nicefrac{1}{\fq}} \\
        & \le 
        2 
        \gamma_{d,\varepsilon} \bpr{K_{\varepsilon}+1} 
        \constantAssumpMainThm d^{rp^2 q} 
        \bpr{1+(T+1)^q}^q \\
        & \quad \cdot 
        \bbpr{
        1 + 3^{3p^2 q}
        c_d 
        \bpr{
        (\fc_d)^{q} 
        \pr{\fm_{N_{\varepsilon},N_{\varepsilon}}}^{q}
        + (T+1)^{\frac{q}{2}} ((\fa_d)^q + (\fb_d)^q)  
        + \constantAssumpMainThm^q d^{pq} 
        } 
        } \\
        & \le 
        2 
        \gamma_{d,\varepsilon} \bpr{K_{\varepsilon}+1} 
        \constantAssumpMainThm 
        d^{rp^2 q} 
        \bpr{1+(T+1)^q}^q 
        \bbpr{
        1 + 3^{3p^2 q}
        c_d 
        \bpr{
        (\fc_d)^{q} 
        + (T+1)^{\frac{q}{2}} ((\fa_d)^q + (\fb_d)^q)  
        + \constantAssumpMainThm^q d^{pq} 
        } 
        } \\
        & = \varepsilon
        \dott 
    \end{split}
\end{equation}
}
\argument{\lref{eq:proofmainthm_delta}; \lref{eq:0001a}; \lref{eq:0001b}; \lref{eq:0001c}}{
that for all 
$d \in \N$, $\varepsilon \in (0,1]$
it holds that 
\begin{equation}\llabel{eq:2042}
    \begin{split}
        & \bbbbr{ \int_{[0,T] \times \R^d} \E\bbbr{ \babs{ u_d(t,x) - \bpr{\functionANN{a}\bpr{\Phi_{\gamma_{d,\delta_{d,\varepsilon}},N_{\delta_{d,\varepsilon}},N_{\delta_{d,\varepsilon}},K_{\delta_{d,\varepsilon}}}^{d,0,\delta_{d,\varepsilon}}}}(t,x) }^{\fq} } \nu_d(\dx t, \dx x)  }^{\nicefrac{1}{\fq}} \\
        & \le 
        \bpr{ 180 (T+1) 2^p 
        \bpr{ \fa_d +\fb_d + \fc_d }
        c_d + 1 } 
        \delta_{d,\varepsilon}
        = \varepsilon 
        \dott
    \end{split}
\end{equation}
}
\argument{\lref{eq:2042}; Fubini's theorem}{
that for all 
$d \in \N$, $\varepsilon \in (0,1]$
it holds that 
\begin{equation}\llabel{eq:2043}
    \begin{split}
        & \E\bbbbr{ \int_{[0,T] \times \R^d} \babs{ u_d(t,x) - \bpr{\functionANN{a}\bpr{\Phi_{\gamma_{d,\delta_{d,\varepsilon}},N_{\delta_{d,\varepsilon}},N_{\delta_{d,\varepsilon}},K_{\delta_{d,\varepsilon}}}^{d,0,\delta_{d,\varepsilon}}}}(t,x) }^{\fq} \, \nu_d(\dx t, \dx x)  } 
        \le \varepsilon^{\fq}  
        \dott
    \end{split}
\end{equation}
}
\argument{\lref{eq:2043}}{
that for all $d \in \N$, $\varepsilon \in (0,1]$ there exists $\omega_{d,\varepsilon} \in \Omega$ such that 
\begin{equation}\llabel{eq:1158}
    \begin{split}
        \int_{[0,T] \times \R^d} \babs{ u_d(t,x) - \bpr{\functionANN{a}\bpr{\Phi_{\gamma_{d,\delta_{d,\varepsilon}},N_{\delta_{d,\varepsilon}},N_{\delta_{d,\varepsilon}},K_{\delta_{d,\varepsilon}}}^{d,0,\delta_{d,\varepsilon}} (\omega_{d,\varepsilon}) }}(t,x) }^{\fq} \, \nu_d(\dx t, \dx x) 
        & \le \varepsilon^{\fq} 
        \dott  
    \end{split}
\end{equation}
}
\startnewargseq 
\argument{\cref{ANN_for_MLP2_item3} in \cref{ANN_for_MLP2}}{
that for all $d \in \N$, $\varepsilon \in (0,1]$ it holds that 
\begin{equation}\llabel{eq:1059}
    \begin{split}
    & \paramANN\bbpr{
    \Phi_{\gamma_{d,\delta_{d,\varepsilon}},N_{\delta_{d,\varepsilon}},N_{\delta_{d,\varepsilon}},K_{\delta_{d,\varepsilon}}}^{d,0,\delta_{d,\varepsilon}} (\omega_{d,\varepsilon})
    } \\
    & \le 16 \bpr{\max\{\fd,\lengthANN(\ftermnn_{d,\delta_{d,\varepsilon}})\} + \lengthANN(\interpolatingDNN_{0,\delta_{d,\varepsilon}})} 
    \bbbr{ \max\bcu{\fd,\normmm{\dims(\interpolatingDNN_{0,\delta_{d,\varepsilon}})},\normmm{\dims(\ftermnn_{d,\delta_{d,\varepsilon}})}} }^2
   \\
    & \quad \cdot
     \bbbr{ \bpr{N_{\delta_{d,\varepsilon}}}^{\nicefrac{1}{2}} \bpr{3 \littleMM_{N_{\delta_{d,\varepsilon}}}}^{N_{\delta_{d,\varepsilon}}}}^2 
     \bbbr{\paramANN\bpr{\Gamma_{\gamma_{d,\delta_{d,\varepsilon}}}}}^3 
     \bbbbr{ \max_{k\in\{0,1,\dots,K_{\delta_{d,\varepsilon}}\}} \paramANN\bbpr{ \ANNhatfct_{K_{\delta_{d,\varepsilon}},k,\gamma_{d,\delta_{d,\varepsilon}}} }}^3 \\
     & \quad \cdot 
    (K_{\delta_{d,\varepsilon}}+1)^2 \fd^2 
    \dott 
    \end{split}
\end{equation}
}
\argument{the fact that for all $d \in \N$, $\varepsilon \in(0,1]$ it holds that $\lengthANN(\interpolatingDNN_{0,\varepsilon}) \le  \constantAssumpMainThm \varepsilon^{-\alpha_0}$, $\normmm{\dims(\interpolatingDNN_{0,\varepsilon})} \le \constantAssumpMainThm \varepsilon^{-\beta_0}$, $\lengthANN(\ftermnn_{d,\varepsilon}) \le \constantAssumpMainThm d^p \varepsilon^{-\alpha_1}$, and $\normmm{\dims(\ftermnn_{d,\varepsilon})} \le \constantAssumpMainThm d^p \varepsilon^{-\beta_1}$}{
that for all $d \in \N$, $\varepsilon \in (0,1]$ it holds that 
\begin{equation}\llabel{eq:1047}
    \begin{split}
        & \bpr{\max\{\fd,\lengthANN(\ftermnn_{d,\delta_{d,\varepsilon}})\} + \lengthANN(\interpolatingDNN_{0,\delta_{d,\varepsilon}})} 
        \bbbr{ \max\bcu{\fd,\normmm{\dims(\interpolatingDNN_{0,\delta_{d,\varepsilon}})},\normmm{\dims(\ftermnn_{d,\delta_{d,\varepsilon}})}} }^2 \\
        & \le \bpr{ \constantAssumpMainThm d^p \pr{\delta_{d,\varepsilon}}^{-\alpha_1} + \constantAssumpMainThm \pr{\delta_{d,\varepsilon}}^{-\alpha_0} }
        \bpr{\max\bcu{ \constantAssumpMainThm, \constantAssumpMainThm \pr{\delta_{d,\varepsilon}}^{-\beta_0}, \constantAssumpMainThm d^p \pr{\delta_{d,\varepsilon}}^{-\beta_1} } }^2 \\
        & \le 2 \constantAssumpMainThm^3 d^{3p} \max\bcu{ \pr{\delta_{d,\varepsilon}}^{-\alpha_1}, \pr{\delta_{d,\varepsilon}}^{-\alpha_0}}
        \bpr{ \max\bcu{ \pr{\delta_{d,\varepsilon}}^{-\beta_0}, \pr{\delta_{d,\varepsilon}}^{-\beta_1}} }^2 \\
        & \le 2 \constantAssumpMainThm^3 d^{3p} \pr{\delta_{d,\varepsilon}}^{-(\max\{\alpha_0,\alpha_1\}+2\max\{\beta_0,\beta_1\})}
        \dott 
    \end{split}
\end{equation}
} 
\argument{\cref{lemma_littleMM_item2a} in \cref{lemma_littleMM}}{
that there exists $\bar{\constantMainThmParam}\in (0,\infty)$ such that for all $d\in\N$, $\varepsilon \in (0,1]$ it holds that 
\begin{equation}\llabel{eq:1807}
    \bbbr{ \bpr{N_{\delta_{d,\varepsilon}}}^{\nicefrac{1}{2}} \bpr{3 \littleMM_{N_{\delta_{d,\varepsilon}}}}^{N_{\delta_{d,\varepsilon}}}}^2 
    \le 
    \pr{\delta_{d,\varepsilon}}^{-2(2+\delta)} \bar{\constantMainThmParam}
    \dott 
\end{equation}
}
\argument{\lref{eq:proofmainthm_constant_a}; \lref{eq:proofmainthm_constant_b}; \lref{eq:proofmainthm_constant_c}; 
\cref{lem:BM_power_bound}}{
that there exists $\fC \in [1,\infty)$ such that for all $d\in\N$ it holds that 
$\fa_d \le \fC d^{3p+1}$, 
$\fb_d \le \fC d^{(p+1)p}$,
$\fc_d \le \fC d^{p}$,
and $c_d \le \fC d^{(r+2) p^2 q}$\dott\,
}
Combining \lref{eq:proofmainthm_delta} and \lref{eq:proofmainthm_gamma} 
\hence 
shows  
that there exists $\bar{\fC} \in [1,\infty)$ such that for all $d\in\N$, $\varepsilon \in (0,1]$ it holds that 
\begin{equation}\llabel{eq:1051}
    \begin{split}
        \pr{\gamma_{d,\delta_{d,\varepsilon}}}^{-1} 
        & \le  \bar{\fC} d^{2(r+1) p^2 q + \max\{(p+1)p,3p+1\} q }
        \pr{\delta_{d,\varepsilon}}^{-1} 
        (K_{\delta_{d,\varepsilon}}+1) 
    \end{split}
\end{equation}
and 
\begin{equation}\llabel{eq:1042}
    \begin{split}
        \pr{\delta_{d,\varepsilon}}^{-1}
        & \le \bar{\fC} \varepsilon^{-1} 
        d^{(r+2) p^2 q + \max\{(p+1)p,3p+1\} }
        \dott
    \end{split}
\end{equation}
\argument{
the fact that for all $\gamma \in (0,1]$ it holds that 
$\paramANN(\Gamma_{\gamma})\le \constantAssumpMainThm \gamma^{-r}$; 
the fact that for all $K\in\N$, $k\in\{0,1,\dots,K\}$, $\gamma \in (0,1]$ it holds that $\paramANN(\ANNhatfct_{K,k,\gamma})\le \constantAssumpMainThm K^r \gamma^{-r}$;  \lref{eq:1051}; the fact that for all $\varepsilon \in (0,1]$ it holds that $K_{\varepsilon} +1 = (K_{\varepsilon}-1) +2 \le \varepsilon^{-2} +2 \le 3\varepsilon^{-2}$}{
that for all $d\in\N$, $\varepsilon \in (0,1]$ it holds that 
\begin{equation}\llabel{eq:1032}
    \begin{split}
        & \bbbr{\paramANN\bpr{\Gamma_{\gamma_{d,\delta_{d,\varepsilon}}}}}^3 
     \bbbbr{ \max_{k\in\{0,1,\dots,K_{\delta_{d,\varepsilon}}\}} \paramANN\bbpr{ \ANNhatfct_{K_{\delta_{d,\varepsilon}},k,\gamma_{d,\delta_{d,\varepsilon}}} }}^3 
        (K_{\delta_{d,\varepsilon}}+1)^2 \\
        & \le 
        \constantAssumpMainThm^6 
        \pr{\gamma_{d,\delta_{d,\varepsilon}}}^{-6r} 
        \bpr{K_{\delta_{d,\varepsilon}}}^{3r}
        (K_{\delta_{d,\varepsilon}}+1)^2 \\
        & \le 
         \constantAssumpMainThm^6 
        \bar{\fC}^{6r} 
        d^{6r\pr{2(r+1) p^2 q + \max\{(p+1)p,3p+1\} q } }
        \pr{\delta_{d,\varepsilon}}^{-6r} 
        (K_{\delta_{d,\varepsilon}}+1)^{9r+2} \\
        & \le 
        3^{9r+2} 
         \constantAssumpMainThm^6 
        \bar{\fC}^{6r} 
        d^{6r q\pr{2(r+1) p^2 + (p+1)^2 } }
        \pr{\delta_{d,\varepsilon}}^{-6r-2(9r+2)} 
        \dott 
    \end{split}
\end{equation}
}
\argument{\lref{eq:1059}; \lref{eq:1047}; \lref{eq:1807}; 
\lref{eq:1032};
the fact that $\constantMainThmDelta=\max\{\alpha_0,\alpha_1\} + 2 \max\{\beta_0,\beta_1\} + 2\delta + 24r + 8$}{ 
that for all $d\in\N$, $\varepsilon \in (0,1]$ it holds that 
\begin{equation}\llabel{eq:2047}
    \begin{split}
    \paramANN\bbpr{
    \Phi_{\gamma_{d,\delta_{d,\varepsilon}},N_{\delta_{d,\varepsilon}},N_{\delta_{d,\varepsilon}},K_{\delta_{d,\varepsilon}}}^{d,0,\delta_{d,\varepsilon}} (\omega_{d,\varepsilon})
    } 
    & \le 32
    \constantAssumpMainThm^3 d^{3p} \pr{\delta_{d,\varepsilon}}^{-(\max\{\alpha_0,\alpha_1\}+2\max\{\beta_0,\beta_1\})}
    \pr{\delta_{d,\varepsilon}}^{-2(2+\delta)} \bar{\constantMainThmParam}   
   \\
    & \quad \cdot
    3^{9r+2}  \constantAssumpMainThm^6  \bar{\fC}^{6r} 
    d^{6r q\pr{2(r+1) p^2 + (p+1)^2 } }
    \pr{\delta_{d,\varepsilon}}^{-6r-2(9r+2)} 
    \fd^2 \\
    & \le  
    32 \constantAssumpMainThm^9  \bar{\fC}^{6r} \fd^2 3^{9r+2} 
    \bar{\constantMainThmParam}
    d^{6r q\pr{2(r+1) p^2 + (p+1)^2 } +3p }
    \pr{\delta_{d,\varepsilon}}^{-\constantMainThmDelta}
    \dott 
    \end{split}
\end{equation}
}
\argument{\lref{eq:2047}; \lref{eq:1042}}{
that there exists $\constantMainThmParam \in (0,\infty)$
such that for all $d\in\N$, $\varepsilon \in (0,1]$ it holds that
\begin{equation}
    \begin{split}
    \paramANN\bbpr{
    \Phi_{\gamma_{d,\delta_{d,\varepsilon}},N_{\delta_{d,\varepsilon}},N_{\delta_{d,\varepsilon}},K_{\delta_{d,\varepsilon}}}^{d,0,\delta_{d,\varepsilon}} (\omega_{d,\varepsilon})
    } 
    & \le 
    \constantMainThmParam  
    d^{6r q\pr{2(r+1) p^2 + (p+1)^2 } +3p }
    d^{\constantMainThmDelta \pr{(r+2) p^2 q + \max\{(p+1)p,3p+1\}} }
    \varepsilon^{-\constantMainThmDelta} \\
    & \le 
    \constantMainThmParam 
    d^{
    (2(r+1)p^2 + (p+1)^2 )(6r+ \constantMainThmDelta)q + 3p
    } 
    \varepsilon^{-\constantMainThmDelta}
    \dott 
    \end{split}
\end{equation}
}
This and \lref{eq:1158} prove \cref{mainThm_item2}. 
\end{aproof}

\subsection{ANN approximations for PDEs with specific activation functions}
\label{subsec:ANN_approx_PDE_specific_activation}

\cfclear
\begin{athm}{corollary}{cor_of_mainthm1}
Let 
$\constantAssumpMainThm, T, \fc \in (0,\infty)$, 
$r \in[1,\infty)$, 
$p \in (1,\infty)$, 
$\fq \in [2,\infty)$, 
$q \in (2,\infty)$, 
$\leaky \in \R\backslash\{-1,1\}$, 
$\indexAct \in \{0,1\}$, 
let
$f\colon \R \to \R$ be Lipschitz continuous, 
for every $d \in \N$ let $u_d \in C^{1,2}([0,T]\times \R^d,\R)$
satisfy for all $t\in[0,T]$, $x \in \R^d$ that 
\begin{equation}\label{eq:cor1:heateq}
\tfrac{\partial}{\partial t} u_d(t,x)
+
\fc \Delta_x u_d(t,x)
+
f\pr{ u_d(t,x) }
=
0, 
\end{equation}
for every 
$d \in \N$ 
let 
$\mu_d \colon \mathcal{B}(\R^{d}) \to [0,\infty)$ 
be a measure with 
\begin{equation}\label{eq:assump_measure_cor1}
\textstyle{\int_{\R^{d}} \pr{1+\norm{x}^{p^2q\fq}} \,\mu_{d}(\dxx x) \le \constantAssumpMainThm d^{rp^2q\fq}}, 
\end{equation} 
let 
$a \colon \R\to\R$
satisfy for all $x \in \R$ that $a(x) = \indexAct \max\{x,\leaky x\} + (1-\indexAct)  \ln(1+\exp(x))$, 
and assume for all $d \in \N$, $\varepsilon \in (0,1]$ that there exists 
$\G \in \ANNsm$ such that for all $t \in [0,T]$, $x \in \R^d$ it holds that 
\begin{equation}
    \functionANN{a}(\G) \in C(\R^d,\R) , \qquad \paramANN(\G) \le \constantAssumpMainThm d^p \varepsilon^{-\constantAssumpMainThm}, \qquad \text{and}
\end{equation}
\begin{equation}\label{eq:assump_approx_cor1}
    \varepsilon \norm{\nabla_x u_d(T,x)} 
    + \varepsilon \abs{u_d(t,x)} 
    + \abs{ u_d(T,x) - (\functionANN{a}(\G))(x) }
    \le \varepsilon \constantAssumpMainThm d^p \pr{1+\norm{x}}^p
\end{equation}
\cfload.
Then 
there exists 
$c \in \R$ 
such that for all $d \in \N$, $\varepsilon \in (0,1]$ 
there exists $\U \in \ANNsm$ 
such that 
\begin{equation}
    \functionANN{a}(\U) \in C(\R^{d+1},\R), \qquad \paramANN(\U) \le c d^c \varepsilon^{-c}, \qquad \text{and} 
\end{equation}
\begin{equation}
    \bbbbr{ \int_0^T \int_{\R^d} \abs{u_d(t,x)-\pr{\functionANN{a}(\U)}(t,x)}^{\fq} \, \mu_d(\dxx x) \, \dxx t  }^{\nicefrac{1}{\fq}}
    \le \varepsilon \dott 
\end{equation}
\end{athm}

\cfclear
\begin{aproof}
    Throughout this proof let $\scrf\colon \R \to \R$ satisfy for all $w\in\R$ that 
    \begin{equation}
        \scrf(w)=(2\fc)^{-1} f(w), 
    \end{equation}
    for every $d\in\N$ let $\scru_d\colon [0,2\fc T]\times \R^d \to \R$ satisfy for all $t \in [0,2\fc T]$, $x \in \R^d$ that 
    \begin{equation}\llabel{eq:defscru1}
        \scru_d(t,x) = u_d((2\fc)^{-1}t,x),
    \end{equation}
    let $\LipConstF \in [1,\infty)$ satisfy for all $v, w\in\R$ that 
    \begin{equation}
        \abs{f(v)-f(w)}\le \LipConstF \abs{v-w}, 
    \end{equation}
    for every $d \in \N$, $\varepsilon \in (0,1]$ let $\F_{d,\varepsilon} \in \ANNsm$ satisfy for all 
    $t \in [0,T]$, $x \in \R^d$ that $\functionANN{a}(\F_{d,\varepsilon}) \in C(\R^d,\R)$, $\paramANN(\F_{d,\varepsilon}) \le \constantAssumpMainThm d^p \varepsilon^{-\constantAssumpMainThm}$, and 
    \begin{equation}\llabel{eq:2104}
        \varepsilon \norm{\nabla_x u_d(T,x)} 
        + \varepsilon \abs{u_d(t,x)}  
        + \abs{ u_d(T,x) - (\functionANN{a}(\F_{d,\varepsilon}))(x) }
        \le \varepsilon \constantAssumpMainThm d^p \pr{1+\norm{x}}^p,
    \end{equation}
    for every $d\in\N$ let 
    $\nu_d\colon \cB(\R^{d+1}) \to \R$ 
    satisfy for all $B \in \cB(\R^{d+1})$ that 
    \begin{equation}\llabel{eq:2207}
        \nu_d(B) = \frac{1}{2\fc} \int_0^{2\fc T} \int_{\R^d} \1_B(t,x) \, \mu_d(\dxx x) \, \dxx t,
    \end{equation} 
    let
    \begin{equation}
        \tilde{r} 
        = \max\bbbcu{ r, \frac{q^2}{(q-2)(q-1)}, \frac{q}{q-1} }, 
    \end{equation}
    and let 
    \begin{equation}
        \tilde{\constantAssumpMainThm}
        = \max\bbbcu{ 
        2^{\frac{p^2q\fq}{2}} \constantAssumpMainThm T (2\fc T+1)^{p^2 q\fq}, 
        2\constantAssumpMainThm + 4 + (2\fc)^{-1} \pr{\LipConstF + \abs{f(0)} },
        1728 \cdot 2^{\tfrac{q^3+3q^2-2q}{(q-2)(q-1)}},
        19 \cdot \bbpr{2+\frac{8}{T}}^{\frac{q}{q-1}} 
        }
    \end{equation}
    \cfload.
    \startnewargseq
    \argument{the fact that for all $d\in\N$ it holds that $u_d\in C^{1,2}([0,T]\times \R^d,\R)$;
    the fact that for all $d \in \N$, $t\in[0,T]$, $x\in\R^d$
    it holds that $\abs{u_d(t,x)} \le \constantAssumpMainThm d^p (1+\norm{x})^p$;
    \cref{eq:cor1:heateq}}{
    that for all $d \in \N$, $t\in[0,2\fc T]$, $x\in\R^d$
    it holds that     
    $\scru_d\in C^{1,2}([0,2\fc T]\times \R^d,\R)$, that 
    $\scru_d$ is at most polynomially growing, 
    and that 
    \begin{equation}
        \tfrac{\partial}{\partial t} \scru_d(t,x)
        + \tfrac12 \Delta_x \scru_d(t,x)
        + \scrf\pr{ \scru_d(t,x) }
        = 0\dott
    \end{equation}
    }
    \argument{the triangle inequality; \lref{eq:2104}}{
    that for all $d\in\N$, $x\in\R^d$, $\varepsilon \in (0,1]$ 
    it holds that 
    \begin{equation}\llabel{eq:2240}
        \begin{split}
            & \varepsilon \norm{\nabla_x \scru_d(2\fc T,x)} 
            + \varepsilon \abs{ (\functionANN{a}(\interpolatingDNN_{d,\varepsilon}))(x) } 
            + \abs{ \scru_d(2\fc T,x)- (\functionANN{a}(\interpolatingDNN_{d,\varepsilon}))(x) } \\
            & \le 
            \varepsilon \norm{\nabla_x u_d(T,x)} 
            + \varepsilon \bpr{ \abs{u_d(T,x)} 
            + \abs{ (\functionANN{a}(\interpolatingDNN_{d,\varepsilon}))(x) - u_d(T,x) } }
            + \abs{ u_d(T,x)- (\functionANN{a}(\interpolatingDNN_{d,\varepsilon}))(x) } \\
            & \le 
            \varepsilon \norm{\nabla_x u_d(T,x)} 
            + \varepsilon 
            \abs{u_d(T,x)} 
            + 2 \abs{ u_d(T,x)- (\functionANN{a}(\interpolatingDNN_{d,\varepsilon}))(x) }
            \\
            &\le 2 \varepsilon \constantAssumpMainThm d^p (1 + \norm{x} )^{p}
            \dott 
        \end{split}
    \end{equation}
    }
    \argument{the fact that for all $d\in\N$, $\varepsilon \in (0,1]$ it holds that $\paramANN(\F_{d,\varepsilon})\le \constantAssumpMainThm d^p \varepsilon^{-\constantAssumpMainThm}$; \eg \cite[Lemma~2.4]{ackermann2023deep}}{
    that for all $d\in\N$, $\varepsilon \in (0,1]$ it holds that 
    \begin{equation}
        \varepsilon^{\constantAssumpMainThm} \lengthANN(\F_{d,\varepsilon}) + \varepsilon^{\constantAssumpMainThm} \normmm{\dims(\F_{d,\varepsilon})}
        \le \varepsilon^{\constantAssumpMainThm} \constantAssumpMainThm d^p \varepsilon^{-\constantAssumpMainThm} = \constantAssumpMainThm d^p
        \dott
    \end{equation}
    }
    \argument{\lref{eq:2207}; 
    the fact that for all $d \in \N$ it holds that $\mu_d$ is a finite measure; 
    Jensen's inequality; Fubini's theorem; \cref{eq:assump_measure_cor1}}{
    that for all $d\in\N$ it holds that $\nu_d$ is a finite measure on $(\R^{d+1},\cB(\R^{d+1}))$ 
    with 
    \begin{equation}
        \begin{split}
        & \int_{\R^{d+1}} \pr{1+ \norm{y}^{p^2q\fq} } \, \nu_d(\dxx y) \\
        & \le  2^{\frac{p^2q\fq}{2}-1} \bbbpr{ \frac{1}{2\fc} \int_0^{2\fc T} \int_{\R^d}  \abs{t}^{p^2q\fq}  
        \, \mu_d(\dxx x) \, \dxx t 
        + \frac{1}{2\fc} \int_0^{2 \fc T} \int_{\R^d} \pr{1+\norm{x}^{p^2q\fq} } 
        \, \mu_d(\dxx x) \, \dxx t 
        } \\
        & \le
        2^{\frac{p^2q\fq}{2}-1} \bbbpr{ \frac{\constantAssumpMainThm d^{r p^2 q \fq}}{2\fc} \int_0^{2\fc T} t^{p^2q\fq} \, \dxx t 
        + T \constantAssumpMainThm d^{rp^2q\fq}
        } \\
        & \le 
         2^{\frac{p^2q\fq}{2}} \constantAssumpMainThm T (2\fc T+1)^{p^2 q\fq} d^{rp^2q\fq} 
         \dott 
        \end{split}
    \end{equation}
    }
    \argument{the fact that for all $v, w\in\R$ it holds that $\abs{f(v)-f(w)}\le \LipConstF \abs{v-w}$}{
    that for all $v, w\in\R$ it holds that $\abs{\scrf(v)-\scrf(w)} \le (2\fc)^{-1} \LipConstF \abs{v-w}$\dott{}
    }
    Hence, \cite[Corollary~4.13]{ackermann2023deep}
    and \cite[Corollary~4.14]{ackermann2023deep}
    ensure 
    that for every $\varepsilon \in (0,1]$ there exists $\F_{0,\varepsilon} \in \ANNsm$ such that 
    \begin{enumerate}[label=(\Roman *)]
        \item 
        \label{cor_of_mainthm1_prop_NN_item1}
        it holds that $\functionANN{a}(\F_{0,\varepsilon}) \in C(\R,\R)$,
        \item
        \label{cor_of_mainthm1_prop_NN_item2}
        it holds for all $v,w\in\R$ that 
        $\abs{(\functionANN{a}(\F_{0,\varepsilon}))(v) - (\functionANN{a}(\F_{0,\varepsilon}))(w)} \le (2\fc)^{-1} \LipConstF \abs{v-w}$,
        \item
        \label{cor_of_mainthm1_prop_NN_item3}
        it holds for all $w \in \R$ that 
        $\abs{ (\functionANN{a}(\F_{0,\varepsilon}))(w) - \scrf(w) } \le 2 \varepsilon \max\cu{1,\abs{w}^p}$,
        and 
        \item
        \label{cor_of_mainthm1_prop_NN_item4}
        it holds that $\paramANN(\F_{0,\varepsilon}) \le 24 \bpr{\max\bcu{ 1, \fc^{-1} \LipConstF }}^{\frac{p}{p-1}} \varepsilon^{-\frac{p}{p-1}}$.
    \end{enumerate}
    \startnewargseq 
    \argument{\eg \cite[Lemma 2.4]{ackermann2023deep}; \cref{cor_of_mainthm1_prop_NN_item4}}{
    that for all $\varepsilon \in (0,1]$ it holds that 
    \begin{equation}
        \varepsilon^{\frac{p}{p-1}} \lengthANN(\F_{0,\varepsilon}) + \varepsilon^{\frac{p}{p-1}} \normmm{\dims(\F_{0,\varepsilon})} \le 24 \pr{\max\cu{1,\fc^{-1}\LipConstF}}^{\frac{p}{p-1}}
        \dott 
    \end{equation}
    }
    \argument{the triangle inequality; the fact that for all $v, w\in\R$ it holds that $\abs{\scrf(v)-\scrf(w)} \le (2\fc)^{-1} \LipConstF \abs{v-w}$; \cref{cor_of_mainthm1_prop_NN_item3}; the fact that for all $w\in\R$ it holds that $1+\abs{w}^p \le (1+\abs{w})^p$}{
    that for all $\varepsilon \in (0,1]$, $w\in\R$ it holds that 
    \begin{equation}\llabel{eq:2100}
        \begin{split}
            & \varepsilon \abs{ (\functionANN{a}(\F_{0,\varepsilon}))(w) }
            + \abs{ \scrf(w) - (\functionANN{a}(\F_{0,\varepsilon}))(w) } \\
            & \le 
            \varepsilon \bpr{ \abs{ (\functionANN{a}(\F_{0,\varepsilon}))(w) - \scrf(w) } 
            + \abs{ \scrf(w) - \scrf(0) } + \abs{\scrf(0)}
            }
            + \abs{ \scrf(w) - (\functionANN{a}(\F_{0,\varepsilon}))(w) } \\
            & \le 2 \abs{ (\functionANN{a}(\F_{0,\varepsilon}))(w) - \scrf(w) } 
            + \varepsilon \bpr{ (2\fc)^{-1} \LipConstF \abs{w} + (2\fc)^{-1} \abs{f(0)} } \\
            & \le 4 \varepsilon \max\cu{1,\abs{w}^p}
            + \varepsilon (2\fc)^{-1} \bpr{\LipConstF + \abs{f(0)} } \bpr{1+\abs{w}} \\
            & \le \varepsilon
            \bpr{ 4 + (2\fc)^{-1} \pr{\LipConstF + \abs{f(0)} } } (1+\abs{w})^p
            \dott 
        \end{split}
    \end{equation}
    }
    \argument{\lref{eq:2100}; \lref{eq:2240}}{
    that for all $\varepsilon \in (0,1]$, $x\in\R$ it holds that 
    \begin{equation}
        \begin{split}
            & \varepsilon 
            \norm{\nabla_x \scru_1(2\fc T,x)} 
            + \varepsilon 
            \abs{ (\functionANN{a}(\F_{0,\varepsilon}))(x) }
            + \abs{ \scrf(x) - (\functionANN{a}(\F_{0,\varepsilon}))(x) } \\
            & \le \varepsilon
            \bpr{ 2\constantAssumpMainThm + 4 + (2\fc)^{-1} \pr{\LipConstF + \abs{f(0)} } } (1+\abs{x})^p
            \dott 
        \end{split}
    \end{equation}
    }
    \argument{items~(iii) and~(v) in Lemma~3.5 in~\cite{ackermann2023deep}; item~(i) in Lemma~3.5 in~\cite{ackermann2023deep}; item~(ii) in Lemma~3.8 in~\cite{ackermann2023deep}}{
    that 
    there exists $\fJ \in \ANNsm$ such that 
    \begin{equation}
        \hiddenLength(\fJ)=1, 
        \qquad \dims(\fJ)=(1,2,1), 
        \qquad \text{and} \qquad \functionANN{a}(\fJ)=\id_{\R}
        \dott
    \end{equation}
    }
    \argument{\cref{ANN_for_product3}; \cref{ANN_for_product4}}{
    that for every $\varepsilon \in (0,1]$ there exists $\Gamma_{\varepsilon} \in \ANNsm$ 
    such that for all 
    $v,w\in\R$ it holds that 
    \begin{equation}
        \begin{split}
            &\functionANN{a}(\Gamma_\varepsilon) \in C(\R^2,\R), \qquad
            \paramANN(\Gamma_\varepsilon)\le
            1728 \cdot 2^{\tfrac{q^3+3q^2-2q}{(q-2)(q-1)}} 
            \varepsilon^{-\tfrac{q^2}{(q-2)(q-1)}},  \\
            &\text{and} \qquad 
            \babs{ vw - (\functionANN{a}(\Gamma_\varepsilon))(v,w) } \le \varepsilon \max\cu{ 1,\abs{v}^q, \abs{w}^q }
            \dott 
        \end{split}
    \end{equation}
    }
    \argument{\cref{ANN_for_hatfct1}; \cref{ANN_for_hatfct2}}{
    that for every $K\in\N$, $k\in\{0,1,\dots,K\}$, $\varepsilon \in (0,1]$ there exists 
    $\ANNhatfct_{K,k,\varepsilon} \in \ANNsm$ 
    such that for all 
    $t\in\R$ it holds that 
    \begin{equation}
        \begin{split}
            &\functionANN{a}(\ANNhatfct_{K,k,\varepsilon}) \in C(\R,\R), \qquad
            \bbabs{ \pr{\functionANN{a}(\ANNhatfct_{K,k,\varepsilon})}(t) - \scrL^{0,1,0}_{\frac{(k-1)T}{K},\frac{kT}{K},\frac{(k+1)T}{K}}(t) } 
            \le \varepsilon \max\{1,\abs{t}^q\}, \\
            & \text{and} \qquad
            \paramANN(\ANNhatfct_{K,k,\varepsilon})\le
            19 \cdot \bbbpr{\max\bbcu{1,\frac{4K}{T}}}^{\frac{q}{q-1}} 2^{\frac{q}{q-1}} 
            \varepsilon^{-\frac{q}{q-1}}
            \le 
            19 \cdot \bbpr{2+\frac{8}{T}}^{\frac{q}{q-1}} 
            K^{\frac{q}{q-1}} 
            \varepsilon^{-\frac{q}{q-1}}
            \dott 
        \end{split}
    \end{equation}
    } 
    \argument{\cref{theorem:main} (applied with 
    $\LipConstF \with (2\fc)^{-1} \LipConstF$, 
    $\constantAssumpMainThm \with \tilde{\constantAssumpMainThm}$, 
    $\alpha_0 \with \frac{p}{p-1}$, $\alpha_1 \with \constantAssumpMainThm$, $\beta_0 \with \frac{p}{p-1}$, $\beta_1 \with \constantAssumpMainThm$, $T\with 2\fc T$, $r\with \tilde{r}$, $p\with p$, $\fq \with \fq$, $q \with q$, 
    $a \with a$, $f_0 \with \scrf$, $(f_d)_{d \in \N} \with (\R^d  \ni x \mapsto \scru_d(2\fc T,x) \in \R)_{d \in \N}$, 
    $(\nu_d)_{d \in \N} \with (\nu_d)_{d \in \N}$, $\fJ \with \fJ$, $(\F_{d,\varepsilon})_{(d,\varepsilon) \in \N_0\times (0,1]} \with (\F_{d,\varepsilon})_{(d,\varepsilon) \in \N_0\times (0,1]}$, 
    $(\Gamma_{\varepsilon})_{\varepsilon \in (0,1]} \with (\Gamma_{\varepsilon})_{\varepsilon \in (0,1]}$, 
    $\delta \with 1$,
    $((\ANNhatfct_{K,k,\varepsilon})_{k\in\{0,1,\dots,K\}})_{(K,\varepsilon) \in \N\times (0,1]} \allowbreak\with \allowbreak ((\ANNhatfct_{K,k,\varepsilon})_{k\in\{0,1,\dots,K\}})_{(K,\varepsilon) \in \N\times (0,1]}$, 
    $\constantMainThmDelta \with 3\max\cu{\frac{p}{p-1},\constantAssumpMainThm} + 2 + 24\tilde{r} + 8$ 
    in the notation of \cref{theorem:main})}{
    that there exist $(\U_{d,\varepsilon})_{(d,\varepsilon) \in \N\times (0,1]} \subseteq \ANNsm$ and $c \in (0,\infty)$ which satisfy for all 
    $d \in \N$, $\varepsilon \in (0,1]$ that 
    $\functionANN{a}(\U_{d,\varepsilon}) \in C(\R^{d+1},\R)$, 
    $\paramANN(\U_{d,\varepsilon}) \le c d^c \varepsilon^{-c}$, and 
    \begin{equation}\llabel{eq:1424}
        \bbbbbr{ \int_{[0,2\fc T]\times \R^{d}} \abs{ \scru_d(t,x) - \pr{\functionANN{a}(\U_{d,\varepsilon})}(t,x) }^{\fq} \, \nu_d(\dxx t, \dxx x) }^{\nicefrac{1}{\fq}} 
        \le \varepsilon
        \dott 
    \end{equation}
    }
    \argument{\cref{lemma_time_reversal_ANN} (applied for every $d\in\N$, $\varepsilon\in(0,1]$
    with $T\with 0$, $c\with 2\fc$, $\F\with \U_{d,\varepsilon}$, 
    $a\with a$, $d\with d$, $\fd \with 2$, $\fJ \with \fJ$
    in the notation of \cref{lemma_time_reversal_ANN})}{
    that for every $d \in \N$, $\varepsilon \in (0,1]$ there exists $\tilde\U_{d,\varepsilon} \in \ANNsm$ 
    such that for all 
    $s \in [0,T]$, $x\in\R^d$ it holds 
    that 
    \begin{equation}\llabel{eq:0047b}
        \begin{split}
        & \functionANN{a}(\tilde\U_{d,\varepsilon}) \in C(\R^{d+1},\R), \qquad 
        \pr{\functionANN{a}(\tilde\U_{d,\varepsilon})}(s,x)
        = \pr{\functionANN{a}(\U_{d,\varepsilon})}(2\fc s,x), \\
        & \text{and} \qquad 
            \paramANN\bpr{\tilde\U_{d,\varepsilon}}
            \le 
            \paramANN\bpr{\U_{d,\varepsilon}}
            384 d^2 
            \le 384 c d^{c+2} \varepsilon^{-c}
            \le (384 c + 2) d^{(384 c + 2)} \varepsilon^{- (384 c + 2) }
            \dott 
        \end{split}
    \end{equation}
    }
    \argument{\lref{eq:defscru1}; \lref{eq:2207}; \lref{eq:0047b}; Fubini's theorem; a change of variables}{
    that for all $d\in\N$, $\varepsilon \in (0,1]$ it holds that 
    \begin{equation}\llabel{eq:2103}
        \begin{split}
            & \int_{[0,2\fc T]\times \R^{d}} \abs{ \scru_d(t,x) - \pr{\functionANN{a}(\U_{d,\varepsilon})}(t,x) }^{\fq} \, \nu_d(\dxx t, \dxx x) \\
            & = 
            \frac{1}{2\fc} \int_{\R^{d}} \int_0^{2\fc T}  \abs{ u_d\pr{(2\fc)^{-1}t,x} - \pr{\functionANN{a}(\tilde\U_{d,\varepsilon})}((2\fc)^{-1}t,x) }^{\fq} \, \dxx t\, \mu_d(\dxx x) \\
            & = 
            \int_{\R^{d}} \int_0^{T}  \abs{ u_d\pr{s,x} - \pr{\functionANN{a}(\tilde\U_{d,\varepsilon})}(s,x) }^{\fq} \, \dxx s \, \mu_d(\dxx x) 
            \dott 
        \end{split}
    \end{equation}
    }
    \argument{\lref{eq:2103}; Fubini's theorem; \lref{eq:1424}}{
    that for all $d\in\N$, $\varepsilon \in (0,1]$ 
    it holds that 
    \begin{equation}
        \bbbbbr{ \int_0^T \int_{\R^d} \abs{u_d(t,x)-\pr{\functionANN{a}(\tilde\U_{d,\varepsilon})}(t,x)}^{\fq} \, \mu_d(\dxx x)\, \dxx t  }^{\nicefrac{1}{\fq}}
        \le \varepsilon \dott 
    \end{equation}
    }
\end{aproof}

\cfclear
\begin{athm}{corollary}{cor_of_mainthm2}
Let 
$\constantAssumpMainThm, T, \fc, p \in (0,\infty)$, 
$b_1 \in\R$, $b_2 \in (b_1,\infty)$, 
$\leaky \in \R\backslash\{-1,1\}$, 
$\indexAct \in \{0,1\}$, 
let
$f\colon \R \to \R$ be Lipschitz continuous, 
for every $d \in \N$ let $u_d \in C^{1,2}([0,T]\times \R^d,\R)$
satisfy for all $t\in[0,T]$, $x \in \R^d$ that 
\begin{equation}\label{eq:cor2:heateq}
\tfrac{\partial}{\partial t} u_d(t,x)
= 
\fc \Delta_x u_d(t,x)
+
f\pr{ u_d(t,x) } , 
\end{equation}
let 
$a \colon \R\to\R$
satisfy for all $x \in \R$ that $a(x) = \indexAct \max\{x,\leaky x\} + (1-\indexAct)  \ln(1+\exp(x))$,  
and assume for all $d \in \N$, $\varepsilon \in (0,1]$ that there exists 
$\G \in \ANNsm$ such that for all $t \in [0,T]$, $x \in \R^d$ it holds that 
\begin{equation}
    \functionANN{a}(\G) \in C(\R^d,\R) , \qquad \paramANN(\G) \le \constantAssumpMainThm d^{\constantAssumpMainThm} \varepsilon^{-\constantAssumpMainThm}, \qquad \text{and}
\end{equation}
\begin{equation}\label{eq:2305}
    \varepsilon \norm{\nabla_x u_d(0,x)} 
    + \varepsilon \abs{u_d(t,x)} 
    + \abs{ u_d(0,x) - (\functionANN{a}(\G))(x) }
    \le \varepsilon \constantAssumpMainThm d^{\constantAssumpMainThm} \pr{1+\norm{x}}^{\constantAssumpMainThm}
\end{equation}
\cfload.
Then 
there exists 
$c \in \R$ 
such that for all $d \in \N$, $\varepsilon \in (0,1]$ 
there exists $\U \in \ANNsm$ 
such that 
\begin{equation}\label{eq:result1_cor2}
    \functionANN{a}(\U) \in C(\R^{d+1},\R), \qquad \paramANN(\U) \le c d^c \varepsilon^{-c}, \qquad \text{and} 
\end{equation}
\begin{equation}\label{eq:result2_cor2}
    \bbbbbr{ \int_{[0,T]\times [b_1,b_2]^d} \frac{\abs{u_d(y)-\pr{\functionANN{a}(\U)}(y)}^{p}}{\pr{b_2 - b_1}^d } \, \dxx y  }^{\nicefrac{1}{p}}
    \le \varepsilon \dott 
\end{equation}
\end{athm}

\cfclear
\begin{aproof} 
    Throughout this proof let $(\delta_{\varepsilon,\fp})_{(\varepsilon,\fp) \in (0,1] \times (0,2)}$
    satisfy for all $\varepsilon \in (0,1]$, $\fp \in(0,2)$ that 
    \begin{equation}\llabel{eq:proof_cor2_def_delta}
        \delta_{\varepsilon,\fp}
        = \varepsilon \bpr{\max\cu{1,T}}^{\frac{1}{2}-\frac{1}{\fp}}, 
    \end{equation} 
    for every $d\in\N$ let 
    $\mu_d \colon \cB(\R^d) \to [0,1]$ be the uniform distribution on $[b_1,b_2]^d$, 
    and 
    for every $d \in \N$ let $\scru_d \colon [0,T] \times \R^d \to \R$
    satisfy for all $t \in [0,T]$, $x \in \R^d$ that 
    \begin{equation}\llabel{eq:0048}
        \scru_d(t,x) = u_d(T-t,x)
        \dott
    \end{equation}
    \startnewargseq 
    \argument{the fact that for all $d \in \N$ it holds that $u_d \in C^{1,2}([0,T]\times \R^d,\R)$; 
    \cref{eq:cor2:heateq}}{
    that for all $d \in \N$, $t \in [0,T]$, $x\in\R^d$ it holds that 
    $\scru_d \in C^{1,2}([0,T]\times \R^d,\R)$ 
    and 
    \begin{equation}
        \tfrac{\partial}{\partial t} \scru_d(t,x)
        = 
        - \fc \Delta_x \scru_d(t,x)
        - f\pr{ \scru_d(t,x) }
        \dott 
    \end{equation}
    }
    \argument{\cref{eq:2305}}{
    that for all $d\in\N$, $\varepsilon \in (0,1]$, $t \in [0,T]$, $x \in \R^d$ it holds that 
    \begin{equation}
        \begin{split}
            \varepsilon \norm{\nabla_x \scru_d(T,x)} + \varepsilon \abs{\scru_d(t,x)}  
            + \abs{\scru_d(T,x)- \pr{\functionANN{a}(\G)}(x)}
            & \le \varepsilon \constantAssumpMainThm d^{\constantAssumpMainThm} \pr{1+\norm{x}}^{\constantAssumpMainThm}
        \end{split}
    \end{equation}
    }
    \argument{the fact that for all $d \in \N$ it holds that $\mu_d$ is the uniform distribution on $[b_1,b_2]^d$; \eg \cite[Lemma 3.15]{GrohsWurstemberger2018} }{
    that for all $d \in \N$, $\fq\in[2,\infty)$, $q\in (2,\infty)$ it holds that 
    \begin{equation}
        \begin{split}
            \int_{\R^d} \pr{1+ \norm{x}^{(\constantAssumpMainThm+1)^2 q\fq} }\, 
            \mu_d(\dxx x) 
            & = 
            1 + 
            \frac{1}{\abs{b_1-b_2}^d} \int_{[b_1,b_2]^d} \norm{x}^{(\constantAssumpMainThm+1)^2 q\fq} \, \dxx x \\
            & \le 
            1 + 
            d^{(\constantAssumpMainThm+1)^2 q\fq} 
            \max\bcu{ \abs{b_1}^{(\constantAssumpMainThm+1)^2 q\fq}, \abs{b_2}^{(\constantAssumpMainThm+1)^2 q\fq} } \\
            & \le 2 d^{(\constantAssumpMainThm+1)^2 q\fq} 
            \max\bcu{ 1, \abs{b_1}^{3\fq(\constantAssumpMainThm+1)^2}, \abs{b_2}^{3\fq (\constantAssumpMainThm+1)^2} } 
            \dott 
        \end{split}
    \end{equation}
    }
    \argument{\cref{cor_of_mainthm1} (applied for every $\fq \in [2,\infty)$ with  
    $T\with T$, $\fc \with \fc$,  $r\with 1$, $p \with \constantAssumpMainThm+1$, $\fq \with \fq$, 
    $q \with 3$, 
    $\alpha \with \leaky$, 
    $\nu \with \nu$, 
    $a \with a$,  
    $f \with f$, $(u_d)_{d\in\N} \with (\scru_d)_{d\in\N}$, $(\mu_d)_{d\in\N} \with (\mu_d)_{d\in\N}$, 
    $\G \with \G$,
    $\constantAssumpMainThm \with 2 \max\cu{ 1, \constantAssumpMainThm, \abs{b_1}^{3\fq (\constantAssumpMainThm+1)^2}, \abs{b_2}^{3\fq (\constantAssumpMainThm+1)^2} }$ 
    in the notation of \cref{cor_of_mainthm1})}{
    that there exist $(\U_{d,\varepsilon,\fq})_{(d,\varepsilon,\fq) \in \N\times (0,1]\times [2,\infty)} \subseteq \ANNsm$ and $(c_{\fq})_{\fq \in [2,\infty)} \subseteq \R$ which satisfy for all 
    $d \in \N$, $\varepsilon \in (0,1]$, $\fq \in [2,\infty)$ that 
    $\functionANN{a}(\U_{d,\varepsilon,\fq}) \in C(\R^{d+1},\R)$, 
    $\paramANN(\U_{d,\varepsilon,\fq}) \le c_{\fq} d^{c_{\fq}} \varepsilon^{-c_{\fq}}$, and 
    \begin{equation}\llabel{eq:1055}
        \bbbbr{ \int_0^T \int_{\R^{d}} \abs{ \scru_d(t,x) - \pr{\functionANN{a}(\U_{d,\varepsilon,\fq})}(t,x) }^{\fq} \, \mu_d(\dxx x) \, \dxx t }^{\nicefrac{1}{\fq}} 
        \le \varepsilon
        \dott 
    \end{equation}
    }
    \argument{items~(iii) and~(v) in Lemma~3.5 in~\cite{ackermann2023deep}; item~(i) in Lemma~3.5 in~\cite{ackermann2023deep}; item~(ii) in Lemma~3.8 in~\cite{ackermann2023deep}}{
    that 
    there exists $\fJ \in \ANNsm$ such that 
    $\dims(\fJ)=(1,2,1)$ and 
    $\functionANN{a}(\fJ)=\id_{\R}$\dott\, 
    }
    \argument{\cref{lemma_time_reversal_ANN} (applied for every $d\in\N$, $\varepsilon\in(0,1]$, $\fq \in [2,\infty)$ with $T\with T$, $c\with -1$, $\F\with \U_{d,\varepsilon,\fq}$, 
    $a\with a$, $d\with d$, $\fd \with 2$, $\fJ \with \fJ$
    in the notation of \cref{lemma_time_reversal_ANN})}{
    that for every $d \in \N$, $\varepsilon \in (0,1]$, $\fq \in [2,\infty)$ 
    there exists $\tilde\U_{d,\varepsilon,\fq}\in\ANNsm$ 
    such that for all 
    $s \in [0,T]$, $x\in\R^d$ it holds 
    that 
    \begin{equation}\llabel{eq:0047}
        \begin{split}
        & \functionANN{a}(\tilde\U_{d,\varepsilon,\fq}) \in C(\R^{d+1},\R), \qquad 
        \pr{\functionANN{a}(\tilde\U_{d,\varepsilon,\fq})}(s,x)
        = \pr{\functionANN{a}(\U_{d,\varepsilon,\fq})}(T-s,x), \\
        & \text{and} \qquad 
            \paramANN\bpr{\tilde\U_{d,\varepsilon,\fq}}
            \le 
            \paramANN\bpr{\U_{d,\varepsilon,\fq}}
            384 d^2 
            \le 384 c_{\fq} d^{c_{\fq}+2} \varepsilon^{-c_{\fq}}
            \le (384 c_{\fq} + 2) d^{(384 c_{\fq} + 2)} \varepsilon^{- (384 c_{\fq} + 2) }
            \dott 
        \end{split}
    \end{equation}
    }
    \argument{\lref{eq:0047}}{
    that for every $\fq \in [2,\infty)$ there exists $\tilde c_{\fq} \in\R$ 
    such that for 
    all $d\in\N$, $\varepsilon\in(0,1]$ 
    it holds that 
    \begin{equation}\llabel{eq:2008}
        \paramANN\pr{\tilde\U_{d,\varepsilon,\fq}} \le \tilde c_{\fq} d^{\tilde c_{\fq}} \varepsilon^{-\tilde c_{\fq}}
        \dott 
    \end{equation}
    }
    \argument{\lref{eq:0048}; \lref{eq:1055}; \lref{eq:0047}; Fubini's theorem; a change of variables; the fact that for all $d \in \N$ it holds that $\mu_d$ is the uniform distribution on $[b_1,b_2]^d$}{
    that for all $d\in\N$, $\varepsilon\in(0,1]$, $\fq \in [2,\infty)$ 
    it holds that
    \begin{equation}\llabel{eq:0050}
        \begin{split}
            \varepsilon
            & \ge 
            \bbbbbr{ \int_0^T \int_{\R^{d}} \abs{ u_d(T-t,x) - \pr{\functionANN{a}(\tilde\U_{d,\varepsilon,\fq})}(T-t,x) }^{\fq} \, \mu_d(\dxx x) \, \dxx t }^{\nicefrac{1}{\fq}} \\
            & = \bbbbbr{ \int_0^T \int_{\R^{d}} \abs{ u_d(s,x) - \pr{\functionANN{a}(\tilde\U_{d,\varepsilon,\fq})}(s,x) }^{\fq} \, \mu_d(\dxx x) \, \dxx s }^{\nicefrac{1}{\fq}} \\
            & = 
            \bbbbbr{ \int_{[0,T]\times [b_1,b_2]^d} \frac{\abs{ u_d(y) - \pr{\functionANN{a}(\tilde\U_{d,\varepsilon,\fq})}(y) }^{\fq}}{ \pr{b_2-b_1}^d } \, \dxx y }^{\nicefrac{1}{\fq}}
            \dott 
        \end{split}
    \end{equation}
    }
    \argument{\lref{eq:0050}; the fact that for all $\varepsilon\in(0,1]$, $\fp\in(0,2)$ it holds that $\delta_{\varepsilon,\fp} \in (0,1]$; Jensen's inequality}{
    that for all $d\in\N$, $\varepsilon\in(0,1]$, $\fp \in (0,2)$ 
    it holds that
    \begin{equation}\llabel{eq:2108}
        \begin{split}
            & \bbbbbr{ \int_{[0,T]\times [b_1,b_2]^d} \frac{\abs{ u_d(y) - \pr{\functionANN{a}(\tilde\U_{d,\delta_{\varepsilon,\fp},2})}(y) }^{\fp}}{ \pr{b_2-b_1}^d } \, \dxx y }^{\nicefrac{2}{\fp}} \\
            & \le 
            T^{\frac{2}{\fp}-1} 
            \bbbbbr{ \int_{[0,T]\times [b_1,b_2]^d} \frac{\abs{ u_d(y) - \pr{\functionANN{a}(\tilde\U_{d,\delta_{\varepsilon,\fp},2})}(y) }^{2}}{ \pr{b_2-b_1}^d } \, \dxx y }
            \le T^{\frac{2}{\fp}-1} \pr{\delta_{\varepsilon,\fp}}^2
            \dott 
        \end{split}
    \end{equation}
    }
    \argument{\lref{eq:2108}; \lref{eq:proof_cor2_def_delta}}{
    that for all $d\in\N$, $\varepsilon\in(0,1]$, $\fp \in (0,2)$ 
    it holds that
    \begin{equation}\llabel{eq:2008b}
        \begin{split}
            & \bbbbbr{ \int_{[0,T]\times [b_1,b_2]^d} \frac{\abs{ u_d(y) - \pr{\functionANN{a}(\tilde\U_{d,\delta_{\varepsilon,\fp},2})}(y) }^{\fp}}{ \pr{b_2-b_1}^d } \, \dxx y }^{\nicefrac{1}{\fp}} 
            \le T^{\frac{1}{\fp} - \frac12} \delta_{\varepsilon,\fp} 
            \le \varepsilon
            \dott 
        \end{split}
    \end{equation}
    }
    \argument{\lref{eq:proof_cor2_def_delta}; \lref{eq:2008}}{
    that for all $d\in\N$, $\varepsilon\in(0,1]$, $\fp \in (0,2)$, and $\hat c_{\fp} = \tilde c_2 \pr{\max\cu{1,T}}^{\pr{\frac{1}{\fp}-\frac12} \tilde c_2}$ 
    it holds that
    \begin{equation}\llabel{eq:0050b}
        \begin{split}
            \paramANN\bpr{\tilde\U_{d,\delta_{\varepsilon,\fp},2}} 
            & \le \tilde c_{2} d^{\tilde c_{2}}
            \varepsilon^{-\tilde c_2}
            \bpr{\max\cu{1,T}}^{\pr{\frac{1}{\fp}-\frac12} \tilde c_2} 
            \le \hat c_{\fp}
            d^{ \hat c_{\fp} }
            \varepsilon^{- \hat c_{\fp} }
            \dott 
        \end{split}
    \end{equation}
    }
    Combining this with \lref{eq:2008}, \lref{eq:0050}, and \lref{eq:2008b} establishes \cref{eq:result1_cor2} and \cref{eq:result2_cor2}.
\end{aproof}


\section*{Acknowledgments}

This project has been partially funded by the Deutsche Forschungsgemeinschaft (DFG, German Research Foundation) under Germany’s Excellence Strategy EXC 2044-390685587, Mathematics Münster: Dynamics-Geometry-Structure (third author). 
This project has also been partially funded by the Deutsche Forschungsgemeinschaft (DFG, German Research Foundation) in the frame of the priority programme SPP 2298 `Theoretical Foundations of Deep Learning' -- Project no.\ 464123384 (third author). 
This work has also been partially supported by the Internal Project Fund from Shenzhen Research Institute of Big Data under Grant T00120220001 (second author).

\bibliographystyle{acm}
\bibliography{bibfile}

\end{document}